\def\paperID{626}
\def\confName{ICCV}
\def\confYear{2025}

\def\paperTitle{Lost in Time: A New Temporal Benchmark for VideoLLMs}

\def\authorBlock{
    Daniel Cores\thanks{Equal contribution. $^{\dagger}$Research conducted while at VISLab, Unversity of Amsterdam.} $^{ \dagger 1}$\enspace
    Michael Dorkenwald$^{*2}$ \enspace
    Manuel Mucientes$^{1}$ \enspace
    Cees G. M. Snoek$^{2}$ \enspace
    Yuki M. Asano$^{3}$\\
    $^{1}$CiTIUS, University of Santiago de Compostela, 
    $^{2}$QUVA Lab, University of Amsterdam, \\
    $^{3}$Fundamental AI Lab, University of Technology Nuremberg 
    \\
    Benchmark available at: \href{https://huggingface.co/datasets/FunAILab/TVBench}{https://huggingface.co/datasets/FunAILab/TVBench}\\
}

\newif\ifreview 
\newif\ifarxiv \newcommand{\arxiv}{\arxivtrue}
\newif\ifcamera 
\newif\ifrebuttal 

\arxiv 

\RequirePackage{iftex}
\ifPDFTeX\pdfoutput=1\fi
\documentclass[10pt,twocolumn,letterpaper]{article}
\ifreview \usepackage[review]{cvpr} \fi
\ifarxiv \usepackage[pagenumbers]{cvpr} \fi
\ifrebuttal \usepackage[rebuttal]{cvpr} \fi
\ifcamera \usepackage{cvpr} \fi

\usepackage[T1]{fontenc}

\usepackage{graphicx}	
\usepackage{amsmath}	
\usepackage{amssymb}	
\usepackage{booktabs}
\usepackage{longtable}
\usepackage{times}
\ifXeTeX
    \catcode`—=\active \protected\def—{\textemdash}
    \catcode`“=\active \protected\def“{\textquotedblleft}
    \catcode`”=\active \protected\def”{\textquotedblright}
    \catcode`’=\active \protected\def’{\textquoteright}
\fi
\usepackage{microtype}
\usepackage{epsfig}
\usepackage{caption}
\usepackage{float}
\usepackage{placeins}
\usepackage{color, colortbl}
\usepackage{stfloats}
\usepackage{enumitem}
\usepackage{tabularx}
\usepackage{xstring}
\usepackage{multirow}
\usepackage{xspace}
\usepackage{url}
\usepackage{xurl}
\usepackage{subcaption}
\usepackage{xcolor}
\usepackage[hang,flushmargin]{footmisc}

\usepackage{amsthm} 
\usepackage{mdframed} 
\usepackage{tikz}
\usepackage{makecell}

\ifcamera \usepackage[accsupp]{axessibility} \fi

% ...

\ifarxiv  \fi

\newcommand{\R}[1]{{%
    \textbf{%
        \ifstrequal{#1}{1}{\textcolor{red}{R#1}}{%
        \ifstrequal{#1}{2}{\textcolor{blue}{R#1}}{%
        \ifstrequal{#1}{3}{\textcolor{magenta}{R#1}}{%
        \ifstrequal{#1}{4}{\textcolor{teal}{R#1}}{%
                           \textcolor{cyan}{R#1}%
        }}}}%
    }%
}}

\newcounter{theo}
\renewcommand{\thetheo}{\arabic{theo}}
\newenvironment{theo}[2][]{%
\refstepcounter{theo}%
\ifstrempty{#1}%
{\mdfsetup{%
frametitle={%
\tikz[baseline=(current bounding box.east),outer sep=0pt]
\node[anchor=east,rectangle,fill=red!20]
{\strut Problem~\thetheo};}}
}%
{\mdfsetup{%
frametitle={%
\tikz[baseline=(current bounding box.east),outer sep=0pt]
\node[anchor=east,rectangle,fill=red!20]
{\strut Problem~\thetheo:~#1};}}%
}%
\mdfsetup{innertopmargin=1pt,linecolor=red!20,%
linewidth=2pt,topline=true,%
frametitleaboveskip=\dimexpr-\ht\strutbox\relax
}
\begin{mdframed}[]\relax%
\label{#2}}{\end{mdframed}}

\usepackage{xr-hyper}

\makeatletter
\newcommand*{\addFileDependency}[1]{
  \typeout{(#1)}
  \@addtofilelist{#1}
  \IfFileExists{#1}{}{\typeout{No file #1.}}
}

\makeatother
\newcommand*{\myexternaldocument}[1]{
    \externaldocument{#1}
    \addFileDependency{#1.tex}
    \addFileDependency{#1.aux}
}

\definecolor{cvprblue}{rgb}{0.21,0.49,0.74}
\usepackage[pagebackref,breaklinks,colorlinks,allcolors=cvprblue]{hyperref}
\usepackage[capitalize]{cleveref}
\crefname{section}{Sec.}{Secs.}
\crefname{table}{Table}{Tables}
\crefname{figure}{Fig.}{Figs.}

\ifarxiv \crefname{appendix}{App.}{Apps.}
\else \crefname{appendix}{Suppl.}{Suppls.} \fi

\ifarxiv
\fi

\unless\ifarxiv \myexternaldocument{_supplementary} \fi

\begin{document}
\title{\paperTitle}
\author{\authorBlock}

\maketitle
\unless\ifarxiv \vspace{-0.1cm} \fi
\noindent\begin{minipage}[t]{\columnwidth}

\begin{abstract}
Large language models have demonstrated impressive performance when integrated with vision models even enabling video understanding. However, evaluating video models presents its own unique challenges, for which several benchmarks have been proposed. In this paper, we show that the currently most used video-language benchmarks can be solved without requiring much temporal reasoning. We identified three main issues in existing datasets: (i) static information from single frames is often sufficient to solve the tasks (ii) the text of the questions and candidate answers is overly informative, allowing models to answer correctly without relying on any visual input (iii) world knowledge alone can answer many of the questions, making the benchmarks a test of knowledge replication rather than video reasoning. In addition, we found that open-ended question-answering benchmarks for video understanding suffer from similar issues while the automatic evaluation process with LLMs is unreliable, making it an unsuitable alternative. As a solution, we propose TVBench, a novel open-source video multiple-choice question-answering benchmark, and demonstrate through extensive evaluations that it requires a high level of temporal understanding. Surprisingly, we find that most recent state-of-the-art video-language models perform similarly to random performance on TVBench, with only a few models such as Qwen2-VL, and Tarsier clearly surpassing this baseline.
\begingroup\textcolor{red!70!black}{\textbf{November 2026 update.}} We add 59 full evaluations across 35 models and reports both task-macro and question-micro accuracy alongside published results. With thinking enabled, task-macro accuracy reaches 87.35\% for Gemini 4 Carbon and 79.09\% for Qwen3.8-Flash-Next.\endgroup
\end{abstract}

\end{minipage}\par
\unless\ifarxiv \vspace{-0.6cm} \fi
\section{Introduction}
\label{sec:intro}

\begin{figure}[t]
    \centering
\includegraphics[width=1\linewidth]{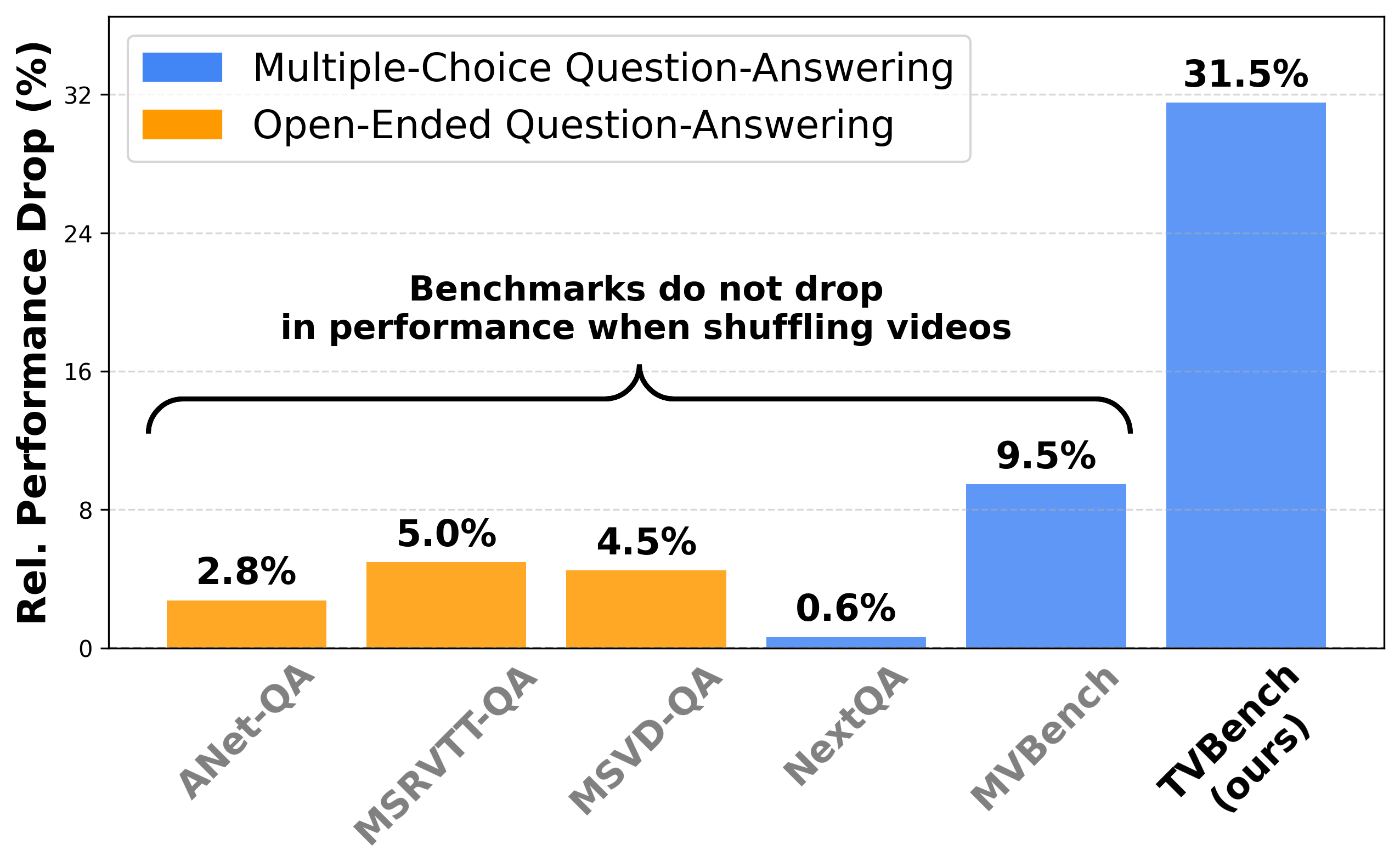}
    \vspace{-0.7cm}
    \caption{\textbf{Existing videoLLM benchmarks are time-invariant.} The performance of a SOTA model \citep{wang2024tarsierrecipestrainingevaluating} on commonly used benchmarks hardly drops when shuffling the input videos. This suggests that these benchmarks do not effectively measure temporal understanding. In contrast, in our proposed TVBench, shuffling input frames results in random accuracy, as it should be.}
    \label{fig:fig1}
    \vspace{-0.5cm}
\end{figure} 

Vision language models \citep{alayrac2022flamingo, openai2024gpt4technicalreport, liu2023llava} have gained popularity, benefiting from both the progress made in natural language processing \cite{gpt3, llama, bert_devlin} and the surge of foundation models for vision \citep{dino, clip, sigclip} tasks with strong generalization capabilities. Recently, video-language models have been introduced \citep{xu2021videoclip, lin2023videollava, xue2023clip}, aiming to replicate the success achieved in the image domain. To evaluate their performance, visual question answering has emerged as a key task requiring both textual and visual reasoning. 
With the rapid model development and release cycles, having a reliable and robust benchmark is crucial in measuring progress and guiding research efforts.


There are two main approaches to designing question-answering benchmarks for videos: multiple-choice question answering (MCQA)~~\citep{wang2024videohallucer, li2024mvbench, xiao2021next, mangalam2024egoschema, patraucean2024perception} and open-ended question answering (OEQA) ~\citep{yu2019activitynet, xu2017video, xu2016msr, wu2017deep}.
Given the critical role these benchmarks play in evaluating video understanding, their reliability is paramount. This raises an important question: To what extent do they truly capture and assess video understanding?

Previous analysis in image question answering benchmarks~\citep{goyal2017making}  has demonstrated that poorly formulated benchmarks could bias the development of new models towards learning strong text representations while ignoring visual information. This is especially relevant for the video-language community, where benchmarks must account not only for visual but also for temporal understanding. 

In this work, we conduct a comprehensive analysis of widely used video question-answering benchmarks, revealing that temporal information is poorly evaluated (see Fig.~\ref{fig:fig1}). Furthermore, in MCQA tasks, prior world knowledge, combined with overly informative questions and answer choices, often allows questions to be answered solely through text without the need for visual input. Our results also indicate that automatic open-ended evaluation is unreliable, with significant evaluation discrepancies in results across different models for the same task.
%

We reveal the shortcomings of existing benchmarks such as MVBench \citep{li2024mvbench}, NextQA \cite{xiao2021next}, MSVD-QA \citep{wu2017deep}, MSRVTT-QA \citep{xu2016msr} and ActivityNet QA \citep{yu2019activitynet} and based on those insights propose a new benchmark, TVBench, that \textit{requires} temporal understanding to be solved, providing an effective evaluation tool for current video-language models:  
\begin{itemize}
    \item We provide only temporal challenging candidate answers, requiring models to leverage temporal information to answer correctly.
    \item {We design task-specific templates to generate questions that are not overly informative such that they cannot be answered solely by text.}
    \item We design questions that can only be answered from the video content, without relying on prior world knowledge.

\end{itemize}
As a result, TVBench measures the temporal understanding of video-language models in contrast to previous benchmarks. In this setting, text-only and single-frame models, such as Gemini 1.5 Pro and GPT-4o, perform at random chance levels on TVBench despite achieving competitive results on other benchmarks. Surprisingly, even recent state-of-the-art video-language models perform close to random chance on TVBench, with only a few models, such as Qwen2-VL \citep{wang2024qwen2vlenhancingvisionlanguagemodels} and Tarsier \citep{wang2024tarsierrecipestrainingevaluating}, outperforming the random baseline. Shuffling and reversing the videos for these models lead to significant performance drops, unlike prior benchmarks, further verifying TVBench as a temporal video benchmark.





\unless\ifarxiv \vspace{-0.15cm} \fi
\section{Related Work}
\label{sec:related}

Traditional video evaluation benchmarks focused on specific tasks such as action recognition~\citep{goyal2017something, kay2017kinetics} or video description~\citep{das2013thousand, xu2016msr, wang2019vatex}. With the emergence of Vision Language Models (VLMs), there is a growing need for more comprehensive evaluation protocols to effectively evaluate models with increasingly advanced generalization capabilities. Current video-language benchmarks focus on solving QA pairs that require a certain level of multimodal understanding. There are two major trends in the QA format: open-ended QA and multiple-choice QA (MCQA).

\noindent \textbf{Open-ended question answering.} Evaluating open-ended QA introduces new challenges, as traditional evaluation metrics such as ROUGE \citep{lin2004rouge}, METEOR \citep{banerjee2005meteor}, and CIDEr \citep{vedantam2015cider} fail to analyze discrepancies of more complex and elaborated answers. Alternatively, \citet{maaz2023video} introduces a novel quantitative evaluation pipeline for open-ended QA datasets. The proposed method relies on GPT-3.5 to determine the correctness of the predicted answer and provides a matching score with the ground truth. Commonly used datasets for evaluating models in this context include MSRVTT-QA \citep{xu2017video}, MSVD-QA \citep{xu2017video}, TGIF-QA \citep{jang2017tgif} and ActivityNet-QA \citep{yu2019activitynet}. In general, any open-ended QA benchmarks can be evaluated following this protocol. Our analysis shows that Large Language Model (LLM) based evaluations are prone to hallucinations, leading to unreliable conclusions. In contrast, MCQA benefits from a more straightforward evaluation process based on the accuracy score.  

\noindent \textbf{Multiple-choice question answering.} CLEVRER \citep{yi2019clevrer} assesses reasoning about object interaction in synthetic videos. Perception Test \citep{patraucean2024perception} was introduced to evaluate visual perception in multimodal settings, mainly in indoor scenes. \citep{piyush_test_of_time} uses synthetically generated data to evaluate the temporal understanding of early video-language models.
EgoSchema \citep{mangalam2024egoschema} focuses on MCQA involving long egocentric videos. Recently, VideoHallucer \citep{wang2024videohallucer} was introduced as a first attempt to define a video-language benchmark specifically designed for hallucination detection.
MVBench~\citep{li2024mvbench} aims to evaluate temporal understanding by defining 20 dynamic tasks designed to require temporal reasoning throughout the entire video. However, our experiments demonstrate that many of these tasks are highly spatial and textual biased, failing to evaluate temporal understanding effectively. We propose a new benchmark that requires a high level of spatiotemporal understanding across different tasks in order to be solved.
\unless\ifarxiv \vspace{-0.15cm} \fi
\section{Problems in Video MCQA Benchmarks}
\label{sec:mcqa}

In this section, we identify two key shortcomings in current video multiple-choice question-answering (MCQA) benchmarks, as demonstrated on MVBench~\citep {li2024mvbench} and NextQA~\citep{xiao2021next}. First, we show that these benchmarks contain strong spatial bias, meaning that questions can be answered without requiring temporal understanding. Secondly, we demonstrate that they also contain a strong textual bias, as many questions can be answered without even looking at the visual input.
  
\subsection{Does Time Matter?}
\label{does_time_matter}


Video benchmarks must define tasks that cannot be solved using solely spatial information to evaluate the temporal understanding of a model effectively. In video MCQA, questions should not be answerable using spatial details from single random or multiple frames, e.g., after shuffling them. However, if no understanding of the sequence of events and temporal localization is needed, the benchmark fails to assess temporal understanding, focusing only on spatial information, which we define as spatial bias.

We analyze the spatial bias in MVBench using state-of-the-art image and video-language models such as GPT-4o~\citep{openai2024gpt4technicalreport}, Gemini 1.5 Pro~\citep{geminiteam2024gemini15unlockingmultimodal}, and Tarsier-34B~\citep{wang2024tarsierrecipestrainingevaluating}. In Table~\ref{tab:mvbench_static_bias}, we focus on four different tasks: i) fine-grained (FG) action, ii) scene transition, iii) fine-grained (FG) pose, and iv) episodic reasoning. In Table \ref{tab:tarsier_nextqa}, we report the performance of the whole NextQA dataset.
To assess whether solving these tasks requires temporal understanding, we compare the performance of such models when receiving only a single random frame, the shuffled videos, and the original videos.

\begin{table}[t]
\setlength{\tabcolsep}{0.2em}
    \centering
    \small
    \resizebox{\columnwidth}{!}{%
    \begin{tabular}{lcccccc}
        \toprule
         & Input & \makecell[c]{FG\\Action} &  \makecell[c]{Scene\\Transition} & \makecell[c]{FG\\Pose} & \makecell[c]{Episodic\\Reasoning} & Average\\
         \midrule
          Random & --& 25.0 & 25.0 & 25.0 & 20.0 &  23.8\\
         \midrule
         Gemini 1.5 & \multirow{3}{*}{image}  & 47.0 & 78.0 & 46.5 & 56.5 & 57.0\\
         GPT-4o &   & 49.0 & 84.0 & 53.0 & 65.0 & 62.8\\
         Tarsier-34B &  & 48.5 & 67.0 & 22.5 & 46.0 & 46.0\\
         \midrule
         Gemini 1.5 &  \multirow{3}{*}{shuffle} &49.5 & 90.0 & 54.5 & 63.0 & 64.3 \\
         GPT-4o & & 52.0 & 84.5 & 69.0 & 64.5 & 67.5\\
         Tarsier-34B &  &  51.0 & 89.0 & 56.5 & 51.5 & 62.0 \\
         \midrule
         Gemini 1.5 & \multirow{3}{*}{video}  & 50.0 & 93.3 & 58.5 & 66.8 & 67.2 \\
         GPT-4o &   & 51.0 & 83.5 & 65.5 & 63.0 & 65.8 \\
         Tarsier-34B &  & 48.5 & 89.5 & 64.5 & 54.5 & 64.3\\
         \bottomrule
    \end{tabular}}
        \vspace{-0.2cm}
    \caption{\textbf{Spatial bias of MVBench.} Our analysis reveals that temporal understanding is not required for solving these temporal tasks as they can be solved with a random frame or shuffled video. The average represents the mean across these four tasks. }
    \label{tab:mvbench_static_bias}
    \vspace{-0.5cm}
\end{table}

The models receiving only a random frame as input show strong performance across all four tasks in Table~\ref{tab:mvbench_static_bias}, surpassing the random baseline. GPT-4o achieves the highest average performance of 62.8\% across the four tasks,  nearly matching its video performance with 65.8\% and other state-of-the-art video-language models. The lower image performance of Tarsier-34B might stem from its training data composition, which contains five times more video data than image data.
These findings are unexpected, as task names like \textit{Fine-grained Action} suggest a need for temporal understanding. For this fine-grained task, the image-model GPT-4o achieves 49\%, which is even slightly better than the state-of-the-art Tarsier model, which scores 48.5\%. Similarly, for the other three tasks. Overall, GPT-4o achieves an average accuracy across all 20 tasks of 47.8\%, which is 20.5\% higher than the random performance of 27.3\% on MVBench. Similarly, on NextQA in Table \ref{tab:tarsier_nextqa}, the Tarsier model significantly outperforms the random baseline of 20.0\% with 71.3\%, processing a single random frame. These results indicate that a large portion of both benchmarks is affected by spatial bias.

Additionally, shuffling the videos has minimal impact on the MVBench performance of all video-language models, with an average difference of 2.3\%, indicating that temporal information is not necessary to solve these tasks. Similarly, for NextQA, the Tarsier model achieves the same performance for shuffling or non-shuffling. Note, as confirmed in Sec.~\ref{tvbench:evaluation}, the Tarsier model shows a significant drop in performance when videos are shuffled for tasks that require temporal understanding.
This problem goes beyond these four tasks as shown in Table~\ref{tab:sota}, Gemini 1.5 Pro and Tarsier achieve an average accuracy across all 20 MVBench tasks of 60.5\% and 67.6\%, respectively. Shuffling video frames causes a performance drop of only 3.8\% and 6.4\%, respectively, indicating that the spatial bias affects not only the tasks analyzed in Table~\ref{tab:mvbench_static_bias} but the entire dataset. Additionally, we verify the agreement between the correct responses of Tarsier-34B across modalities: 91.0\% between image and video inputs, and 93.9\% between video and shuffled video. This confirms that current models heavily rely on spatial biases to solve MVBench.


\begin{figure}
    \centering
    \includegraphics[width=1\linewidth]{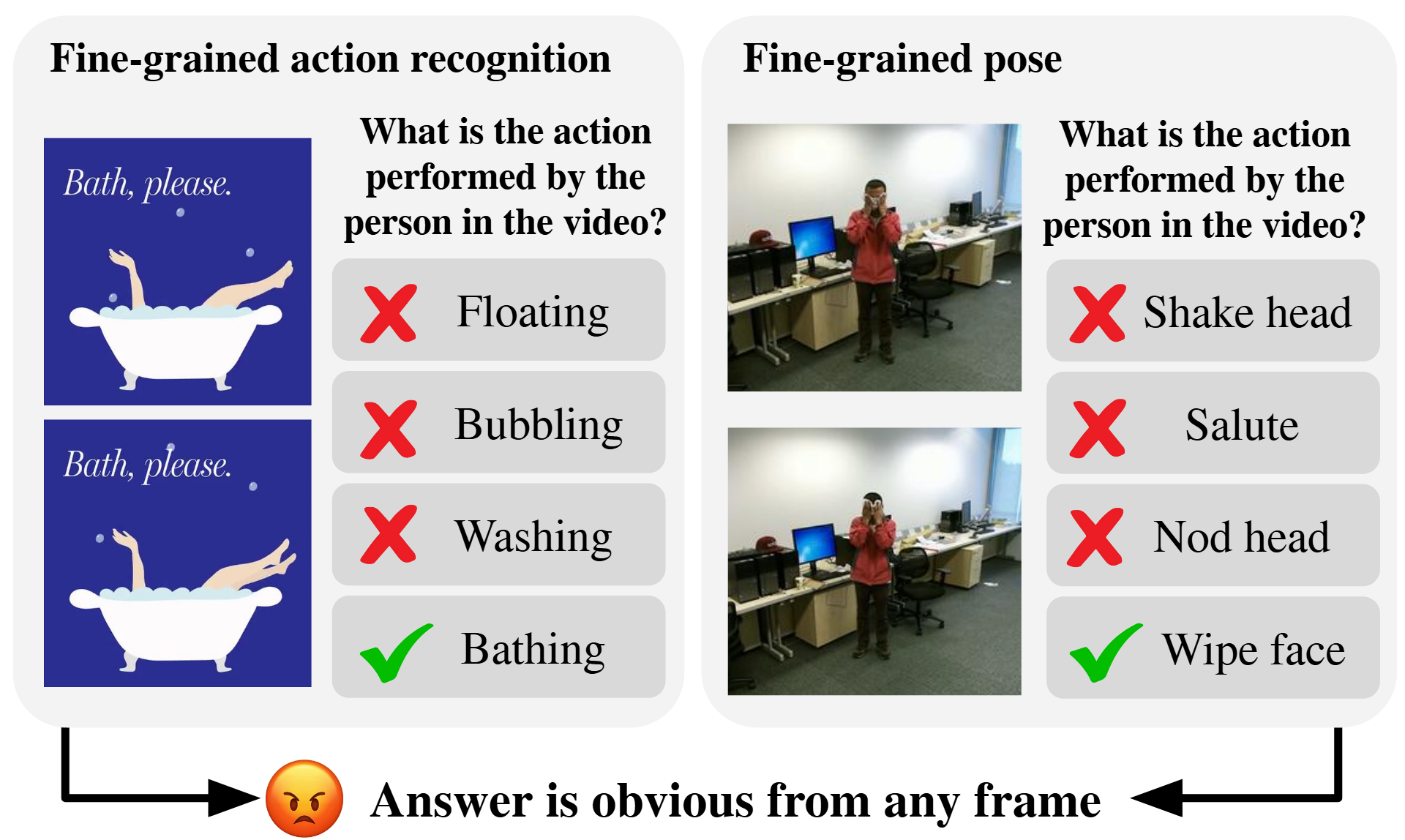}
    \vspace{-0.6cm}
    \caption{\textbf{Spatial bias of MVBench.} We show different tasks of the MVBench benchmark and observe that the question can be answered without requiring temporal understanding.}
    \vspace{-0.6cm}
    \label{fig:static_bias}
\end{figure}

In Fig.~\ref{fig:static_bias} we show examples corresponding to tasks analyzed in Table~\ref{tab:mvbench_static_bias}. 
The example from the left is taken from the \textit{Fine-grained Action recognition} task
implying that temporal understanding is needed.
The example on the right is from the \textit{Fine-grained Pose} task. In both cases, the information provided from any frame is enough to correctly answer the question. In the Appendix, in Fig.~\ref{fig:tvbench_first} -\ref{fig:tvbench_spatial_bias_last} we show 34 more examples of spatial bias in MVBench.

\begin{theo}{thm:temporal_problem}
MVBench and NextQA have a strong spatial bias, meaning questions can be answered without requiring temporal understanding.
\label{theo:problem1}
\end{theo}




\vspace{-0.4cm}
\subsection{Does Vision Matter?}
\label{sec:vision_matter}

Video benchmarks must be designed to prevent questions from being answered solely through common sense reasoning. Modern LLMs possess strong reasoning skills, which can exploit the information within the question and candidate sets in MCQA video language evaluation benchmarks. This creates textual bias, enabling models to answer questions without leveraging the video content.

\begin{figure}
    \centering
    \includegraphics[width=1\linewidth]{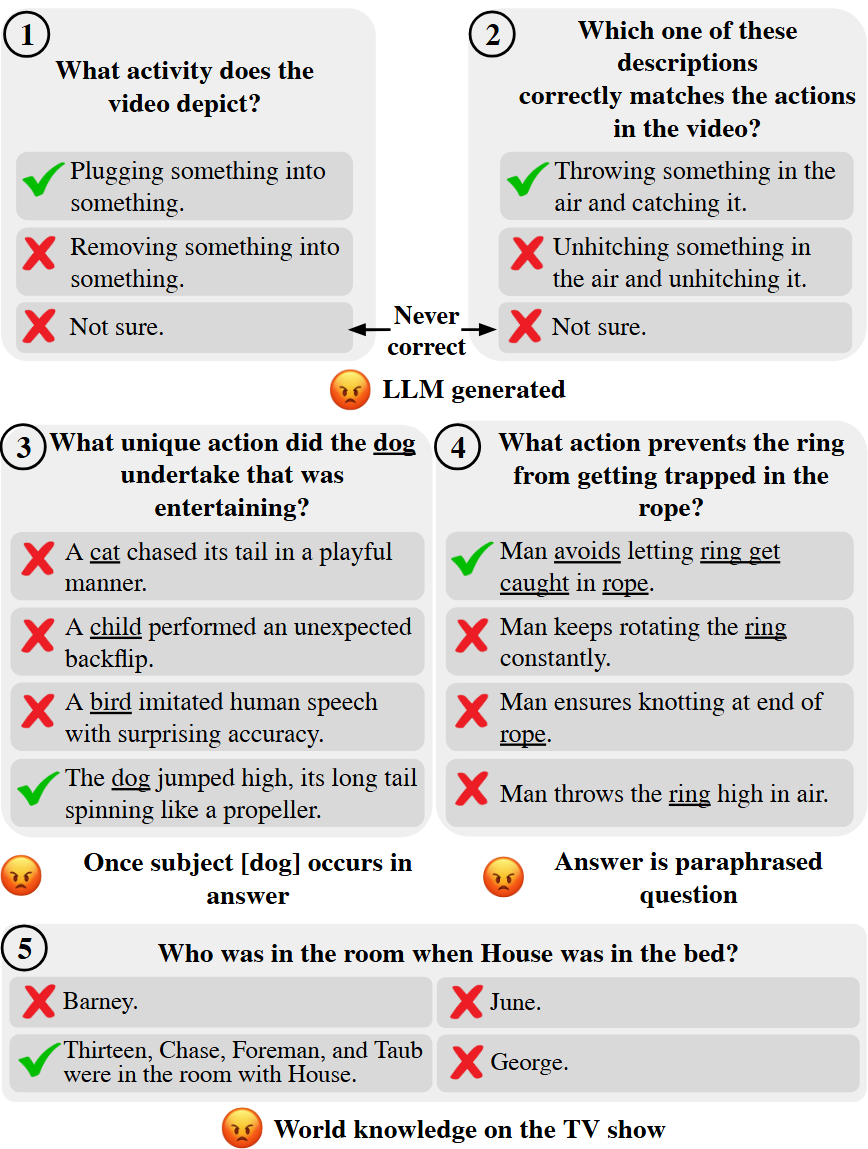}
    \caption{{\textbf{Textual bias of MVBench.} We show the shortcomings of the QA generated by MVBench and find that questions can be answered without considering the visual part.}}
    \vspace{-0.6cm}
    \label{fig:text-bias}
\end{figure}

We analyze the impact of textual bias on MVBench in Table~\ref{tab:mvbench_text_bias}. We evaluate the performance of state-of-the-art text-only LLMs, Llama 3~\citep{dubey2024llama3herdmodels}, and multi-modal LLMs such as Gemini 1.5 Pro~\citep{geminiteam2024gemini15unlockingmultimodal}, GPT-4o~\citep{openai2024gpt4technicalreport} and Tarsier~\citep{wang2024tarsierrecipestrainingevaluating}. Our findings reveal that LLMs can eliminate incompatible candidates easily, greatly outperforming the random baseline. 
Models using only text achieve competitive results compared to video-language models across these four tasks (Action Count, Unexpected Action, Action Antonym, and Episodic Reasoning). For instance, Gemini 1.5 Pro achieves an average performance of 62.3\% using text-only, compared to Tarsier-34B's 67.4\% using videos. Additionally, we verify an 85.3\% agreement between Tarsier-34B's correct text and video responses, confirming its strong reliance on textual biases in MVBench.
This goes beyond the four tasks, as Gemini Pro 1.5 achieves an average performance across all 20 tasks of 38.2\% {with text-only}, which is 10.9\% higher than the random chance baseline of 27.3\%. Similarly, for NextQA in Table \ref{tab:tarsier_nextqa}, Tarsier achieves 47.6\% performance across the whole dataset, an increase of 27.6\% over the random baseline. We have identified three key sources of this textual bias on MVBench:

\noindent \textbf{Bias from LLM-based QA generation.} Collecting and manually annotating large datasets for training and evaluation is very costly. Automatic and semi-automatic collection and annotation processes are commonly used~\citep{li2024mvbench, mangalam2024egoschema}. This includes techniques such as automatic QA pair generation with LLMs. ChatGPT plays a fundamental role in QA generation for 11 of the 20 tasks in the MVBench dataset. However, this introduces unrealistic candidates and QA pairs with excessive information. Fig.~\ref{fig:text-bias} presents examples of QA pairs that can be resolved merely with text information. Questions 1 and 2 belong to the \textit{Action Antonym} task, where an LLM is prompted to generate the antonym of the actual action shown in the video. The answers generated are either unrealistic, as one cannot ``remove something into something,'' or consistently incorrect, such as ``not sure.''
\begin{table}[t]
    \small
    \centering
        \setlength{\tabcolsep}{1.5mm} 

        \resizebox{\columnwidth}{!}{ 

    \begin{tabular}{lcccccc}
        \toprule
         & Input & \makecell[c]{Action\\Count} &  \makecell[c]{Unexp.\\Action} & \makecell[c]{Action\\Antonym} & \makecell[c]{Episodic\\Reasoning} & Average \\
         \midrule
         Random & -- & 33.3 & 25.0 & 33.3 & 20.0 & 27.9 \\
         \midrule
         Llama3 70B & \multirow{4}{*}{\shortstack{text- \\ only}}
         & 44.5 & 63.5 & 74.5 & 50.5 & 58.3 \\
         Gemini 1.5 & & 49.0 & 68.0 & 85.5 & 49.0 & 62.3\\
         GPT-4o & & 44.0 & 69.5 & 57.5 & 51.5 & 55.6\\
         Tarsier-34B & & 37.0 & 39.5 & 66.0 & 44.0 & 46.6\\
         \midrule
         Gemini 1.5 & \multirow{3}{*}{video} & 41.2 & 82.4 & 64.5 & 66.8 & 63.7\\
         GPT-4o & & 43.5 & 75.5 & 72.5 & 63.0 & 63.6\\
         Tarsier-34B & &  46.5 & 72.0 & 97.0 & 54.5 & 67.4\\
         \bottomrule
    \end{tabular}
    }
    \vspace{-0.2cm}
    \caption{\textbf{Textual bias of MVBench.} Our analysis reveals that vision is not required for solving tasks from MVBench as text-only LLMs score high above the random baseline and nearly on par with video models. The average is with respect to these four tasks.}
    \label{tab:mvbench_text_bias}
    \vspace{-0.5cm}
\end{table}
Questions 3 and 4 are from the \textit{Unexpected Action} task, where the LLM is prompted to generate textual questions and candidates based on the dataset annotations. However, in Question 3, the subject inquired in the question only occurs in the correct answer, while in Question 4, one of the candidates is a paraphrased version of the question. Hence, the text-only model can identify the correct answer without visual information.

\noindent \textbf{Bias from unbalanced sets.} Unbalanced QA sets also hinder a robust evaluation process. For instance, the correct answer for the Action Count task on MVBench is `3' for 90 out of 200 questions, while `9' is only the correct answer for one question. A model with a similar bias might get higher results than random by chance. We have observed in our experiments that some text-only models such as GPT-4o have this bias, predicting `3' for 88 out of 200 samples. This makes GPT-4o perform on par with the best video model with an accuracy of 44.0\% and 46.5\% respectively.

\noindent \textbf{Overreliance on world knowledge in questions.}
Video benchmarks should ensure models cannot rely solely on memorized world knowledge from an LLM to guess answers without using visual input. Even with well-designed questions, models might bypass visual reasoning and rely on prior knowledge to answer correctly.
An example of this can be seen in question 5 of Fig.~\ref{fig:text-bias}. The question does not exhibit an obvious bias in the QA generation. Still, it can be correctly answered if the model has world knowledge of the TV show from which the question was derived as answer 2 are character names from the House TV show.  

In the Appendix, in Fig.~\ref{fig:text_bias_one} -\ref{fig:tvbench_last}, we show 26 more examples of textual bias in MVBench.

\begin{theo}{thm:visual_problem}
MVBench and NextQA can be partially solved without visual information due to the bias from LLM QA generation, unbalanced dataset, and overreliance on world knowledge.
\label{theo:problem2}

\end{theo}
\vspace{-0.3cm}
\section{Open-ended QA to the rescue?}
\vspace{-0.2cm}
\label{sec:oeqa}

\begin{table}[t]
    \centering
    \small
    \begin{tabular}{lcc}
        \toprule
     & Input & NextQA \\
        \midrule
        Random & -- & 20.0 \\
    \midrule
        \multirow{5}{*}{Tarsier-34B} &text-only  & 47.6 \\
         & image & 71.3 \\
         & video shuffle  & 78.5 \\
         & video reverse  & 77.6 \\
         & video & 79.0 \\
        \bottomrule
    \end{tabular}
        \vspace{-0.2cm}
    \caption{\textbf{Spatial and textual bias on NextQA~\citep{xiao2021next}.} We find a strong spatial bias, with image performance nearly on par with video, and a textual bias with a strong improvement on random.}
    \label{tab:tarsier_nextqa}
    \vspace{-0.5cm}
\end{table}

Contrary to multiple-choice question answering (MCQA), open-ended question answering can be seen as an alternative to solving the aforementioned issues. Without a predefined candidate answer set, the model cannot rely on textual information to eliminate implausible candidates. However, open-ended evaluation presents new challenges compared to MCQA. Following \citet{maaz2023video}, LLMs have been widely used for the evaluation of open-ended question-answering in video datasets such as MSVD-QA \citep{wu2017deep}, MSRVTT-QA \citep{xu2016msr} and ActivityNet QA \citep{yu2019activitynet}. Specifically, \citet{maaz2023video} proposed GPT-3.5 as the evaluator model, which makes the entire evaluation process rely on a private API model. The evaluation model is prompted to determine if the predicted answer is correct given the question and the ground-truth answer. In addition, the evaluator also computes a score to measure the answer quality.

We conducted a comparative analysis to assess the influence of the evaluation model on the results. Table~\ref{tab:o_qa_diff_evals} shows the accuracy and average score for different models on three open-ended datasets, using two evaluators: GPT-3.5 and Llama3-70B. The evaluators produced significantly different results for the same method on the same dataset, with discrepancies of more than 20 points. Specifically, Llama3 highly increases the accuracy of text-only and single-image models, while providing similar or even lower results than GPT-3.5 for video models. It is also evident that Llama3 assigns better metrics to predictions made by the same model. If both the prediction model and the evaluator contain similar biases, the hallucinations in the predictions may be classified as correct responses by the evaluator ignoring the correct answer. These findings raise doubt about the reliability of these evaluations, as different models give completely different results.

Moreover, as shown in Table~\ref{tab:o_qa_diff_evals} open-ended QA does not solve the main issues of MCQA. The performance of text-only models is surprisingly strong; state-of-the-art LLMs can guess the answer solely from the question text for a significant number of questions, even without a candidate list. This includes questions such as \textit{Which hand of the person in black wears a watch?} or \textit{What color is the pants of a person wearing black clothes?}, which correct answers are \textit{Left hand} and \textit{Black}. The first question can be answered just with prior knowledge as people commonly wear the watch on the left hand, while in the second one, the question gives too much information, the person is wearing black clothes. 
Similar to the findings for MCQA on spatial bias in Sec.~\ref{does_time_matter}, when using a single random frame for image-text models such as GPT-4o, performance reaches 60.6\% and 46.4\%, approaching the video-language model's 80.3\% and 61.6\%, respectively. In addition, the performance of Tarsier-34B does not significantly drop—on average less than 3\%—when the input videos are shuffled, indicating the low temporal understanding required for solving the benchmarks. These results show that open-ended video-language benchmarks also exhibit strong spatial bias, not requiring temporal understanding to be solved.

\begin{table}[t]   
    \small
    \centering
    \setlength{\tabcolsep}{0.75mm} 
    \resizebox{\columnwidth}{!}{ 
    \begin{tabular}{>{\centering\arraybackslash}p{3mm} p{16mm} p{10mm} p{8mm} p{8mm} p{8mm} p{8mm} p{8mm}}
        \toprule
         & & & \multicolumn{5}{c}{\textbf{Evaluation method}}\\
         \cmidrule{4-8} 
        & \multirow{2}{*}{\textbf{Model}} & \multirow{2}{*}{\textbf{Input}} & \multicolumn{2}{c}{\textbf{GPT-3.5}} & \multicolumn{2}{c}{\textbf{Llama3-70B}} & \multirow{2}{*}{\textbf{$\Delta$Acc.}} \\
         & & &  \textbf{Acc.} & \textbf{Score} & \textbf{Acc.} & \textbf{Score} \\
         \midrule
         \multirow{5}{*}{\rotatebox{90}{\textbf{MSVD}}} 
         & Llama3 70B & text & 29.1 & 2.6 & 51.9 & 2.8 & +22.8 \\
         & GPT-4o & text & 29.0 & 2.5 & 44.2 & 2.4 & +15.2\\
         & GPT-4o & image & 60.6 & 3.6 & 75.3 & 3.8 & +14.7 \\
         & Tarsier-34B & shuffle & 76.7 & 4.0 & 78.3 & 4.1 & +1.6 \\
         & Tarsier-34B & video & 80.3 & 4.2 & 78.3 & 4.1 & -2.0\\
         \midrule
         \multirow{5}{*}{\rotatebox{90}{\textbf{MSRVTT}}} 
         & Lama3 70B & text & 23.9 & 2.4 & 47.8 & 2.6 & +23.9 \\
         & GPT-4o & text & 23.3 & 2.3 & 42.4 & 2.3 & +19.1\\
         & GPT-4o & image & 34.2 & 2.7 & 50.8 & 2.7 & +16.6\\
         & Tarsier-34B & shuffle & 63.1 & 3.5 & 62.7 & 3.4 & -0.4\\
         & Tarsier-34B & video & 66.4 & 3.7 & 63.0 & 3.4 & -3.4\\
         \midrule
         \multirow{5}{*}{\rotatebox{90}{\textbf{ActivityNet}}} 
         & Lama3 70B & text & 25.3 & 2.6 & 32.6 & 1.9 & +7.3 \\
         & GPT-4o & text & 27.1 & 2.5 & 33.9 & 1.8 & +6.8\\
         & GPT-4o & image & 46.4 & 3.2 & 56.2 & 2.9 & +9.8\\
         & Tarsier-34B & shuffle & 59.9 & 3.6 & 60.8 & 3.4 & +0.9\\
         & Tarsier-34B & video & 61.6 & 3.7 & 61.3 & 3.4 & -0.3\\
         \midrule
         \multicolumn{8}{c}{\shortstack{Standard deviation between GPT-3.5 and Llama3\\score differences: $\pm$ 9.3}} \\
         \bottomrule
    \end{tabular}
    }
        \vspace{-0.2cm}
\caption{\textbf{Unreliability and biases of open-ended video-language benchmark evaluation.} Different LLMs used for evaluation produce varying results, see $\Delta$ column. Additionally, open-ended benchmarks also exhibit spatial and textual bias, similar to MCQA.}
    \label{tab:o_qa_diff_evals}

\vspace{-0.6cm}
\end{table}

\begin{figure*}
    \centering
    \includegraphics[width=\linewidth]{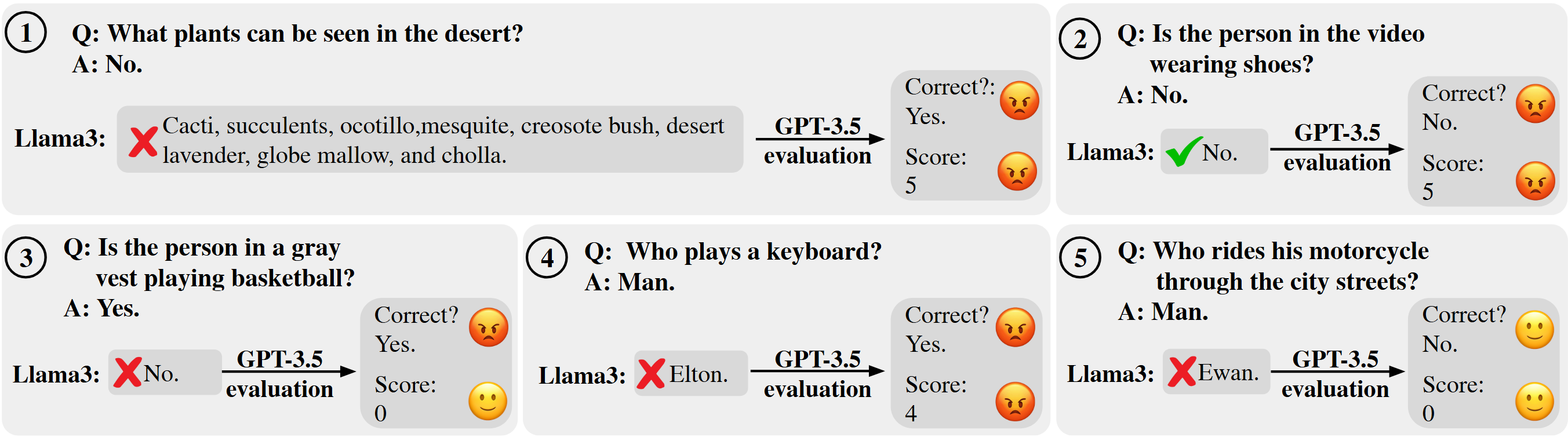}
    \caption{\textbf{Unreliability of open-ended benchmarks.} GPT 3.5 is commonly used as an evaluator; here, we use Llama 3 in a text-only setting to generate answers. GPT gives confusing accuracies and scores. Smiley emoji shows truthful or unreliable evaluation from GPT 3.5.}
    \label{fig:qualitative_oeqa}
    \vspace{-0.6cm}
\end{figure*}

Fig.~\ref{fig:qualitative_oeqa} shows incorrect evaluations on ActivityNet QA and MSVD QA. In these experiments, we prompt a Llama3 model to answer the questions in a text-only setting and perform the evaluation with GPT-3.5. In the first question, it can be seen how the model provides an answer completely unrelated to the video: ``Cacti, succulents, ocotillo, mesquite, creosote bush, ...,'' offering general information about desert plants. This is expected as the model only receives the question without visual information. The evaluation model classifies this question as correct with the highest matching score. It seems that the correct answer is disregarded in the evaluation of this example, focusing on the relationship between the questions and the predicted answer. Questions 2 and 3 contain two instances of no/yes QA pairs where the model correctly identifies whether they are correct, but assigns completely contradictory scores.
Questions 4 and 5 contain evaluations in which the correct answer is \textit{man}. However, it can be seen how the text-only model answers with specific names rather than a more general concept. The evaluations are completely different, most probably due to a context bias, as both Llama3 and the evaluation model --GPT3.5-- links the concept \textit{keyboard} with \textit{Elton}. 

In summary, current open-ended benchmarks are unreliable due to their use of LLMs as evaluators. This makes them unsuited for evaluating video-language models, especially as they also suffer from spatial and textual bias. In addition, they rely on closed-source LLMs for evaluation, which incurs costs to access, and becomes unreproducible when newer versions are released. 

\vspace{-0cm}
\vspace{-0.3cm}
\section{TVBench: A temporal VQA Benchmark}
\vspace{-0.2cm}


We propose TVBench, a new benchmark for evaluating temporal understanding in video QA. 
We adopt a multiple-choice QA approach to prevent the problems of open-ended VQA  described in Sec.~\ref{sec:oeqa}. The main design principles of TVBench are derived from and address the problems listed in Sec.~\ref{sec:mcqa}. 
Appendix~\ref{sec:details_TVBENCH}
provides an overview of the tasks, questions, and answers candidates used in our benchmark. We verify our choice of tasks and QA templates in Sec.~\ref{tvbench:evaluation} by the performance of multi-modal LLMs with solely a random frame or shuffled and reversed videos. 
\subsection{Designing TVBench}

This section explains the key strategies implemented in TVBench to address the issues identified in Sec.~\ref{sec:mcqa} of current video MCQA evaluation benchmarks.



\noindent \textbf{Strategy 1: Define Temporally Hard Answer Candidates.}
To address Problem \ref{theo:problem1}, it is crucial that the temporal constraints in the question are essential for determining the correct answer. This involves designing time-sensitive questions and selecting temporal challenging answer candidates.

\begin{figure}[t]
    \centering
    \vspace{-0.5em}
    \includegraphics[width=1\linewidth]{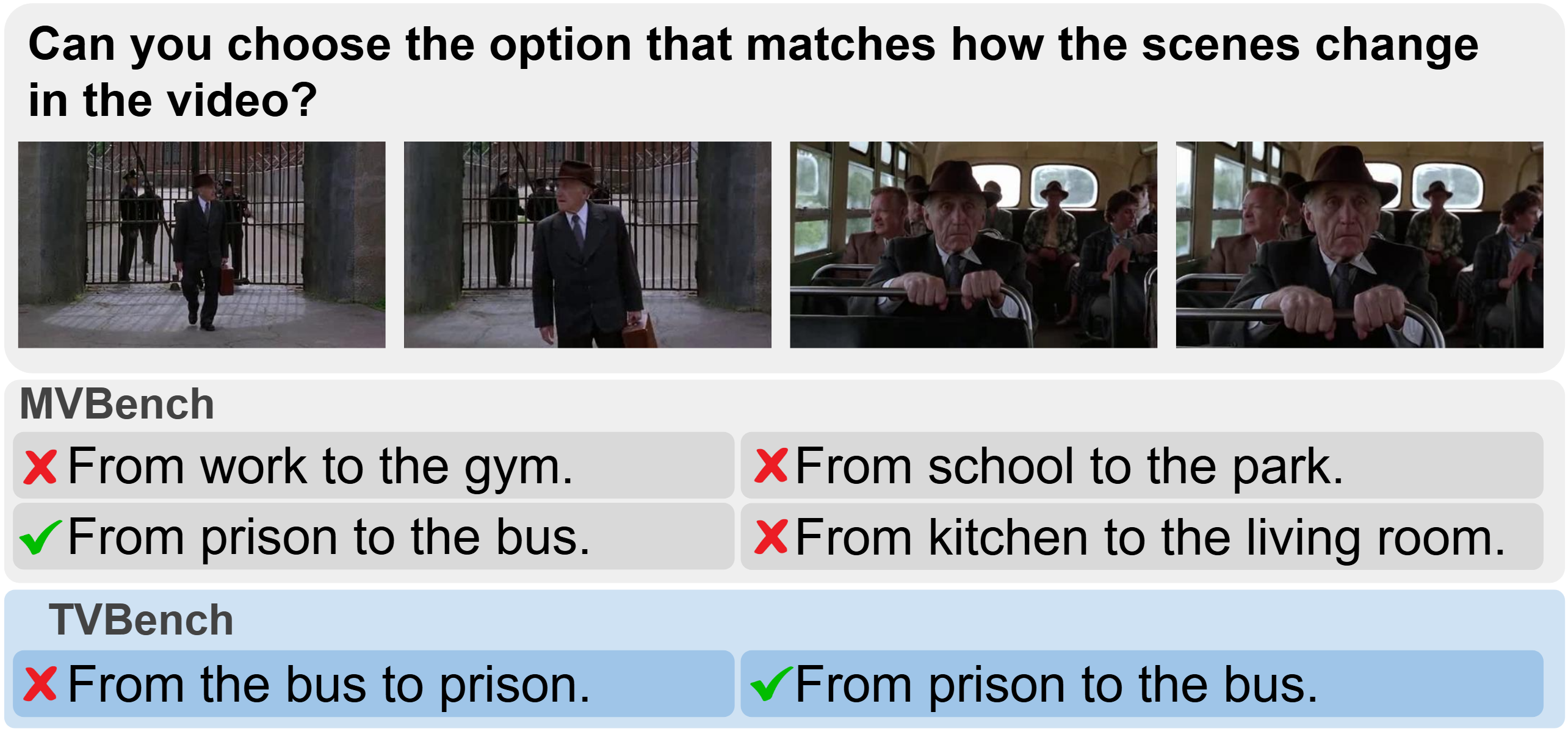}
    \vspace{-2em}
    \caption{\textbf{Strategy 1 of TVBench: Define temporally hard answer candidates.} To address the spatial bias in MVBench, we design answer candidates that are temporally challenging.}
    \label{fig:scene_transition}
    \vspace{-0.6cm}
\end{figure}

\begin{enumerate}[left=0pt, labelindent=0pt, itemindent=0pt, labelwidth=0pt, align=parleft]

    \item We select 10 temporally challenging tasks that require: repetition counting (Action Count), properties of moving objects (Object Shuffle, Object Count, Moving Direction), temporal localization (Action Localization, Unexpected Action), sequential ordering (Action Sequence, Scene Transition, Egocentric Sequence), and distinguishing between similar actions (Action Antonyms).
    \item We define hard-answer candidates based on the original annotations to ensure realism and relevance, rather than relying on LLM-generated candidates that are often random and easily disregarded, as seen in MVBench. For example, in the Scene Transition task (Fig.~\ref{fig:scene_transition}), we design a QA template that provides candidates based on the two scenes occurring in the videos for this task, rather than implausible options like “From work to the gym.” Similarly, for the Action Sequence task, we include only two answer candidates corresponding to the actions that occurred in the video. More details for the remaining tasks are in Appendix \ref{sec:details_TVBENCH}.
\end{enumerate}


\noindent \textbf{Strategy 2: Define QA pairs that are not overly informative.} Contrary to LLM-based generation, we apply basic templates to mitigate the effect of text-biased QA pairs, mitigating Problem \ref{theo:problem2}.

\begin{figure}[t]
    \centering
    \includegraphics[width=1\linewidth]{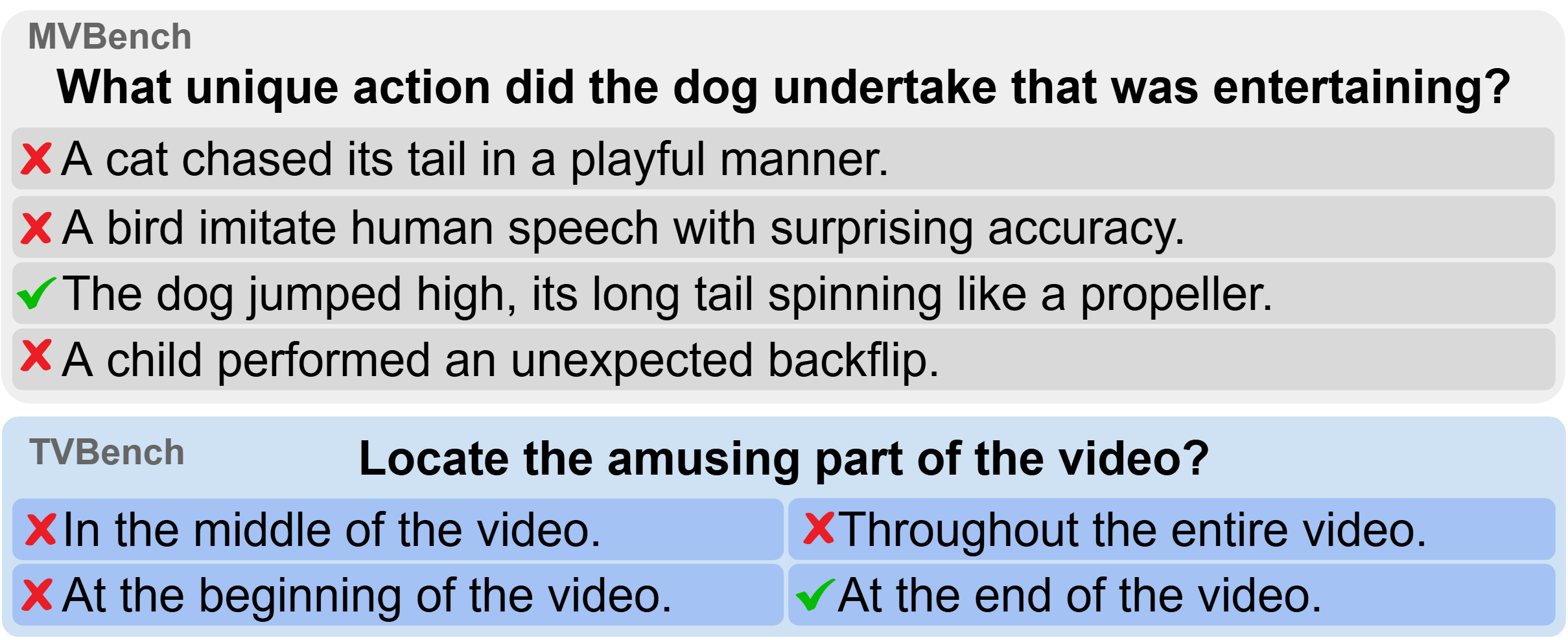}
    
    \caption{\textbf{Strategy 2 of TVBench: Define questions that are not overly informative.} To address the textual bias in MVBench, we design QA templates that do not contain unnecessary information.}
    \label{fig:strategy2}
    \vspace{-0.6cm}
\end{figure}

\begin{enumerate}[left=0pt, labelindent=0pt, itemindent=0pt, labelwidth=0pt, align=parleft]
    \item We design QA pairs that are concise and not unnecessarily informative by applying task-specific templates. These templates ensure that the QA pairs lack sufficient information to determine the correct answer purely from text. 
    An example of Unexpected Action is illustrated in Fig.~\ref{fig:strategy2}. QA pairs require the same level of understanding for the model to identify what is amusing in the video, but without providing additional textual information. Unlike MVBench, the model cannot simply select the only plausible option containing a dog.
    We use the same candidate sets across tasks like Action Count, Object Count, Object Shuffle, Action Localization, Unexpected Action, and Moving Direction, to ensure balanced datasets with an equal distribution of correct answers, keeping visual complexity while reducing textual bias.
    Appendix Table~\ref{tab:t_mvbench_samples} provides an overview of all tasks, demonstrating that the QA templates are carefully crafted without unnecessary textual information. 

    
    \item Solving the overreliance on world knowledge requires providing questions and candidates that contain only the necessary information, specifically removing factual information that the LLM can exploit. 
    We remove tasks such as Episodic Reasoning, that are based on QA pairs about TV shows or movies. 
\end{enumerate}



\vspace{-0.2cm}
\subsection{Dataset Source}
Videos in TVBench are sourced from Perception Test \citep{patraucean2024perception}, CLEVRER \citep{yi2019clevrer}, STAR \citep{wu2024star}, MoVQA \citep{zhang2023movqa}, Charades-STA \citep{gao2017tall}, NTU RGB+D \citep{liu2019ntu}, FunQA \citep{xie2023funqa} and CSV \citep{qian2022svip}. Overall, TVBench comprises first and third-person perspectives, indoor and outdoor scenes, and real and synthetic data with 2,654 QA pairs among 10 different tasks.
QA pairs are generated based on original annotations following the model provided in Appendix~\ref{sec:details_TVBENCH} for each task.

\vspace{-0.2cm}
\subsection{TVBench Evaluation}
\label{tvbench:evaluation}
Table~\ref{tab:sota} provides a detailed performance breakdown of state-of-the-art text~\citep{dubey2024llama3herdmodels} and multi-modal LLMs~\citep{internlmxcomposer2_5,li2024aria,zhang2024video,openai2024gpt4technicalreport, maaz2024videogpt+, li2024mvbench, xu2024pllava, geminiteam2024gemini15unlockingmultimodal, wang2024tarsierrecipestrainingevaluating, lin2023videollava, cheng2024videollama2advancingspatialtemporal,ye2024mplugowl3longimagesequenceunderstanding, wang2024qwen2vlenhancingvisionlanguagemodels, liu2024stllmlargelanguagemodels, su2023pandagptmodelinstructionfollow} across the 10 TVBench tasks.
In addition, we also report a human baseline to verify the quality of the benchmark, more details see appendix \ref{sec:human_baseline}. We also include the average performance of these models on MVBench and TVBench, with the upward arrow $_{\uparrow}$ indicating the improvement over random chance, and a human baseline to verify our benchmark's solveability.

\noindent \textbf{Does time matter?} For TVBench, multi-modal LLMs with a single image perform at random chance, verifying that a random frame is not sufficient for accurate question answering. Specifically, Gemini 1.5 Pro, the top image-language model on TVBench, outperforms random chance by only 3.0\%, compared to a 21.2\% improvement on MVBench.
Shuffling videos has minimal impact on the performance of video-language models on MVBench, but significantly degrades their accuracy on TVBench, where it drops to near-random levels. For example, Tarsier-34B's accuracy is 33.9\% higher than the random baseline on MVBench when videos are shuffled, while on TVBench, it is only 4.7\% higher under the same conditions.
This suggests that temporal understanding is crucial for TVBench, where visual data alone is insufficient to outperform random chance, unlike MVBench.
In addition, we analyze the effect of reversing videos instead of shuffling them. This significantly degrades model performance on TVBench, resulting in accuracy below the random baseline for even the best models. For instance, Tarsier-7B and Tarsier-34B perform worse than random by 6.1\% and 4.9\%, respectively. This is expected, as top models on TVBench rely on temporal information. Reversing frames misleads them with incorrect cues while shuffling disrupts the temporal structure entirely. These results demonstrate that temporal understanding is crucial for TVBench which is in contrast to MVBench where reversed and non-reversed videos obtain the same performance.

\noindent \textbf{Does vision matter?} 
For TVBench, state-of-the-art LLMs with text-only perform at random levels, highlighting the effectiveness of our Strategy 2 for Problem 2. Notably, Llama 3 achieves the best performance, just 1.4\% above random chance on TVBench, whereas it performs 10.8\% better on MVBench.
This indicates that LLMs cannot determine the answer solely by analyzing the question and answer candidates or by relying on prior world knowledge. Thus, visual information becomes key for solving TVBench.

\clearpage\onecolumn
\begin{table*}[t]
\vspace{-0.5em}
\footnotesize
\centering
\setlength{\tabcolsep}{5pt}
\begin{tabular}{lcccccccccccccc}
\toprule
& & \multirow{2}{*}{ \cellcolor{white} \textbf{\textbf{MVBench}}} & \multirow{2}{*}{\cellcolor{white} \textbf{\textbf{TVBench}}} & & \multicolumn{9}{c}{\textbf{TVBench}} \\
\cmidrule{5-14}
\textbf{Model} & \textbf{Input} & \cellcolor{white} Average & \cellcolor{white}Average & \textbf{AC} & \textbf{OC} & \textbf{AS} & \textbf{OS} & \textbf{ST} & \textbf{AL} & \textbf{AA} & \textbf{UA} & \textbf{ES} & \textbf{MD} \\ \midrule
Random & -- & 27.3& 33.3 & 25.0     & 25.0     & 50.0 & 33.3       & 50.0 & 25.0    & 50.0 & 25.0   & 25.0     & 25.0   \\
\midrule
GPT-3.5 Turbo &  \multirow{5}{*}{\shortstack{text- \\ only}} & 35.0$_{\uparrow7.7}$ & 33.1$_{\downarrow0.2}$ & 27.2     & 18.2     & 44.9 & 32.0       & 53.5 & 26.9    & 45.9 & 29.2   & 26.5     & 26.7   \\
Llama 3 70B & & 38.1$_{\uparrow10.8}$& 34.7$_{\uparrow1.4}$ & 30.2     & 27.0     & 48.7 & 32.9       & 55.1 & 26.9    & 49.1 & 23.3   & 25.5     & 28.0   \\
GPT-4o  & & 34.8$_{\uparrow7.5}$& 33.8$_{\uparrow0.5}$ & 28.0     & 21.0     & 48.7 & 33.3       & 53.5 & 25.6    & 50.9 & 25.8   & 28.5     & 22.4   \\

Gemini 1.5 Pro &  & 38.2$_{\uparrow 10.9}$& 33.6$_{\uparrow 0.3}$ & 25.0 & 17.6 & 53.0 & 33.3 & 51.9 & 25.0 & 54.7 & 23.3 & 28.0 & 24.1 \\
Tarsier-34B &  & 35.7$_{\uparrow 8.4}$ & 34.4$_{\uparrow 1.1}$ & 28.7 & 25.0 & 50.9 & 33.3 & 49.7 & 26.2 & 54.7 & 22.5 & 23.5 & 29.7\\
\midrule
Idefics3  & \multirow{4}{*}{image} & 44.2$_{\uparrow16.9}$& 34.5$_{\uparrow1.2}$ & 27.1     & 23.0     & 56.8 & 36.9       & 49.2 & 25.6    & 52.2 & 29.2   & 25.5     & 19.8   \\
GPT-4o & & 47.8$_{\uparrow20.5}$& 35.8$_{\uparrow2.5}$ & 30.8     & 19.6     & 62.1 & 33.3       & 52.4 & 25.0    & 52.5 & 27.5   & 33.0     & 21.6   \\
Gemini 1.5 Pro & & 48.5$_{\uparrow21.2}$ & 36.3$_{\uparrow3.0}$ & 22.8 & 21.0 & 55.9 & 36.4 & 51.4 & 36.9 & 54.7 & 35.8 & 30.0 & 23.7\\
Tarsier-34B &  & 45.1$_{\uparrow17.8}$ & 35.0$_{\uparrow1.7}$ & 30.0 & 26.4 & 61.0 & 27.1 & 53.0 & 32.5 & 49.4 & 27.5 & 21.5 & 22.0 \\
\midrule
VideoChat2
& \multirow{8}{*}{\shortstack{video \\ reverse}}
& 46.7$_{\uparrow19.4}$ & 32.1$_{\downarrow1.2}$ & 28.2 & 22.3 & 50.6 & 33.3 & 38.4 & 23.1 & 45.6 & 33.3 & 23.0 & 22.8\\
ST-LLM&  & 53.4$_{\uparrow26.1}$& 34.7$_{\uparrow1.4}$ & 25.4     & 32.4     & 55.1 & 36.9       & 43.6 & 30.5    & 47.5 & 31.4   & 23.5     & 20.3   \\
PLLaVA-7B&  & 45.8$_{\uparrow18.5}$& 34.1$_{\uparrow0.8}$ & 32.6     & 29.1     & 53.4 & 33.3       & 48.6 & 24.4    & 48.1 & 28.3   & 22.0     & 21.6   \\
PLLaVA-13B&  & 48.4$_{\uparrow21.1}$& 34.6$_{\uparrow1.3}$ & 36.9     & 27.7     & 51.9 & 33.3       & 52.4 & 25.0    & 46.6 & 34.2   & 19.5     & 19.0   \\
PLLaVA-34B&  & 56.2$_{\uparrow28.9}$& 33.4$_{\uparrow0.1}$ & 26.1 & 27.7 & 51.7 & 35.1 & 33.0 & 26.9 & 54.1 & 29.2 & 26.5 & 23.7\\
Gemini 1.5 Pro &  & 53.1$_{\uparrow25.8}$  & 27.0$_{\downarrow6.3}$ & 29.6 & 16.9 & 45.1 & 32.7 & 24.7 & 22.7 & 26.9 & 39.8 & 23.5 & 7.7\\
Tarsier-7B&  & 62.5$_{\uparrow35.2}$& 28.4$_{\downarrow4.9}$ & 24.3     & 19.6     & 41.1 & 36.0       & 38.9 & 24.4    & 29.4 & 30.0   & 24.5     & 15.9   \\
Tarsier-34B&  & 67.7$_{\uparrow40.4}$& 27.2$_{\downarrow6.1}$ & 30.2     & 25.0     & 45.1 & 32.9       & 31.9 & 21.9    & 19.1 & 38.3   & 23.0     & 4.3    \\

\midrule

VideoChat2&\multirow{8}{*}{\shortstack{video \\ shuffle}}
 & 49.8$_{\uparrow22.5}$ & 34.7$_{\uparrow1.4}$ & 25.6 & 27.0 & 54.0 & 32.9 & 56.2 & 23.1 & 48.1 & 33.3 & 24.5 & 22.4 \\

ST-LLM&   & 53.9$_{\uparrow26.6}$ & 35.0$_{\uparrow1.7}$ & 25.6 & 35.1 & 50.8 & 36.9 & 49.2 & 31.5 & 45.6 & 32.4 & 23.5 & 19.4 \\

PLLaVA-7B&   & 46.5$_{\uparrow19.2}$ & 34.4$_{\uparrow1.1}$ & 34.0 & 29.1 & 51.5 & 33.3 & 49.7 & 24.4 & 49.7 & 28.3 & 22.5 & 21.6\\

PLLaVA-13B&   & 49.5$_{\uparrow22.2}$ & 35.1$_{\uparrow1.8}$ & 36.8 & 25.0 & 55.5 & 33.3 & 51.9 & 27.5 & 45.9 & 35.0 & 21.0 & 19.0\\

PLLaVA-34B&   & 56.7$_{\uparrow29.4}$ & 37.2$_{\uparrow3.9}$ & 25.9 & 29.1 & 58.5 & 35.1 & 56.2 & 34.4 & 55.9 & 28.3 & 25.0 & 23.7\\

Gemini 1.5 Pro & & 56.8$_{\uparrow29.5}$ & 36.1$_{\uparrow2.8}$ & 26.5 & 23.7 & 55.9 & 35.1 & 51.4 & 31.3 & 51.9 & 31.7 & 29 & 24.6 \\
Tarsier-7B&   & 56.9$_{\uparrow29.6}$ & 36.0$_{\uparrow2.7}$ & 21.1 & 38.5 & 54.5 & 38.2 & 53.5 & 30.0 & 55.6 & 27.5 & 24.0 & 16.8\\

Tarsier-34B&   & 61.2$_{\uparrow33.9}$ & 38.0$_{\uparrow4.7}$ & 30.2 & 35.8 & 59.8 & 34.7 & 52.4 & 35.6 & 55.6 & 35.0 & 23.0 & 18.1\\
\midrule

VideoLLaVA & & 42.5$_{\uparrow15.2}$ & 33.8$_{\uparrow0.5}$ & 26.3 & 27.0 & 54.7 & 33.8 & 57.8 & 23.1 & 45.7 & 24.1 & 22.5 & 23.3\\
{VideoChat2} &  \multirow{22}{*}{video}  & 51.0$_{\uparrow23.7}$& 35.0$_{\downarrow1.7}$ & 25.9 & 27.0 & 56.8 & 32.9 & 56.2 & 23.1 & 48.1 & 32.9 & 24.5 & 22.4  \\

ST-LLM &  & 54.9$_{\uparrow27.6}$& 35.7$_{\uparrow2.4}$ & 25.0     & 35.1     & 51.7 & 36.0       & 54.4 & 31.0    & 45.6 & 34.2   & 24.0     & 20.3   \\

GPT-4o &  & 49.1$_{\uparrow21.8}$& 39.9$_{\uparrow6.6}$ & 26.1 & 21.3 & 59.3 & 33.2 & 52.4 & 25.0 & 78.4 & 41.7 & 31.0 & 30.6\\

{PLLaVA-7B}&  & 46.6$_{\uparrow19.3}$& 34.9$_{\uparrow0.9}$ & 32.1     & 25.7     & 55.6 & 33.3       & 52.4 & 23.8    & 53.1 & 30.5   & 20.5     & 21.6   \\

PLLaVA-13B &  & 50.1$_{\uparrow22.8}$& 36.4$_{\uparrow3.1}$ & 37.3     & 24.3     & 61.8 & 33.3       & 55.1 & 28.1 & 47.8 & 39.0   & 19.5     & 17.2   \\

PLLaVA-34B  &  & 58.1$_{\uparrow30.8}$& 42.3$_{\uparrow9.0}$ & 27.6     & 32.4     & 67.0 & 35.6       & 77.8 & 44.4    & 58.8 & 32.9   & 27.0     & 19.4   \\

mPLUG-Owl3 & & 54.5$_{\uparrow27.2}$ & 42.2$_{\uparrow8.9}$ & 27.4 & 32.4 & 69.8 & 37.3 & 76.8 & 43.1 & 56.9 & 34.2 & 18.0 & 26.3\\

VideoLLaMA2.1 & & 57.3$_{\uparrow30.0}$ & 42.1$_{\uparrow8.8}$ & 25.4 & 30.4 & 71.2 & 29.8 & 76.6 & 46.9 & 58.1 & 36.8 & 23.5 & 22.4\\
VideoLLaMA2 7B & & 54.6$_{\uparrow27.3}$ & 42.9$_{\uparrow9.6}$ & 30.6 & 37.8 & 69.6 & 33.8 & 68.1 & 40.6 & 56.9 & 43.9 & 24.5 & 22.8\\
VideoLLaMA2 72B & & 62.0$_{\uparrow34.7}$ & 48.4$_{\uparrow15.1}$ & 26.7 & 46.6 & 73.5 & 40.4 & 81.6 & 53.1 & 76.6 & 36.6 & 19.5 & 29.7\\

VideoGPT+ &  & 58.7$_{\uparrow31.4}$& 41.7$_{\uparrow8.4}$ & 30.6     & 52.0     & 61.3 & 36.9       & 69.2 & 40.0    & 53.1 & 29.7   & 16.0     & 28.5   \\
Gemini 1.5 Pro &  & 60.5$_{\uparrow33.2}$& 47.6$_{\uparrow14.3}$ & 33.5     & 22.3     & 71.5 & 34.5       & 82.6 & 51.6    & 77.8 & 43.9   & 25.0     & 33.6   \\
Qwen2-VL 7B & & 67.0$_{\uparrow39.7}$ & 43.8$_{\uparrow10.5}$ & 27.1 & 61.5 & 63.8 & 38.7 & 75.7 & 41.3 & 63.1 & 22.0 & 22.5 & 22.4\\
Qwen2-VL 72B & & 73.6$_{\uparrow46.3}$ & 52.7$_{\uparrow19.4}$ & 32.6 & 73.0 & 68.6 & 31.1 & 82.2 & 45.0 & 75.3 & 37.8 & 19.5 & 62.1\\
LLaVA-Video 7B & & 58.6$_{\uparrow31.3}$ & 45.6$_{\uparrow12.3}$ & 32.6 & 23.6 & 78.5 & 32.9 & 81.6 & 58.8 & 71.9 & 29.3 & 22.5 & 24.1 \\
LLaVA-Video 72B & & 64.1$_{\uparrow36.8}$ & 50.0$_{\uparrow16.7}$ & 38.6 & 27.0 & 80.3 & 33.3 & 85.9 & 65.6 & 66.6 & 35.4 & 25.5 & 41.8 \\
IXC-2.5-7B & & 69.1$_{\uparrow41.8}$ & 51.6$_{\uparrow18.3}$ & 32.1 & 50.7 & 77.1 & 37.3 & 78.4 & 46.9 & 60.0 & 41.7 & 21.0 & 70.3 \\
Aria & & 69.7$_{\uparrow42.4}$ & 51.0$_{\uparrow17.7}$ & 58.4 & 60.1 & 75.1 & 42.2 & 81.6 & 43.1 & 70.3 & 32.9 & 21.0 & 25.0 \\
Tarsier-7B &  & 62.6$_{\uparrow35.3}$& 46.9$_{\uparrow13.6}$ & 22.9     & 58.8     & 73.2 & 35.6       & 64.9 & 46.9    & 75.6 & 40.2   & 25.5     & 25.0   \\
Tarsier-34B &  & 67.6$_{\uparrow40.3}$& 55.5$_{\uparrow22.2}$ & 31.7     & 64.2     & 81.7 & 32.1       & 77.8 & 58.1    & 84.4 & 52.4   & 24.5     & 48.3   \\ 
Human Baseline &  & - & 94.8$_{\uparrow61.5}$  & 100.0  & 94.9  & 100.0  & 90.6  & 90.0  & 96.0  & 100.0  & 86.0  & 90.0  & 100.0 \\
\bottomrule
\end{tabular}
\vspace{-0.2cm}
\caption{\textbf{Results on TVBench.} On TVBench, text-only and image models perform near-random. Surprisingly, several recent state-of-the-art video-language models, such as ST-LLM, also perform close to random. With TVBench we can identify temporally strong models like Gemini and Tarsier. In addition, these models drop significantly in accuracy when the videos are shuffled or reversed, indicated by the upward arrow $_{\uparrow}$ showing the difference to random chance.}
\label{tab:sota}
\vspace{-0.5cm}
\end{table*}

\clearpage
\thispagestyle{plain}
\begingroup
\newsavebox{\expandedtablebox}\begin{lrbox}{\expandedtablebox}
\begin{minipage}{\linewidth}
\captionsetup{font=footnotesize,skip=3pt}
\captionof{table}{\textbf{Expanded TVBench evaluation.} Macro averages the ten task accuracies; micro averages questions. Our paired scores cover the same 2,525 questions. External reports are included only when their aggregation is established. Model citations identify origins; Source identifies our evaluation or the reporting paper. Settings and historical dataset versions differ.}\label{tab:sota-update}
\fontsize{9}{10}\selectfont\setlength{\tabcolsep}{3pt}\renewcommand{\arraystretch}{1.02}
\setlength{\aboverulesep}{0.4pt}\setlength{\belowrulesep}{0.4pt}
\begin{tabularx}{\linewidth}{@{}>{\raggedright\arraybackslash}p{0.35\linewidth}>{\raggedright\arraybackslash}Xrrl@{}}
\toprule Model / checkpoint & Model setting & Macro (\%) & Micro (\%) & Source\\\midrule
Gemini 4 Carbon~\citep{vfour_gemini4_carbon} & 2 fps; HIGH media; Thinking (medium) & 87.35 & 88.12 & Ours\\
Gemini 4 Argon~\citep{vfour_gemini4_argon} & 2 fps; HIGH media; Thinking (medium) & 86.82 & 87.56 & Ours\\
Gemini 3.8 Flash~\citep{vfour_gemini38_flash} & 2 fps; HIGH media; Thinking (medium) & 82.59 & 82.30 & Ours\\
Gemini 3.5 Flash-Lite~\citep{vfour_gemini35_flash_lite} & 2 fps; HIGH media; Thinking (medium) & 66.81 & 68.20 & Ours\\
\cmidrule(lr){1-5}
GPT-6.1 Sol~\citep{vfour_openai_sol_api} & 2 fps; original images; Thinking (med.) & 86.71 & 88.63 & Ours\\
\cmidrule(lr){1-5}
Intern-S2-397B~\citep{vfour_model_b2b901ebf1} & 2 fps\(\leq\)512f; native; Thinking & 77.87 & 77.19 & Ours\\
\cmidrule(lr){1-5}
Qwen3.8-Flash-Next~\citep{vfour_model_de56e022f7} & 2 fps\(\leq\)512f; native; Direct & 64.74 & 64.20 & Ours\\
Qwen3.8-Flash-Next~\citep{vfour_model_de56e022f7} & 2 fps\(\leq\)512f; native; Thinking (xhigh) & 79.09 & 79.45 & Ours\\
Qwen3.8-27B~\citep{vfour_29578a8593} & 2 fps\(\leq\)512f; native; Direct & 67.52 & 67.13 & Ours\\
Qwen3.5-27B~\citep{vfour_model_554611c1c1} & 2 fps\(\leq\)512f; native; Direct & 67.48 & 67.13 & Ours\\
Qwen3.5-27B~\citep{vfour_model_554611c1c1} & 2 fps\(\leq\)512f; native; Thinking & 74.91 & 74.69 & Ours\\
Qwen3.5-397B-A17B~\citep{vfour_model_554611c1c1} & 2 fps\(\leq\)512f; native; Direct & 69.88 & 70.18 & Ours\\
Qwen3.5-122B-A10B~\citep{vfour_model_554611c1c1} & 2 fps\(\leq\)512f; native; Direct & 69.01 & 68.59 & Ours\\
Qwen3.5-35B-A3B~\citep{vfour_model_554611c1c1} & 2 fps\(\leq\)512f; native; Direct & 68.66 & 68.40 & Ours\\
Qwen3.5-4B~\citep{vfour_model_554611c1c1} & 2 fps\(\leq\)512f; native; Direct & 65.98 & 66.85 & Ours\\
Qwen3.5-9B~\citep{vfour_model_554611c1c1} & 2 fps\(\leq\)512f; native; Direct & 64.08 & 63.72 & Ours\\
Qwen3-VL-235B-A22B-Instruct~\citep{vfour_model_f2c33147d1} & 2 fps\(\leq\)512f; native; Direct & 72.51 & 73.23 & Ours\\
Qwen3-VL-235B-A22B-Thinking~\citep{vfour_model_f2c33147d1} & 2 fps\(\leq\)512f; native; Thinking & 69.93 & 70.46 & Ours\\
Qwen3-VL-8B~\citep{vfour_model_f2c33147d1} & 2 fps\(\leq\)512f; native; Direct & 61.86 & 61.98 & Ours\\
Qwen3-VL-4B~\citep{vfour_model_f2c33147d1} & 8 fps\(\leq\)2048f; \(\leq\)448²; 80K visual tokens & -- & 53.70 & \citep{vfour_6ad33070a4}\\
\cmidrule(lr){1-5}
GLM-4.6V~\citep{vfour_model_0b2768f023} & dynamic\(\leq\)640f; native; Direct & 67.02 & 67.56 & Ours\\
GLM-4.6V-Flash~\citep{vfour_model_0b2768f023} & dynamic\(\leq\)640f; native; Direct & 63.12 & 63.80 & Ours\\
GLM-4.6V-Flash~\citep{vfour_model_0b2768f023} & dynamic\(\leq\)640f; native; Thinking & 66.96 & 67.17 & Ours\\
\cmidrule(lr){1-5}
Molmo2-4B~\citep{vfour_model_17dceab39a} & 2 fps\(\leq\)384f; 378\(\times\)378/frame; Direct & 65.83 & 66.89 & Ours\\
Molmo2-8B~\citep{vfour_model_17dceab39a} & 2 fps\(\leq\)384f; 378\(\times\)378/frame; Direct & 64.20 & 65.82 & Ours\\
\cmidrule(lr){1-5}
Gemma4-31B~\citep{vfour_model_cc62f90cdf} & 1 fps\(\leq\)60f; native; Direct & 54.79 & 54.53 & Ours\\
Gemma4-31B~\citep{vfour_model_cc62f90cdf} & 1 fps\(\leq\)60f; native; Thinking & 65.61 & 65.03 & Ours\\
Gemma4-26B-A4B~\citep{vfour_model_cc62f90cdf} & 32f; native; Direct & 51.00 & 50.34 & Ours\\
\cmidrule(lr){1-5}
PLM-8B~\citep{vfour_c828841319} & 32f; native; Direct & 64.45 & 66.73 & Ours\\
PLM-3B~\citep{vfour_c828841319} & 32f; native; Direct & 57.44 & 60.48 & Ours\\
PLM-1B~\citep{vfour_c828841319} & 32f; native; Direct & 47.68 & 50.57 & Ours\\
\cmidrule(lr){1-5}
MiniCPM-V-4.7-35B-A3B~\citep{vfour_model_ffcc020a58} & 1 fps\(\leq\)128f; native; Direct & 62.02 & 62.65 & Ours\\
MiniCPM-V-4.5-8B~\citep{vfour_model_cfd3ba4c95} & 2 fps\(\leq\)540f; groups of 3 & -- & 53.00 & \citep{vfour_6ad33070a4}\\
\cmidrule(lr){1-5}
LongVU + RRPO~\citep{vfour_3696f97359} & 1 fps\(\leq\)100f; 384²+378² (code) & -- & 56.50 & \citep{vfour_3696f97359}\\
LongVU + DPO~\citep{vfour_3696f97359} & 1 fps\(\leq\)100f; 384²+378² (code) & -- & 54.30 & \citep{vfour_3696f97359}\\
LongVU~\citep{vfour_model_bada4b0ae5} & 1 fps\(\leq\)100f; 384²+378² (code) & -- & 53.70 & \citep{vfour_3696f97359}\\
\cmidrule(lr){1-5}
Nemotron-Nano-12B-v2-VL~\citep{vfour_model_f9db5082dd} & 2 fps\(\leq\)128f; 512\(\times\)512/frame; Direct; EVS50 & 53.71 & 54.30 & Ours\\
Nemotron-3.5-Super-VL-120B-A12B-NVFP4~\citep{vfour_model_5fb74237ce} & 1 fps\(\leq\)128f; native; Direct & 50.73 & 52.20 & Ours\\
\cmidrule(lr){1-5}
Tarsier2-7B~\citep{vfour_dbf5e7ad7d} & 16f; \(\leq\)460,800 px (code) & 52.61 & 54.70 & \citep{vfour_dbf5e7ad7d,vfour_152f4cc5c9}\\
Tarsier2-Recap-7B~\citep{vfour_152f4cc5c9} & -- & 50.38 & 54.00 & \citep{vfour_152f4cc5c9}\\
\cmidrule(lr){1-5}
LLaVA-Video-7B + RRPO (32f training variant)~\citep{vfour_3696f97359} & 64f; 384\(\times\)384 (code) & -- & 52.20 & \citep{vfour_3696f97359}\\
LLaVA-OneVision-1.5-4B~\citep{vfour_onevision15} & \(\leq\)32f uniform; 200,704 px/f; Direct & 41.77 & 43.17 & Ours\\
LLaVA-OneVision-1.5-8B~\citep{vfour_onevision15} & \(\leq\)32f uniform; 200,704 px/f; Direct & 41.73 & 42.89 & Ours\\
LLaVA-OneVision-1.5-8B-RL~\citep{vfour_onevision15} & \(\leq\)32f uniform; 200,704 px/f; Fast & 40.60 & 41.94 & Ours\\
LLaVA-OneVision-1.5-8B-RL~\citep{vfour_onevision15} & \(\leq\)32f uniform; 200,704 px/f; Thinking & 39.81 & 41.35 & Ours\\
\cmidrule(lr){1-5}
Eagle2.5-8B~\citep{vfour_model_be1b63daf2} & 2 FPS; \(\leq\)512 frames & -- & 62.00 & \citep{vfour_6ad33070a4}\\
InternVideo2.5-8B~\citep{vfour_model_7b7899c91b} & 128--512 uniform frames; full video & -- & 58.10 & \citep{vfour_6ad33070a4}\\
InternVL3.5-4B~\citep{vfour_model_85cd34f428} & 64 uniform frames & -- & 57.10 & \citep{vfour_6ad33070a4}\\
VideoChat3-4B~\citep{vfour_6ad33070a4} & 8 fps\(\leq\)2048f; \(\leq\)448²; 80K visual tokens & -- & 55.40 & \citep{vfour_6ad33070a4}\\
VideoLLaMA3-7B~\citep{vfour_model_76003ef63d} & 1 fps\(\leq\)180f; native; Direct & 53.90 & 54.61 & Ours\\
Video-R1~\citep{vfour_model_c3843796b0} & 2 fps\(\leq\)32f; \(\leq\)200,704 px/frame (author-confirmed); Thinking & 53.60 & 53.54 & \citep{vfour_1e7a4da468}\\
Video-XL2-8B~\citep{vfour_model_209b27ee7f} & 1 FPS; \(\leq\)512 frames & -- & 47.50 & \citep{vfour_6ad33070a4}\\
\bottomrule\end{tabularx}
\par\vspace{3pt}\scriptsize
f: frames; $\leq$: cap; native: model-specific spatial processing; EVS50: 50\% visual-token pruning. --: unavailable. Tarsier2 and Recap macro values are reconstructed from rounded task scores. (code): released defaults, not historical-run logs. RRPO micro follows its released scorer. Invalid answers count as wrong.
\end{minipage}\end{lrbox}
\typeout{TVBench expanded table: natural height=\the\dimexpr\ht\expandedtablebox+\dp\expandedtablebox\relax; page limit=\the\dimexpr\textheight-12pt\relax}
\noindent\makebox[\linewidth][c]{%
\ifdim\dimexpr\ht\expandedtablebox+\dp\expandedtablebox\relax>\dimexpr\textheight-12pt\relax
\resizebox*{!}{\dimexpr\textheight-12pt\relax}{\usebox{\expandedtablebox}}%
\else\usebox{\expandedtablebox}\fi}
\endgroup

\clearpage\twocolumn

\begingroup
\subsection{Expanded Model Evaluation (v4)}
\label{sec:v4-results}
Table~\ref{tab:sota-update} adds 52 completed settings for recent models,
including direct and thinking runs. Our evaluations use all 2,525 questions
and report task-macro accuracy (mean of ten task accuracies) and question-micro
accuracy (total correct divided by 2,525) from the same predictions.
Clear final choices are accepted; missing, ambiguous or truncated answers count
as incorrect. External scores retain their source citations and verified
aggregation, with reconstruction details noted below the table.

Gemini 4 Carbon, Gemini 4 Argon and GPT-6.1 Sol reach 87.35\%, 86.82\%
and 86.71\% macro accuracy, respectively. The strongest evaluated open-weight setting is
Qwen3.8-Flash-Next with thinking at 79.09\%. Thinking raises Qwen3.5-27B from
67.48\% to 74.91\%, while Qwen3-VL-235B-Thinking scores 69.93\% versus
72.51\% for its separately released Instruct checkpoint. Model-specific inputs,
generation budgets and historical dataset versions differ, so these are not
controlled comparisons of reasoning modes.
\par\endgroup

\vspace{-0.15cm}
\section{Discussion}
\label{sec:discussion}

\textbf{A sobering view on current models.}
With our new TVBench, we can accurately assess the temporal understanding of existing video-language models.
Surprisingly, we find that recent state-of-the-art and highly popular models, such as VideoChat2, ST-LLM, PLLava, VideGPT+, GPT-4o, VideoLLaMA2 7B, mPLUG-Owl3 perform close to random chance on our temporal benchmark, with the largest gain being only +8.9\%. Only five models, Qwen2-VL, LLaVA-Video, IXC-2.5, Aria, and Tarsier, achieve above 50\% accuracy, significantly outperforming the random baseline. From these results, we observe that TVBench amplifies the performance gaps between models with the strongest temporal understanding and those with weaker capabilities, as a good benchmark should.

\noindent \textbf{Conclusion.} In this work, we highlight major limitations in existing language-video benchmarks, particularly in the widely used MVBench and open-ended benchmarks. Key issues include inadequate temporal evaluation and tasks that do not require visual information, making tracking progress in this domain ineffective. To address these problems, we introduce TVBench, a benchmark designed to assess the temporal understanding of video-language models explicitly. Our experiments reveal that on TVBench, text-only and visual models lacking temporal reasoning perform at random levels, and only a handful of models achieve moderately high scores, showing the potential for progress. TVBench thus provides a reliable yardstick for evaluating future advancements in video-language models.

  \section*{Acknowledgement}

  We acknowledge funding from Spain's Agencia Estatal de Investigación
  (grant PID2023-149549NB-I00), the Xunta de Galicia through its
  Consellería de Educación, Ciencia, Universidades e Formación
  (Centro de Investigación de Galicia accreditation 2024--2027,
  ED431G-2023/04), and the European Union through the European Regional
  Development Fund (ERDF).
  We also acknowledge support from Qualcomm Technologies Inc., the
  University of Amsterdam, and the Dutch Ministry of Economic Affairs
  and Climate Policy through its Top Consortia for Knowledge and
  Innovation (TKI) allowance.

{\small
\bibliographystyle{ieeenat_fullname}
\bibliography{11_references,11_update_references,11_model_references}
}

\ifarxiv \clearpage 
\appendix \section{Appendix} 
\label{apendix:tvbench}

\subsection{Reproducibility Statement}
To ensure the reproducibility of our results, we will make the full dataset publicly accessible in raw form, along with comprehensive documentation detailing its structure and usage. Additionally, we will provide open-sourced evaluation code under a permissive license, enabling researchers to replicate our experiments and adapt the tools for their own studies.


\subsection{Details on TVBench}
\label{sec:details_TVBENCH}
\subsubsection{Benchmark Creation}
\label{sub:benchmark_creation}
Table~\ref{tab:t_mvbench_samples} provides the template used to generate the QA pairs for each task on TVBench. 
For the tasks \textit{Action Count}, \textit{Object Count}, \textit{Object Shuffle}, \textit{Action Localization}, \textit{Unexpected Action}, and \textit{Moving Direction}, we use the same set of options for every question. For \textit{Action Sequence} and Scene Transition, we generate candidates by selecting an action and a scene from the video to create negative candidates. The \textit{Action Antonym} candidate set consists of two similar actions that are difficult to distinguish without time understanding, such as 'Wear a jacket' and 'Take off a jacket.' Finally, for \textit{Egocentric Sequence}, we generate random modifications to the correct sequence of actions to create three negative candidates. Table~\ref{tab:t_mvbench_samples} also details the datasets from which videos for each task were sourced.
Here we give more details for each task and their template:

\textbf{Action Antonym.}
We source videos from the NTU \citep{liu2019ntu} RGB+D 120 (action classification dataset). Videos are filtered to retain only those belonging to one of 16 selected categories (e.g., take off bag, put on bag, putting something into a bag, take something out of a bag). For each QA, we select the correct option as the label and the alternative option as the opposite action (e.g., take off bag vs put on bag).

\textbf{Action Count.}
Questions are sourced from the Perception \citep{patraucean2024perception} dataset. QAs are filtered to always include the same candidate set with four options, each being correct for 25\% of the questions.

\textbf{Action Localization.}
Questions are sourced from MVBench\citep{li2024mvbench} The candidate set is balanced, with four options, each being correct 25\% of the time.

\textbf{Action Sequence.}
Videos are selected from the STAR \citep{wu2024star} dataset, containing two actions. The question is always: "What did the person do first?" Two candidates are provided, each representing one of the actions depicted in the video.

\textbf{Egocentric Sequence.}  
We ask for the correct order of tasks performed in egocentric videos taken from CSV \citep{qian2022svip}. The question is: "What is the sequence of actions shown in the video?" The correct order is taken from the original annotations. Three negative candidates are generated by swapping two tasks in the correct order.

\textbf{Moving Direction.}  
Videos and annotations are taken from CLEVRER\citep{yi2019clevrer}, including annotations for the positions and directions of each object in the video. The question is: "Which direction does the \{yellow sphere\} move in the video?" Candidates are always the same four options, each being correct for 25\% of the questions:  
1. Down and to the left.  
2. Down and to the right.  
3. Up and to the right.  
4. Up and to the left.

\textbf{Object Count.}  
Videos and annotations are taken from CLEVRER\citep{yi2019clevrer} The question is: "How many moving cylinders are there when the video begins?" The candidate set always consists of four options, each being correct for 25\% of the questions. QAs are selected so that ignoring the time constraint (e.g., "when the video begins") would lead to incorrect answers (e.g., the number of objects at the beginning is different from at the end).

\textbf{Object Shuffle.}  
Questions are sourced from Perception\citep{patraucean2024perception} QAs are filtered to always include the same candidate set with four options, each being correct for 25\% of the questions.

\textbf{Scene Transition.}  
Questions are sourced from MVBench\citep{li2024mvbench} (with videos originally from MoVQA). The original question and correct answer are retained, while incorrect options are discarded. A hard negative is generated by reversing the sequence in the correct answer.

\textbf{Unexpected Action.}  
Annotations are parsed from the FunQA \citep{xie2023funqa} dataset, which provides the start and end times for each action. The question is: "Locate the creative | mesmerizing | surprising part of the video." The candidate set always consists of the following four options:  
1. "Throughout the entire video."  
2. "At the beginning of the video."  
3. "At the end of the video."  
4. "In the middle of the video."  
The correct option is set based on the original annotations.

\begin{table*}[t]
    \centering
    \caption{Overview of all tasks and datasets used in our TVBench benchmark.}
    \label{tab:t_mvbench_samples}
    \small
    \scriptsize
    \setlength{\tabcolsep}{4pt}
    \begin{tabular}{llll}
        \toprule
        Task & Source & Question & Candidates\\
        \midrule
        \makecell[l]{Action\\Count} & Perception & \makecell[l]{The person makes sets of repeated actions.\\How many distinct repeated actions did the person do?} &2; 3; 4; 5\\
                \midrule
        \makecell[l]{Object\\Shuffle} & Perception & \makecell[l]{The person uses multiple similar objects to play an occlusion game.\\Where is the hidden object at the end of the game from the person's\\point of view?} & \makecell[l]{Under the third object from the left.\\Under the second object from the left.\\Under the first object from the left.}\\
        \midrule
        
        \makecell[l]{Object\\Count} & CLEVRER & How many \{metal\} objects are moving when \{the video begins\}? & 0; 1; 2; 3\\
                \midrule
        \makecell[l]{Moving\\Direction} & CLEVRER & Which direction does the \{gray cube\} move in the video? & \makecell[l]{Down and to the right.\\Down and to the left.\\Up and to the right.\\Up and to the left.}\\
        \midrule
        \makecell[l]{Action\\Sequence} & STAR & What did the person do first? & \makecell[l]{2 actions that actually happened\\on the video}\\

        \midrule
        \makecell[l]{Scene\\Transition} & MoVQA & What's the right option for how the scenes in the video change? & \makecell[l]{From \{scene1\} to \{scene2\}.\\From \{scene2\} to \{scene1\}.}\\ 
        \midrule
        \makecell[l]{Action\\Localization} & Charades & \makecell[l]{During which part of the video does\\the action \{person takes a blanket\} occur?} & \makecell[l]{Throughout the entire video.\\At the end of the video.\\In the middle of the video.\\At the beginning of the video.}\\
        \midrule
        \makecell[l]{Action\\Antonym} &  \makecell[l]{NTU\\RGB+D} & What is the action being performed in the video? & Wear jacket; Take off jacket\\
        \midrule
        \makecell[l]{Unexpected\\Action} & FunQA & Locate the \{creative$|$amusing$|$mesmerizing\} part of the video. & \makecell[l]{Throughout the entire video.\\At the end of the video.\\At the beginning of the video.\\In the middle of the video.}\\
        \midrule
        \makecell[l]{Egocentric\\Sequence} & CSV & What is the sequence of actions shown in the video? & \makecell[l]{GT + Correct sequence changing\\the order of actions} \\

        \bottomrule
    \end{tabular}
\end{table*}

\begin{table}[]
    \caption{\textbf{TVBench Video Statistics for each task.}}
    \centering
    \begin{tabular}{l|cc}
    \toprule
    Task & \#Videos & \makecell{Average\\Frame Length} \\
    \midrule
    Action Count & 536 & 628 \\
    Object Count & 148 & 127 \\
    Action Sequence & 437 & 409 \\
    Object Shuffle & 225 & 710 \\
    Scene Transition & 185 & 599 \\
    Action Localization & 160 & 423 \\
    Action Antonym & 320 & 171 \\
    Unexpected Action & 82 & 726 \\
    Egocentric Sequence & 200 & 562 \\
    Moving Direction & 232 & 127 \\
    \bottomrule
    \end{tabular}
    \label{tab:my_label}
\end{table}


\subsubsection{Human Baseline Study}
\label{sec:human_baseline}
With 14 annotators we label 400 videos of the TVBench out of the 2,654 videos in the benchmark. We achieved an average human accuracy of 94.8\%.

\textbf{Estimating Stastical Error:} By having each annotator annotate 40 different videos, resulting in 400 unique annotations across the dataset, we aim to estimate the human performance baseline with acceptable statistical precision. The total number of videos in the dataset is \( N = 2654 \), and the number of annotated videos is \( n = 400 \). Each video is annotated by a single annotator. In the absence of prior knowledge about the true average human accuracy \( p \), we adopt a conservative approach by assuming \( p = 0.5 \), which maximizes the variance \( p(1 - p) \). This assumption ensures that our estimation of the standard error is not underestimated, providing a robust margin of error. The estimated average accuracy \( \hat{p} \) is given by:
\[
\hat{p} = \frac{1}{n} \sum_{i=1}^{n} y_i,
\]
where \( y_i \) is an indicator variable that equals 1 if the annotator answered video \( i \) correctly, and 0 otherwise. The variance of \( \hat{p} \) with finite population correction is calculated as:
\[
\text{Var}(\hat{p}) = \frac{p(1 - p)}{n} \times \left( \frac{N - n}{N - 1} \right).
\]
Substituting \( p = 0.5 \), \( n = 400 \), and \( N = 2654 \), we have:
\[
p(1 - p) = 0.5 \times 0.5 = 0.25,
\]
\[
\frac{N - n}{N - 1} = \frac{2654 - 400}{2654 - 1} = \frac{2200}{2599} \approx 0.8496.
\]
Therefore, the variance becomes:
\[
\text{Var}(\hat{p}) = \frac{0.25}{400} \times 0.8496 = \frac{0.25 \times 0.8465}{400} = \frac{0.211625}{400} = 0.000531.
\]
The standard error (SE) is the square root of the variance:
\[
\text{SE} = \sqrt{\text{Var}(\hat{p})} = \sqrt{0.000531} \approx 0.0230.
\]
At a 95\% confidence level (with \( z = 1.96 \)), the margin of error (ME) is:
\[
\text{ME} = z \times \text{SE} = 1.96 \times 0.0230 \approx 0.0451.
\]
Thus, the statistical error in estimating human performance is characterized by a standard error of approximately 0.0230, leading to a margin of error of \( \pm 4.51\% \). This conservative estimate provides an upper bound on the margin of error, ensuring our results are statistically robust even without prior knowledge of \( p \).

\subsection{Extended Analysis of Video MCQA Benchmarks}
\label{app:nextqa}

We extend our analysis beyond MVBench to evaluate the NextQA benchmark \citep{xiao2021next}, using the best-performing model, Tarsier-34B. Specifically, we assess performance across five settings: text-only, image-only (random video frame), video, shuffled video, and reversed video. The results, summarized in Table~\ref{tab:tarsier_nextqa}, reveal significant spatial and textual biases in the NextQA benchmark. 
First, we observe a strong spatial bias: the image-only setting achieves a performance nearly on par with the video setting (71.3\% vs. 79.0\%). Furthermore, altering the temporal structure of videos through shuffling or reversing has minimal impact, with performance dropping by only 0.5\% and 1.4\%, respectively. 
Second, the benchmark exhibits a notable textual bias, as many questions can be answered using text alone. In the text-only setting, Tarsier-34B achieves 47.6\%, substantially outperforming the random chance baseline by 27.6\%. This highlights significant room for improvement in the benchmark’s ability to evaluate genuine temporal reasoning.

\subsection{TVBench qualitative examples}

This section provides qualitative examples for the different TVBench tasks. As shown in Fig.\ref{fig:tvbench_appendix1}-\ref{fig:tvbench_appendix7}, temporal information from the input video is indispensable for correctly answering the given questions.

\subsection{Spatial and Textual Bias in MVBench}

Fig.~\ref{fig:tvbench_first}-\ref{fig:tvbench_last} contain several evaluation examples from MVBench that suffer from the issues analyzed in Sec.~\ref{sec:mcqa}. This qualitative analysis complements the quantitative results discussed in Sec~\ref{sec:discussion} showing that these issues affect a large portion of MVBench.
\begin{figure*}
\centering
\vspace{-0.6cm}
\centering
\begin{subfigure}{0.95\textwidth}
    \centering
    \includegraphics[width=\linewidth]{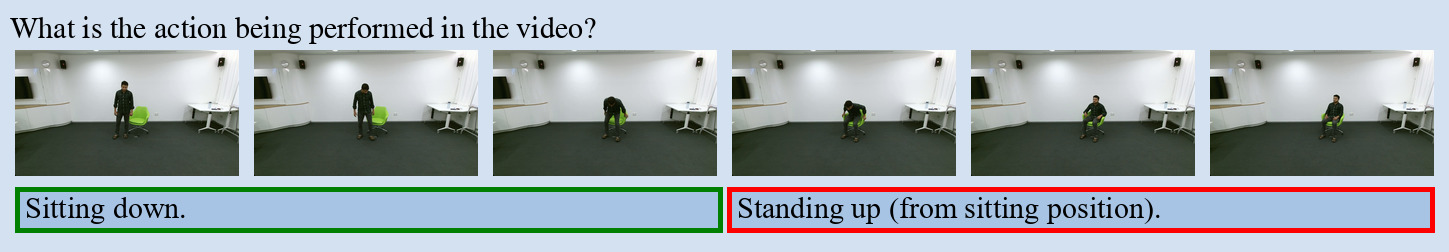}
\end{subfigure}

\begin{subfigure}{0.95\textwidth}
    \centering
    \includegraphics[width=\linewidth]{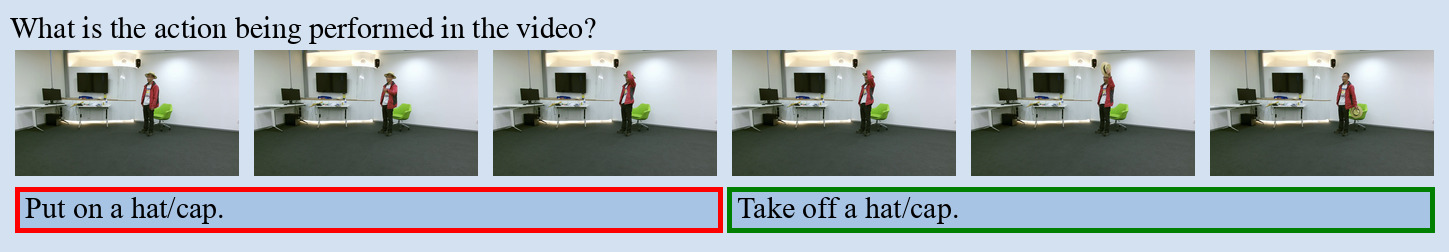}
\end{subfigure}

\begin{subfigure}{0.95\textwidth}
    \centering
    \includegraphics[width=\linewidth]{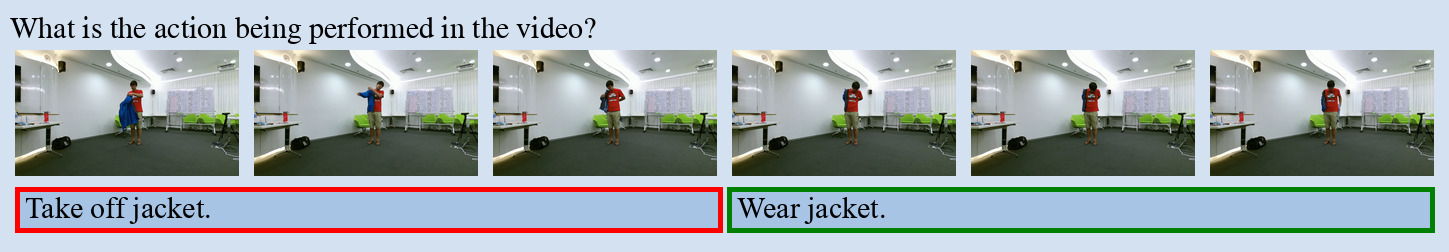}
\end{subfigure}

\begin{subfigure}{0.95\textwidth}
    \centering
    \includegraphics[width=\linewidth]{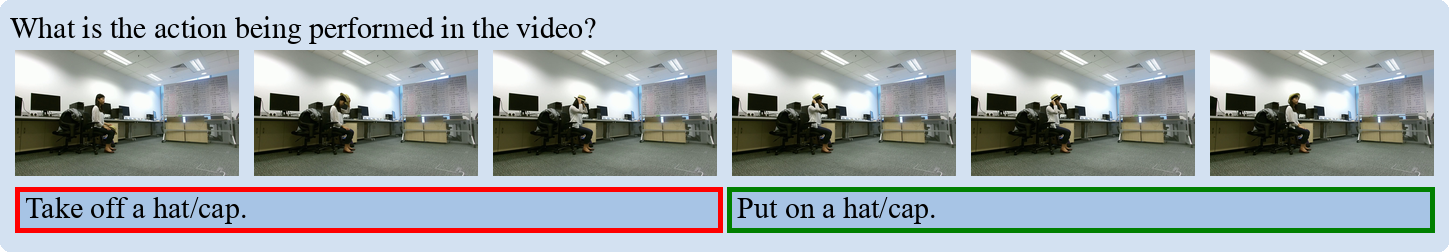}
\end{subfigure}

\begin{subfigure}{0.95\textwidth}
    \centering
    \includegraphics[width=\linewidth]{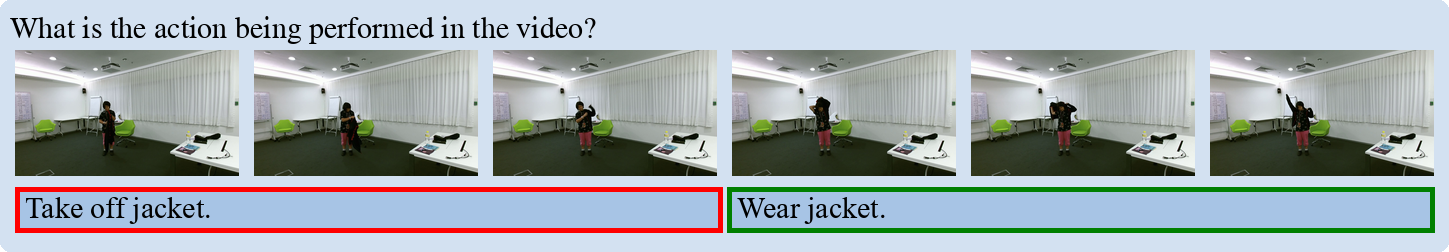}
\end{subfigure}

    \label{fig:aa_first}
\begin{subfigure}{0.95\textwidth}
    \centering
    \includegraphics[width=\linewidth]{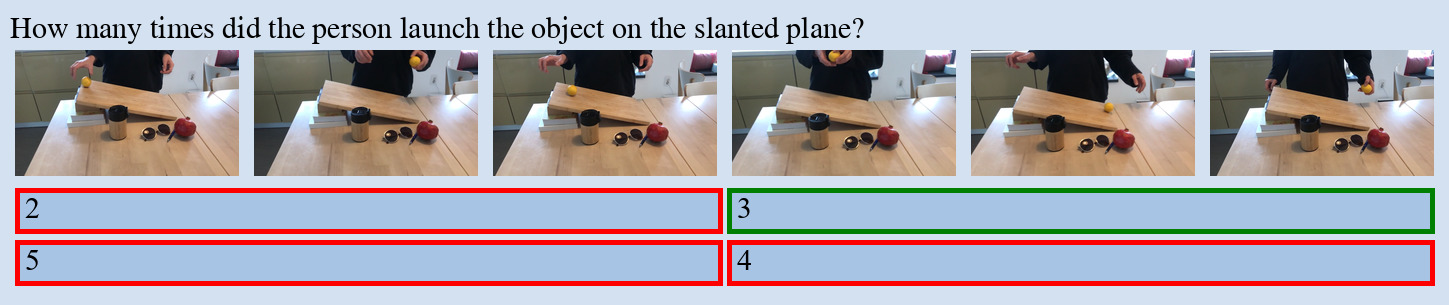}
\end{subfigure}

\begin{subfigure}{0.95\textwidth}
    \includegraphics[width=\linewidth]{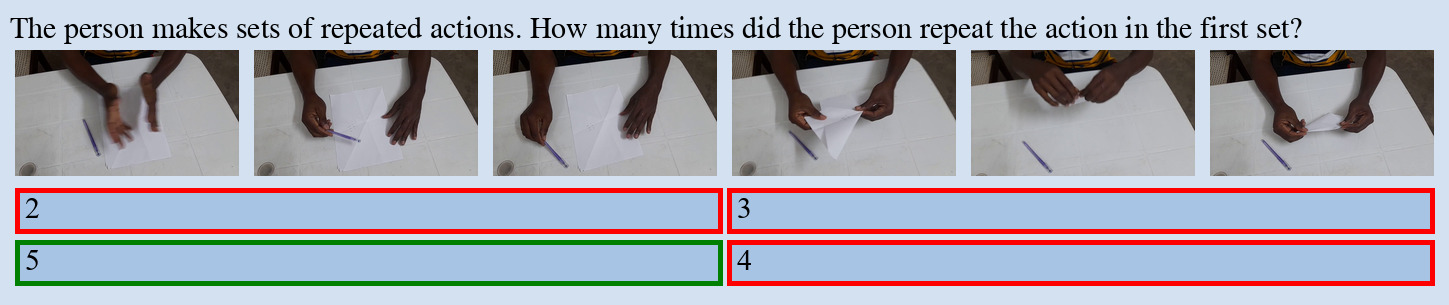}
\end{subfigure}

    \caption{\textbf{TVBench: Samples from our benchmark (1).} Row 1-5: Action Antonym; Row 6-7: Action Count.}
     \label{fig:tvbench_appendix1}

\end{figure*}

\begin{figure*}
\centering

\begin{subfigure}{0.95\textwidth}
    \includegraphics[width=\linewidth]{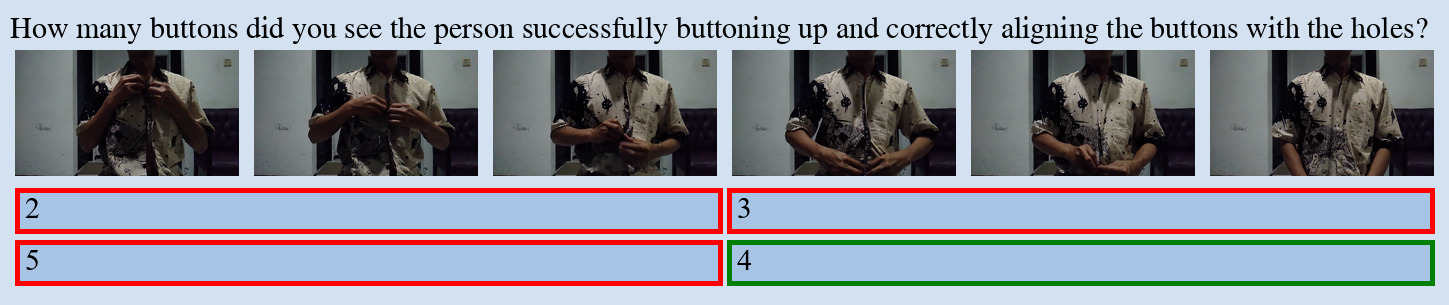}
\end{subfigure}

\begin{subfigure}{0.95\textwidth}
    \includegraphics[width=\linewidth]{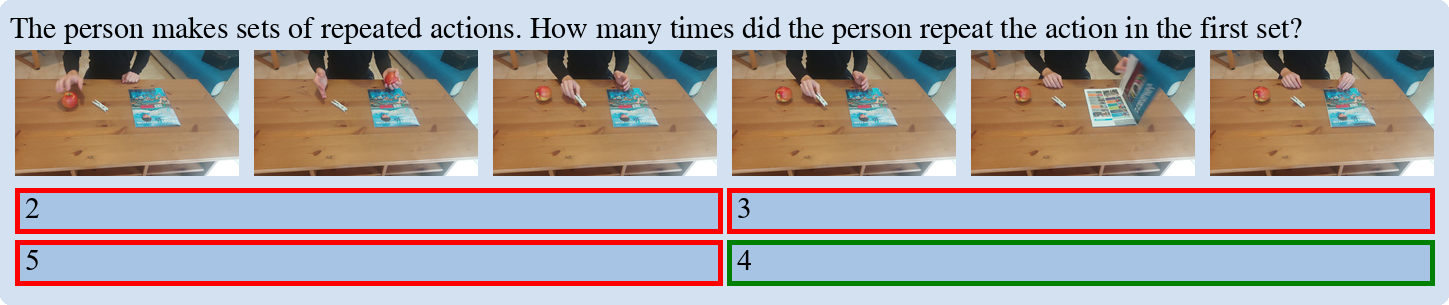}
\end{subfigure}

\begin{subfigure}{0.95\textwidth}
    \includegraphics[width=\linewidth]{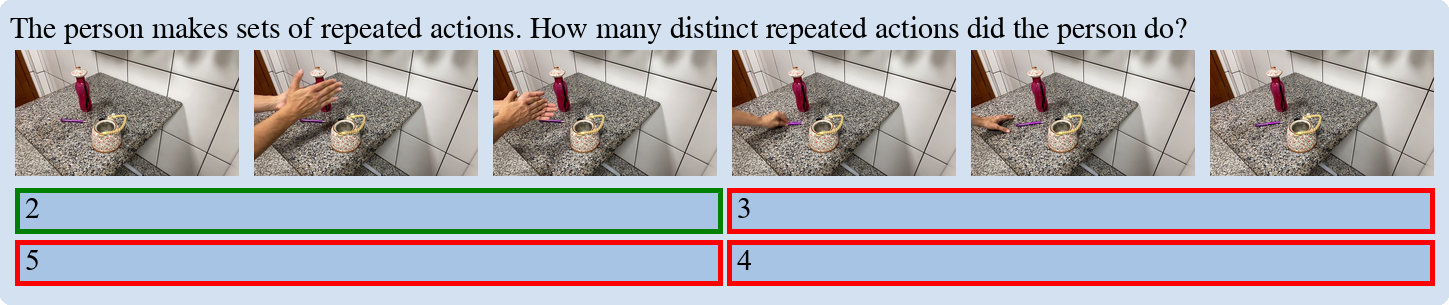}
\end{subfigure}
\centering
\begin{subfigure}{0.95\textwidth}
    \centering
    \includegraphics[width=\linewidth]{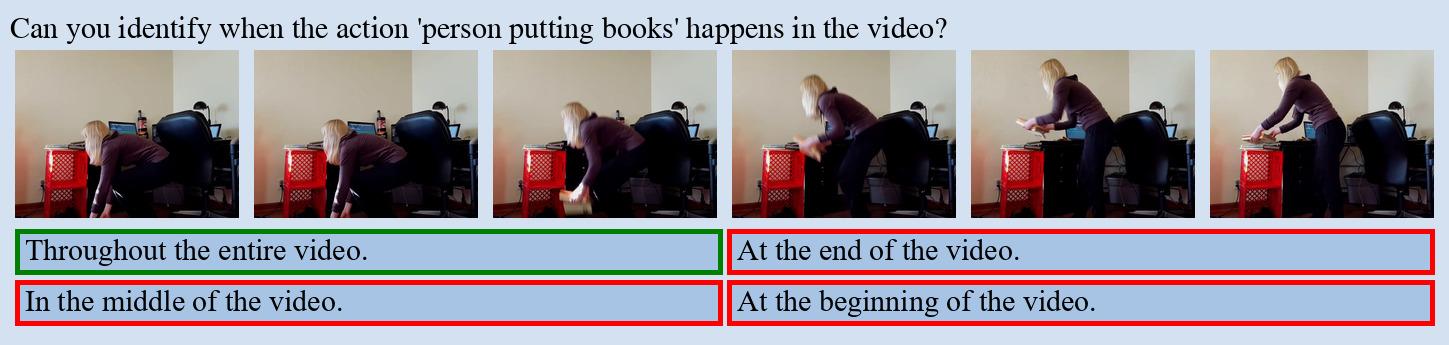}
\end{subfigure}

\begin{subfigure}{0.95\textwidth}
    \includegraphics[width=\linewidth]{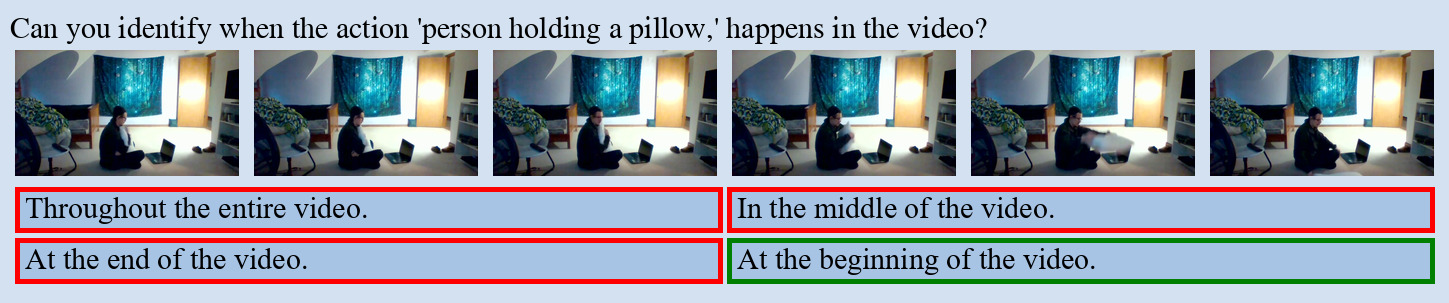}
\end{subfigure}

\begin{subfigure}{0.95\textwidth}
    \includegraphics[width=\linewidth]{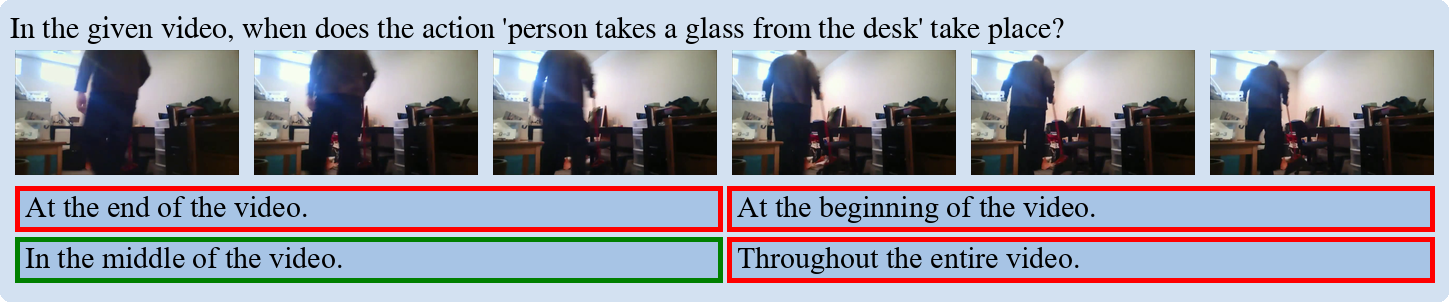}
\end{subfigure}

    \caption{\textbf{TVBench: Samples from our benchmark (2).} Row 1-3: Action Count; Row 4-6: Action Localization.}
        \label{fig:tvbench_appendix2}

\end{figure*}

\begin{figure*}
\centering

\begin{subfigure}{0.95\textwidth}
    \includegraphics[width=\linewidth]{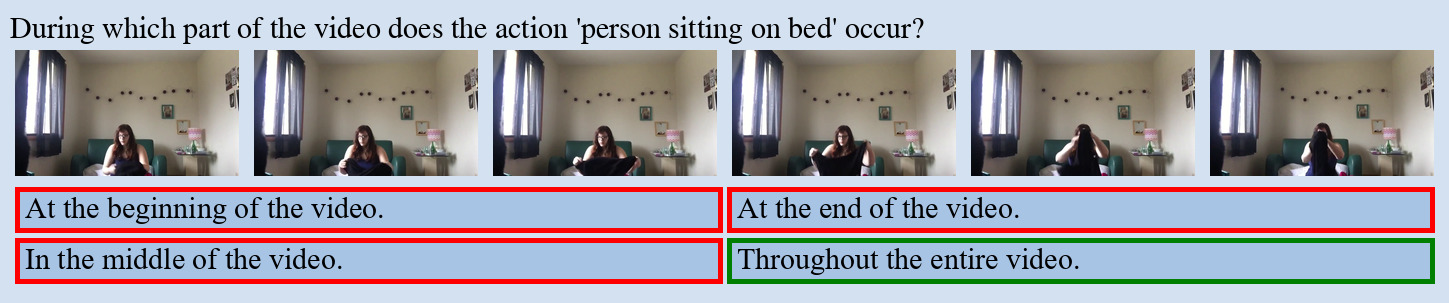}
\end{subfigure}

\centering
\begin{subfigure}{0.95\textwidth}
    \centering
    \includegraphics[width=\linewidth]{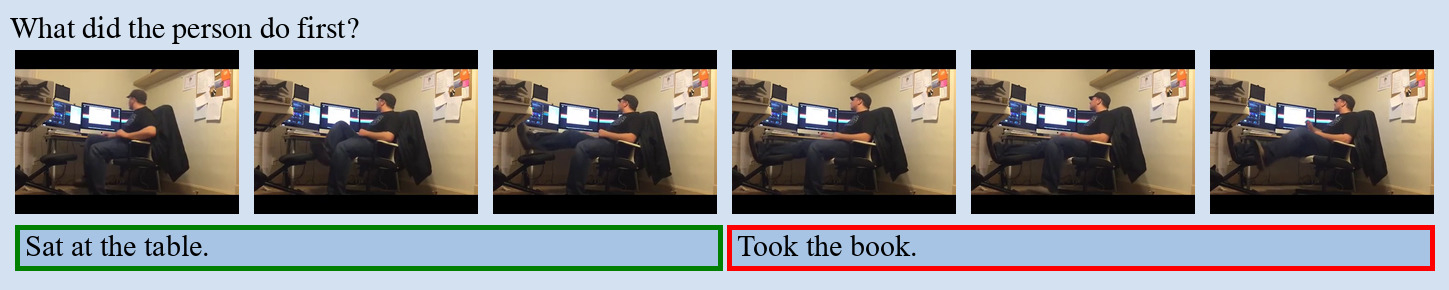}
\end{subfigure}

\begin{subfigure}{0.95\textwidth}
    \includegraphics[width=\linewidth]{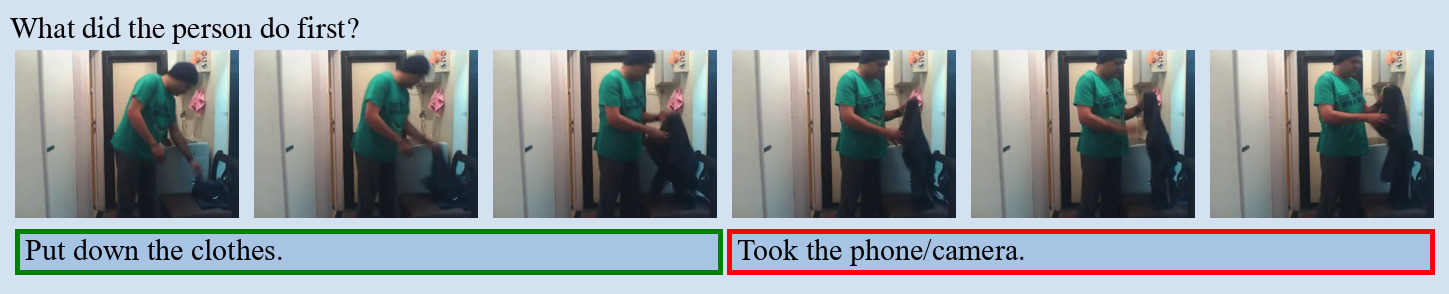}
\end{subfigure}

\begin{subfigure}{0.95\textwidth}
    \includegraphics[width=\linewidth]{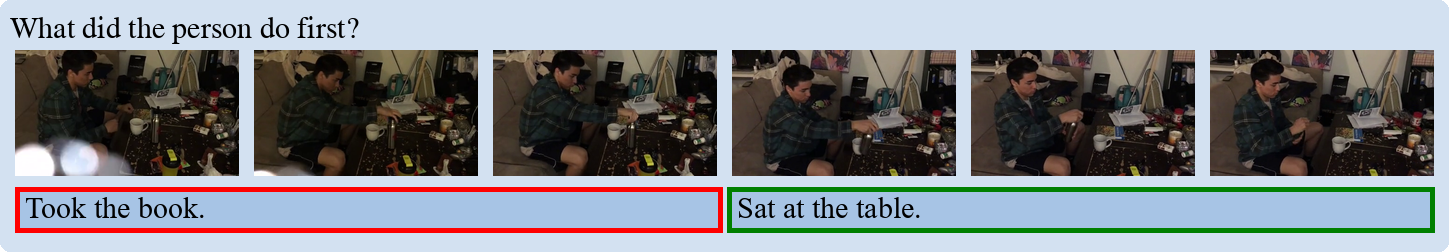}
\end{subfigure}

\begin{subfigure}{0.95\textwidth}
    \includegraphics[width=\linewidth]{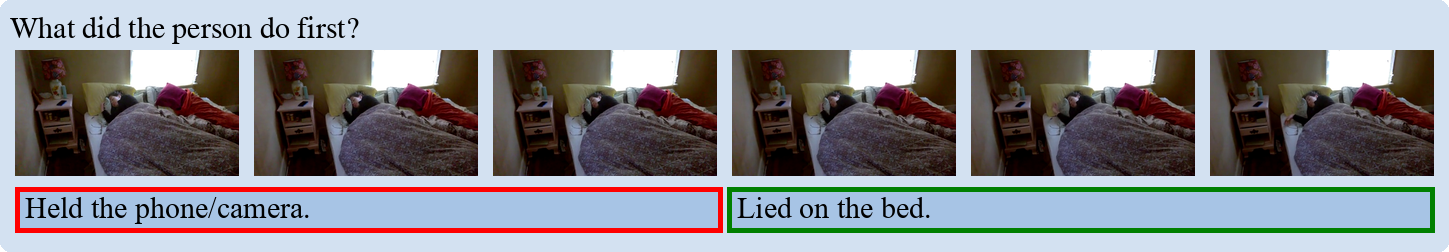}
\end{subfigure}

\begin{subfigure}{0.95\textwidth}
    \includegraphics[width=\linewidth]{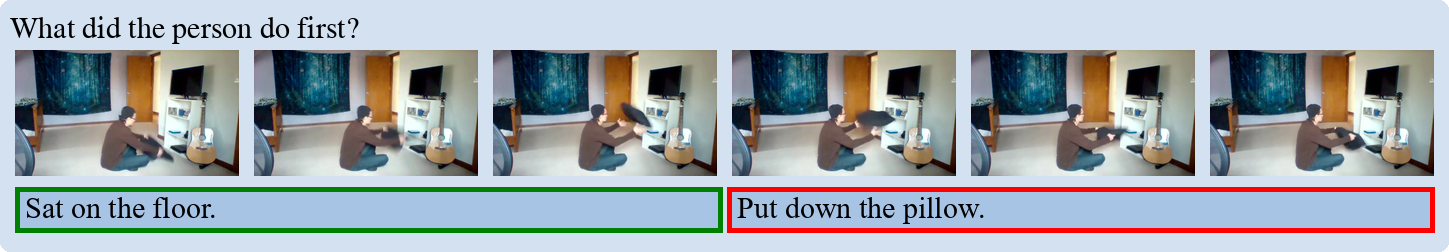}
\end{subfigure}

    \caption{\textbf{TVBench: Samples from our benchmark (3).} Row 1: Action Localization; Row 2-6: Action Sequence.}
        \label{fig:tvbench_appendix3}

\end{figure*}

\begin{figure*}
\centering
\centering
\begin{subfigure}{0.95\textwidth}
    \centering
    \includegraphics[width=\linewidth]{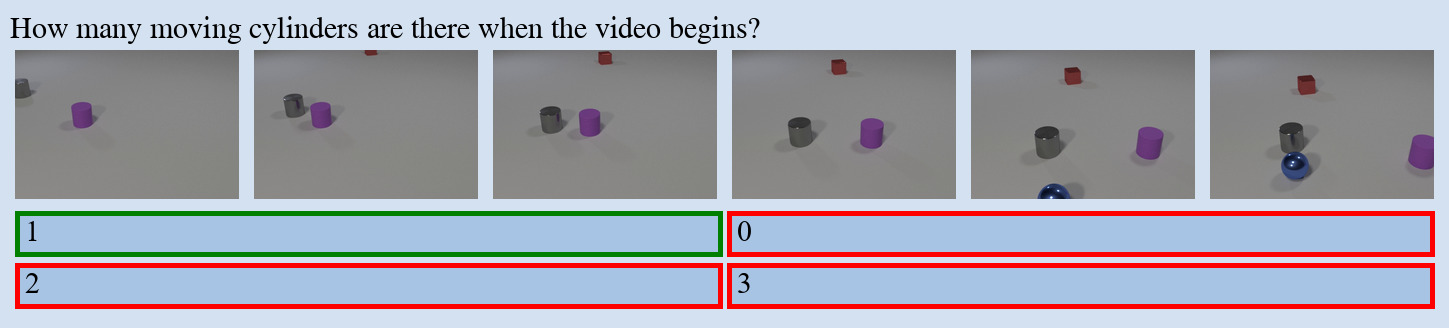}
\end{subfigure}

\begin{subfigure}{0.95\textwidth}
    \includegraphics[width=\linewidth]{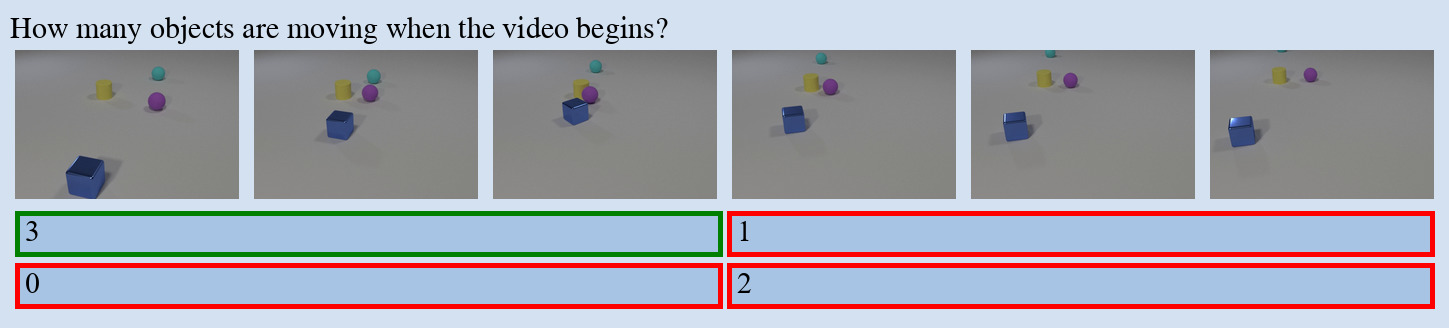}
\end{subfigure}

\begin{subfigure}{0.95\textwidth}
    \includegraphics[width=\linewidth]{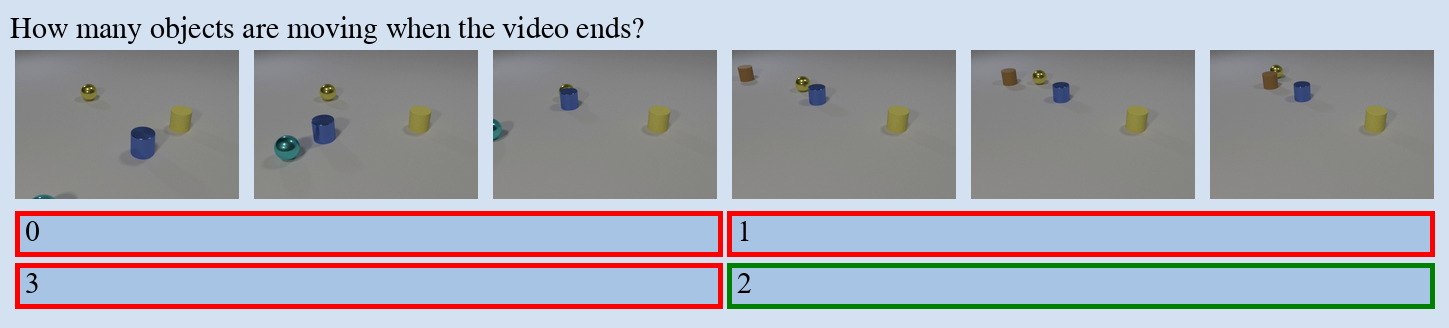}
\end{subfigure}

\begin{subfigure}{0.95\textwidth}
    \includegraphics[width=\linewidth]{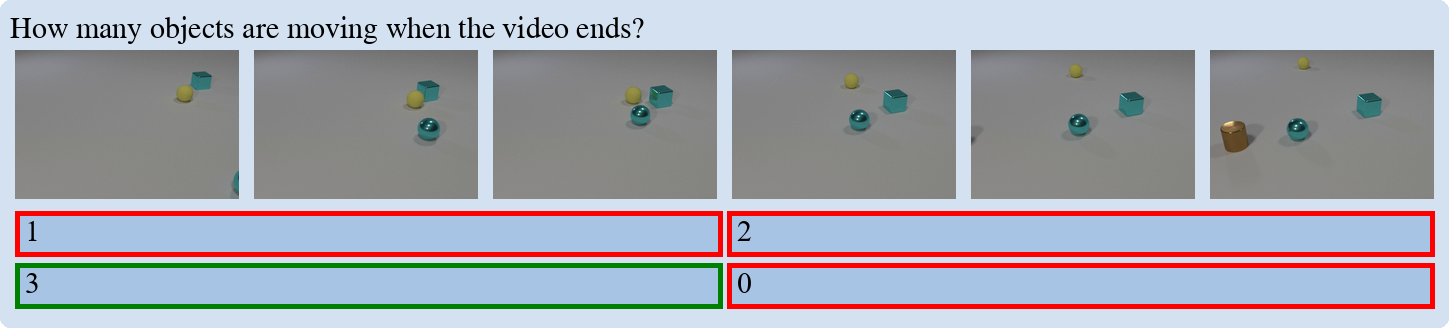}
\end{subfigure}

\begin{subfigure}{0.95\textwidth}
    \includegraphics[width=\linewidth]{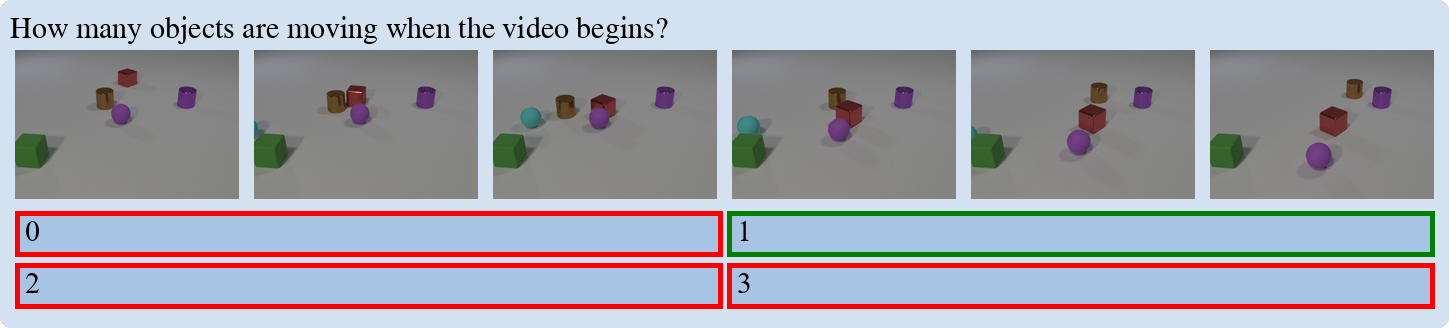}
\end{subfigure}


    \caption{\textbf{TVBench: Samples from our benchmark (4).} Row 1-5: Object Count.}
        \label{fig:tvbench_appendix4}

\end{figure*}

\begin{figure*}
\centering

\centering
\begin{subfigure}{0.95\textwidth}
    \centering
    \includegraphics[width=\linewidth]{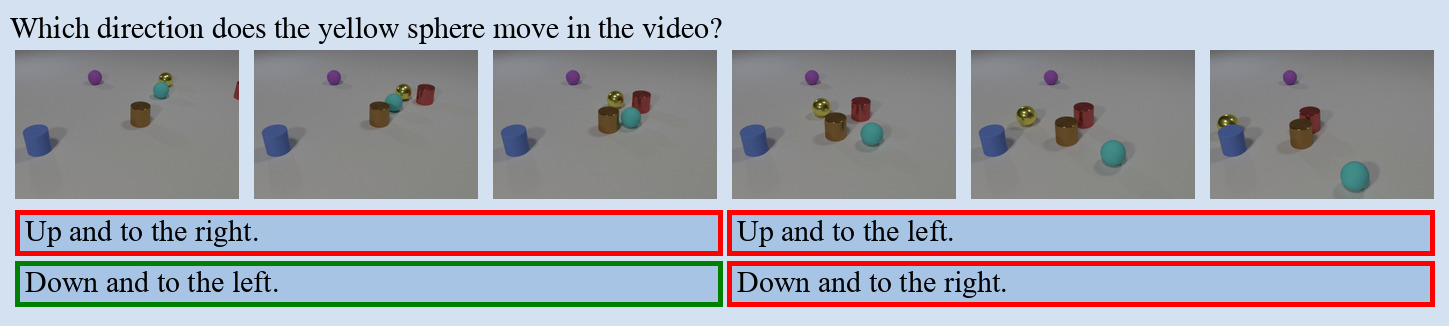}
\end{subfigure}

\begin{subfigure}{0.95\textwidth}
    \includegraphics[width=\linewidth]{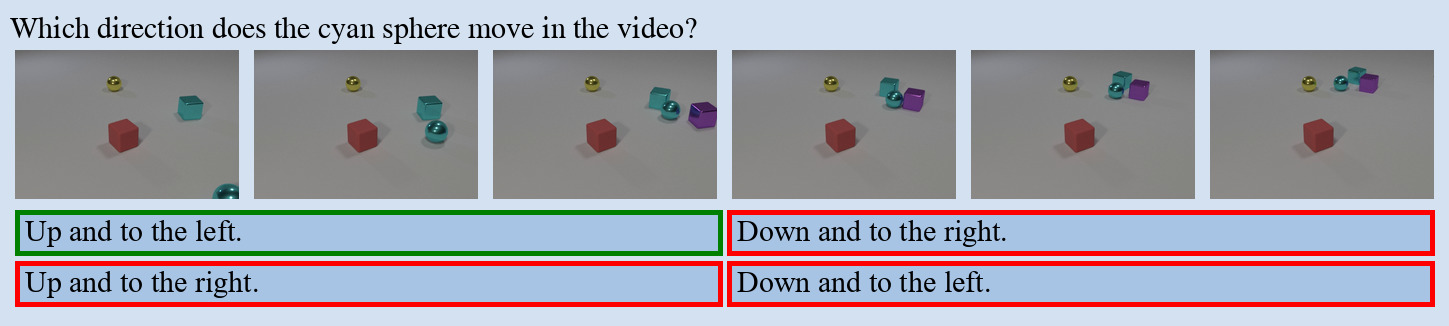}
\end{subfigure}

\begin{subfigure}{0.95\textwidth}
    \includegraphics[width=\linewidth]{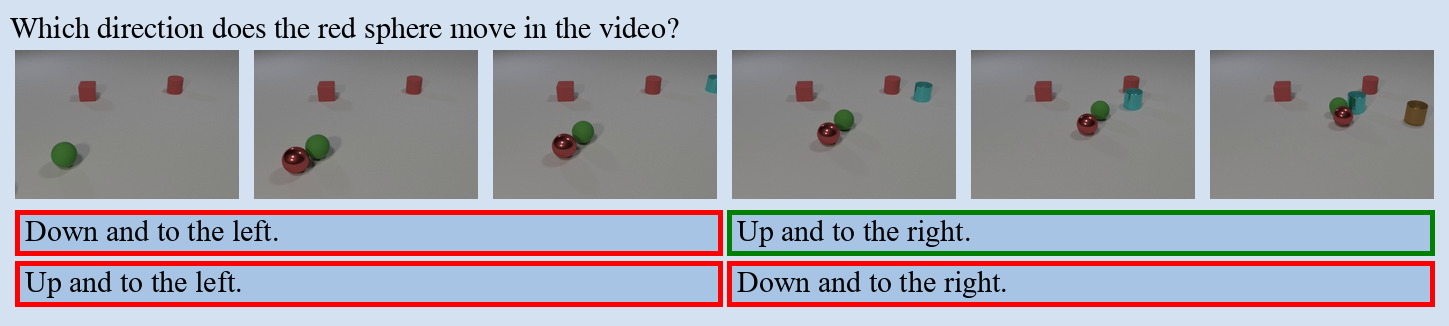}
\end{subfigure}

\begin{subfigure}{0.95\textwidth}
    \includegraphics[width=\linewidth]{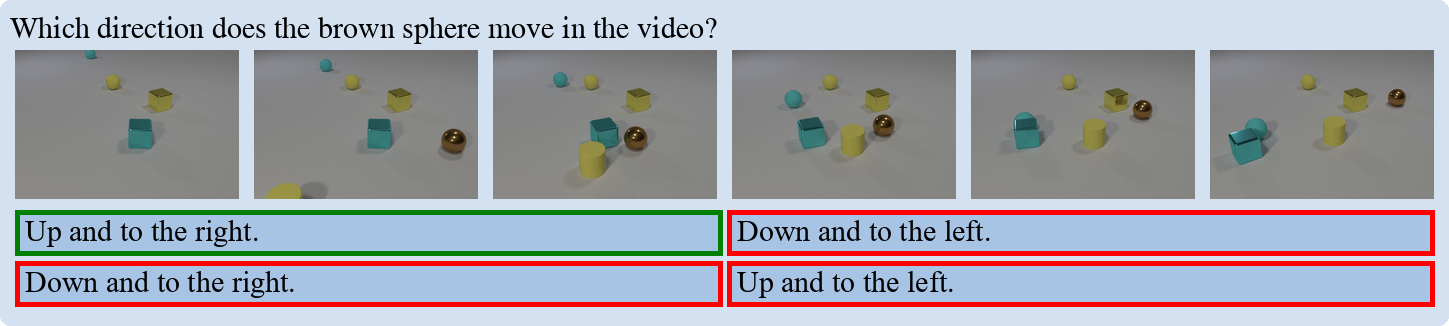}
\end{subfigure}

\begin{subfigure}{0.95\textwidth}
    \includegraphics[width=\linewidth]{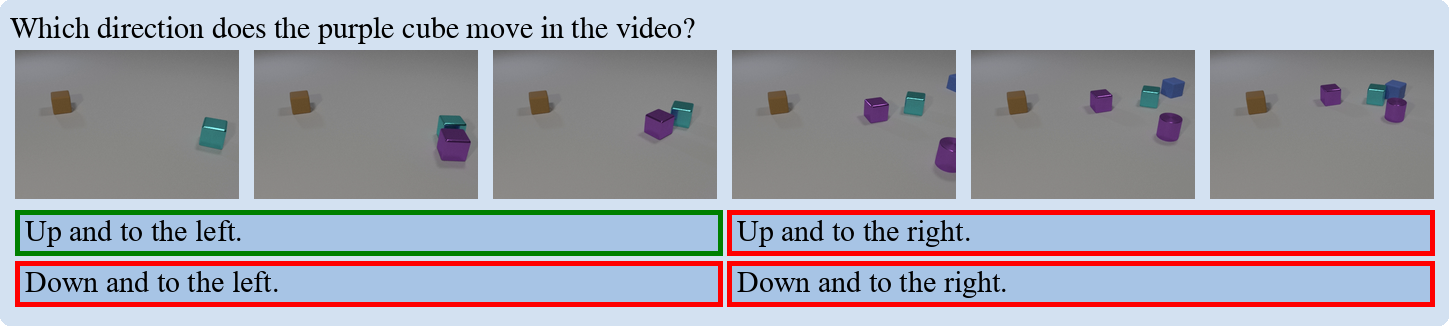}
\end{subfigure}
    \caption{\textbf{TVBench: Samples from our benchmark (5).} Row 1-5: Moving Direction.}
        \label{fig:tvbench_appendix5}

\end{figure*}

\begin{figure*}
\centering
    
\centering
\begin{subfigure}{0.95\textwidth}
    \centering
    \includegraphics[width=\linewidth]{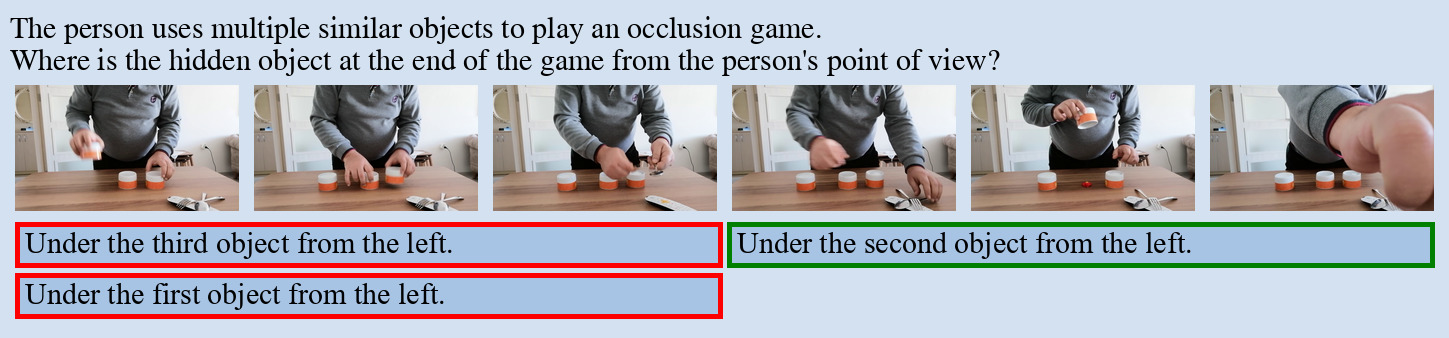}
\end{subfigure}

\begin{subfigure}{0.95\textwidth}
    \includegraphics[width=\linewidth]{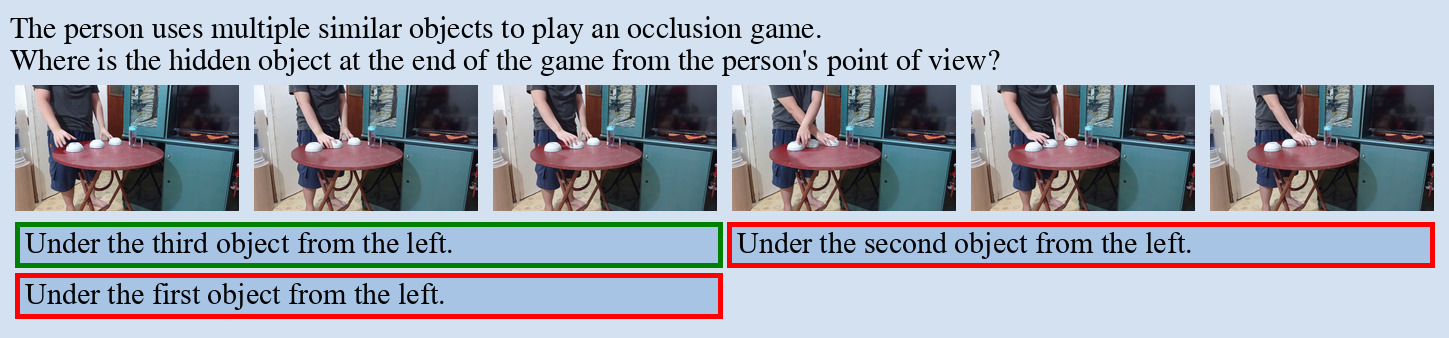}
\end{subfigure}

\begin{subfigure}{0.95\textwidth}
    \includegraphics[width=\linewidth]{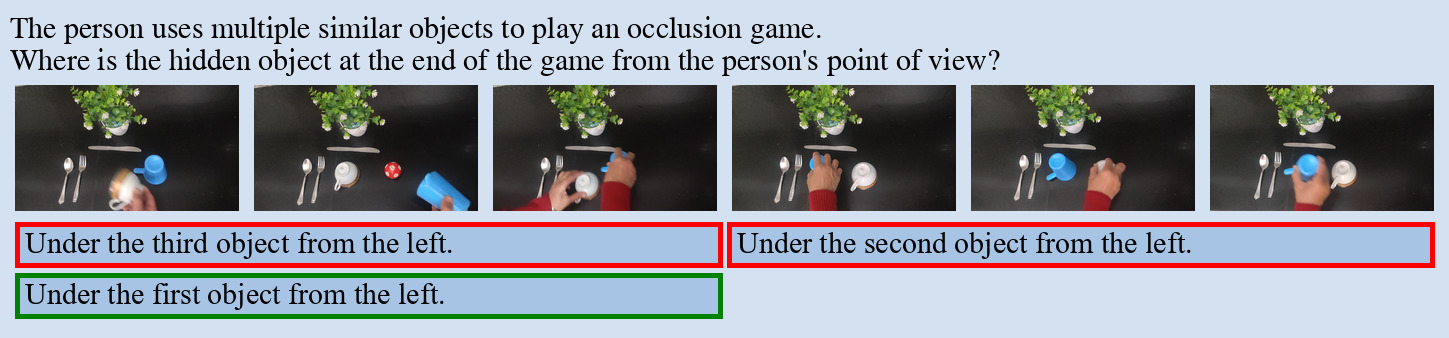}
\end{subfigure}

\begin{subfigure}{0.95\textwidth}
    \includegraphics[width=\linewidth]{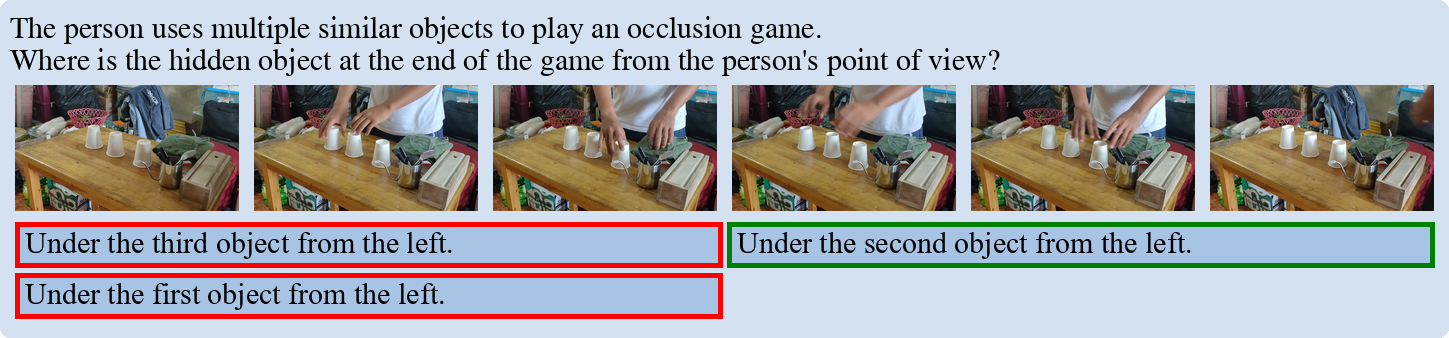}
\end{subfigure}

\begin{subfigure}{0.95\textwidth}
    \includegraphics[width=\linewidth]{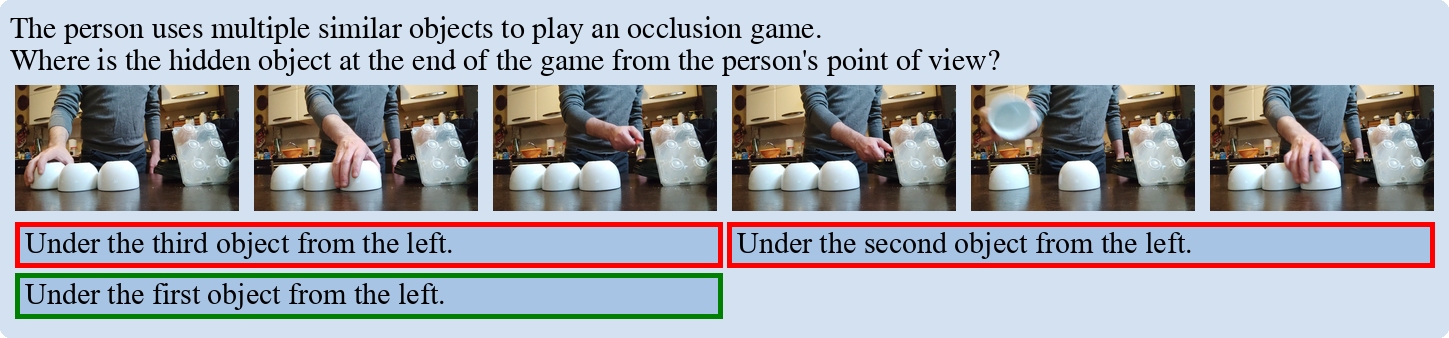}
\end{subfigure}

    \caption{\textbf{TVBench: Samples from our benchmark (6).} Row 1-5: Object Shuffle.}
        \label{fig:tvbench_appendix5-continued}

\end{figure*}

\begin{figure*}
\centering

\begin{subfigure}{0.95\textwidth}
    \centering
    \includegraphics[width=\linewidth]{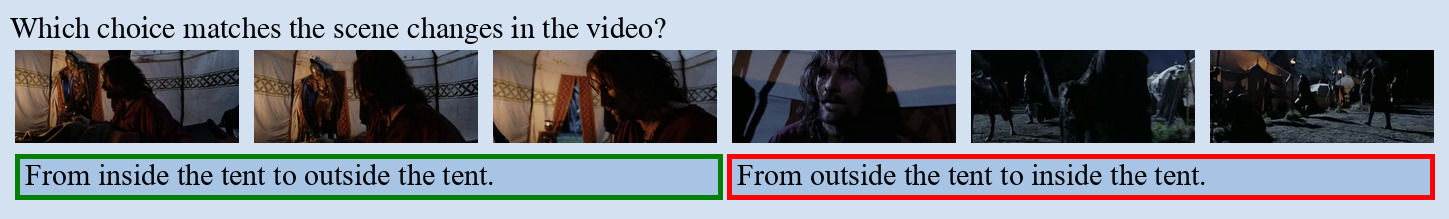}
\end{subfigure}

\begin{subfigure}{0.95\textwidth}
    \includegraphics[width=\linewidth]{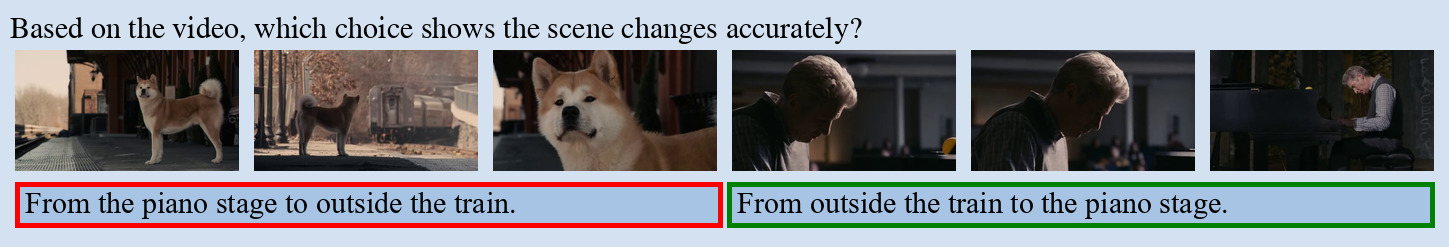}
\end{subfigure}

\begin{subfigure}{0.95\textwidth}
    \centering
    \includegraphics[width=\linewidth]{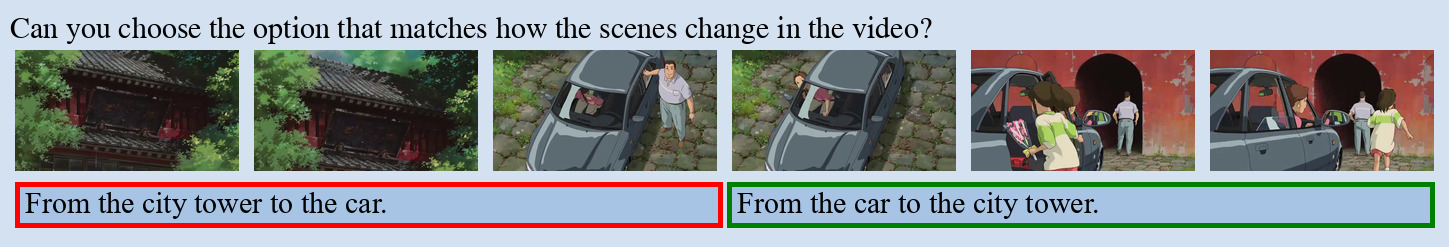}
\end{subfigure}

\begin{subfigure}{0.95\textwidth}
    \includegraphics[width=\linewidth]{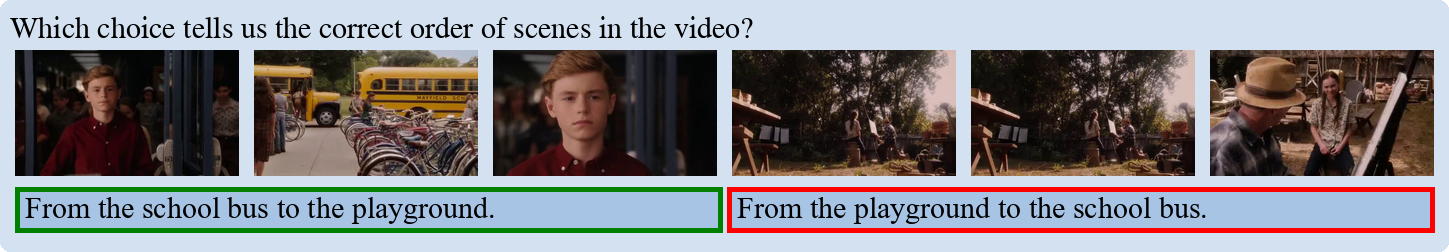}
\end{subfigure}

\begin{subfigure}{0.95\textwidth}
    \centering
    \includegraphics[width=\linewidth]{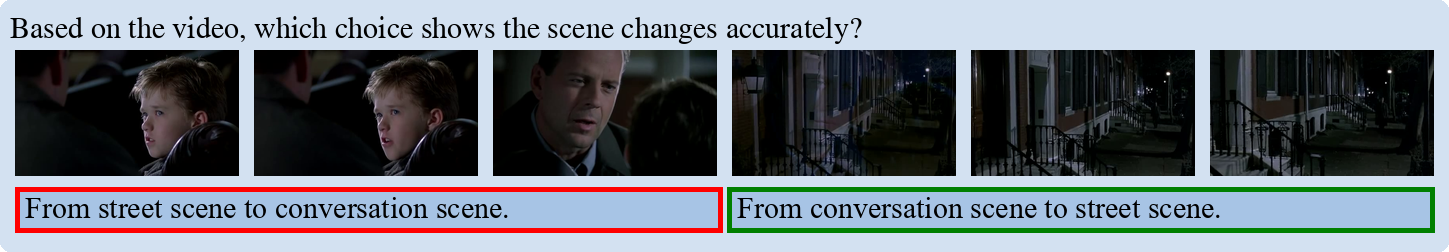}
\end{subfigure}

        \caption{\textbf{TVBench: Samples from our benchmark (7).} Row 1-5: Scene Transition.}
            \label{fig:tvbench_appendix6}

\end{figure*}

\begin{figure*}
\centering[t]
\centering


\centering
\begin{subfigure}{0.95\textwidth}
    \centering
    \includegraphics[width=\linewidth]{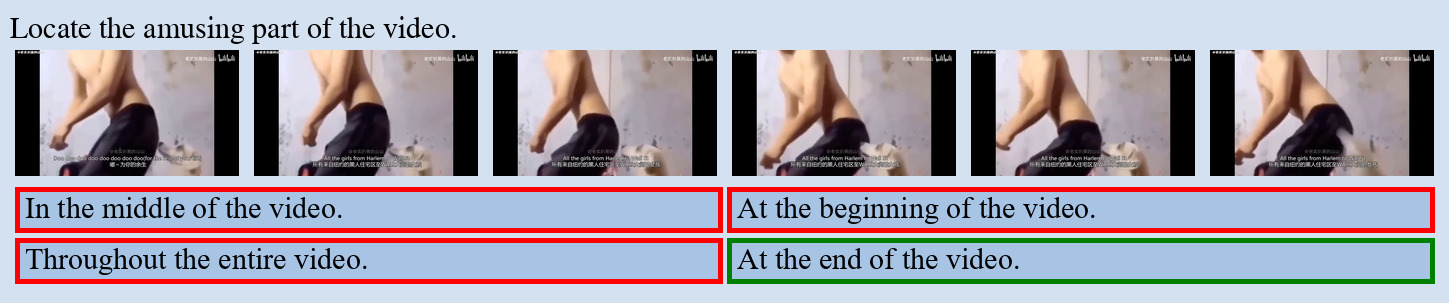}
\end{subfigure}

\begin{subfigure}{0.95\textwidth}
    \includegraphics[width=\linewidth]{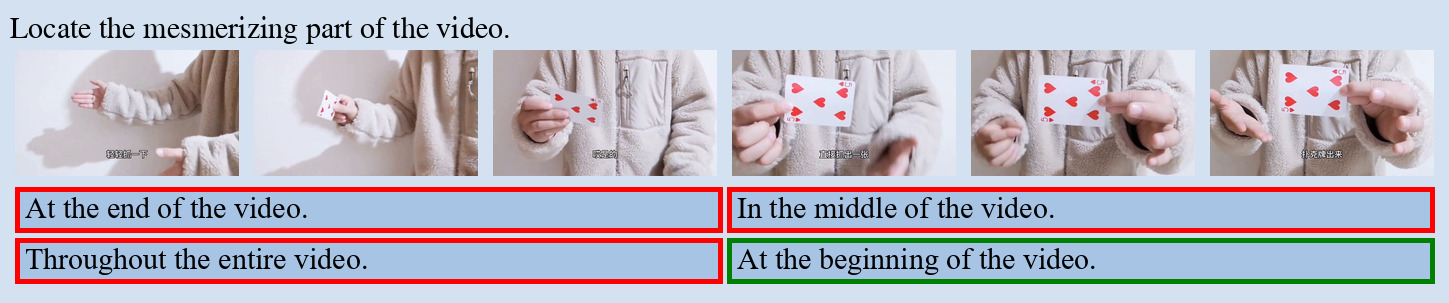}
\end{subfigure}

\begin{subfigure}{0.95\textwidth}
    \centering
    \includegraphics[width=\linewidth]{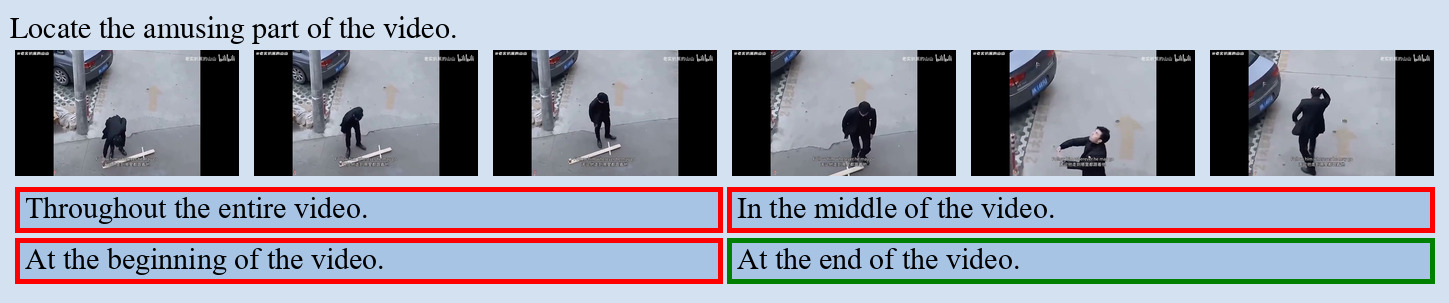}
\end{subfigure}

\begin{subfigure}{0.95\textwidth}
    \includegraphics[width=\linewidth]{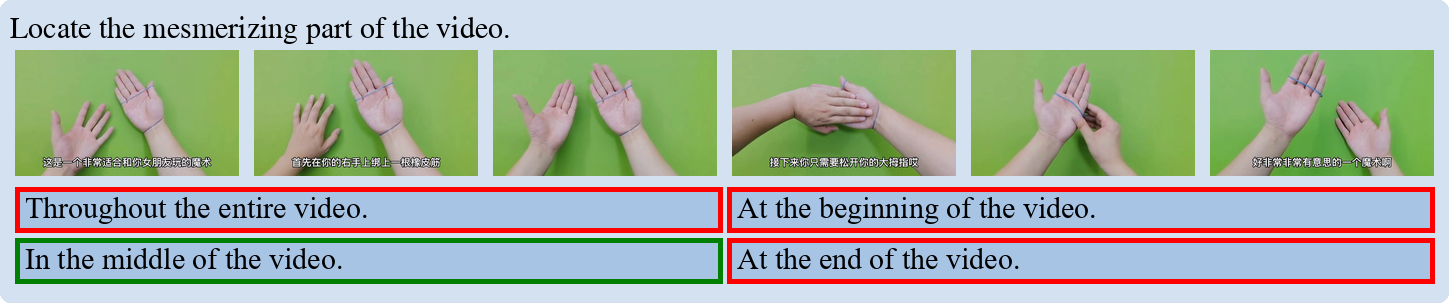}
\end{subfigure}

\begin{subfigure}{0.95\textwidth}
    \centering
    \includegraphics[width=\linewidth]{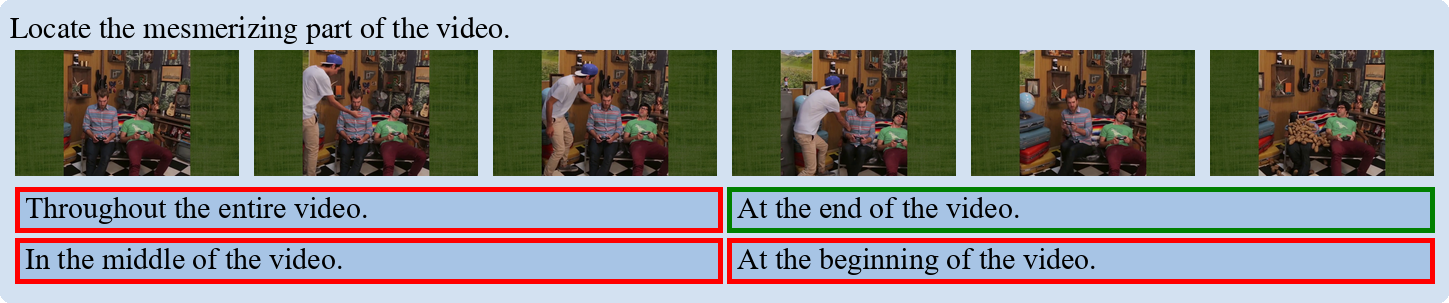}
\end{subfigure}

        \caption{\textbf{TVBench: Samples from our benchmark (8).} Row 1: Scene Transition; Row 2-6: Unexpected Action.}
    \label{fig:tvbench_appendix7}
\end{figure*}

\begin{figure*}
\centering
\centering
\begin{subfigure}{0.95\textwidth}
    \centering
    \includegraphics[width=\linewidth]{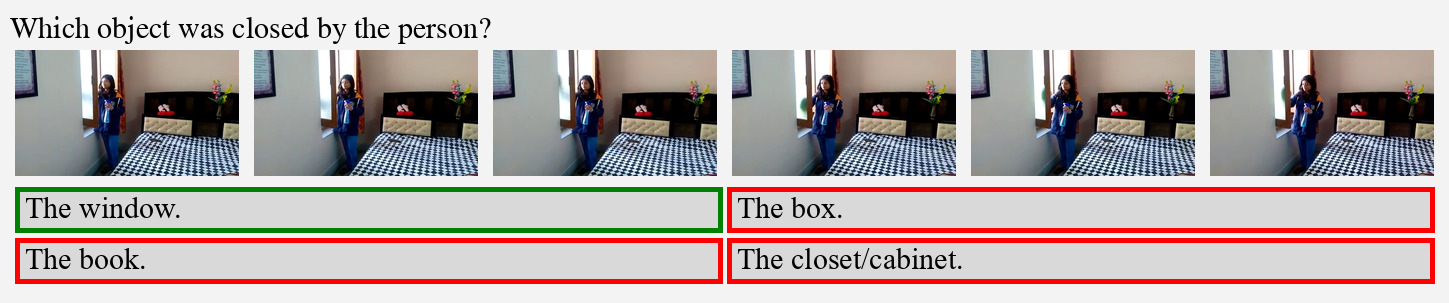}
\end{subfigure}
\begin{subfigure}{0.95\textwidth}
    \includegraphics[width=\linewidth]{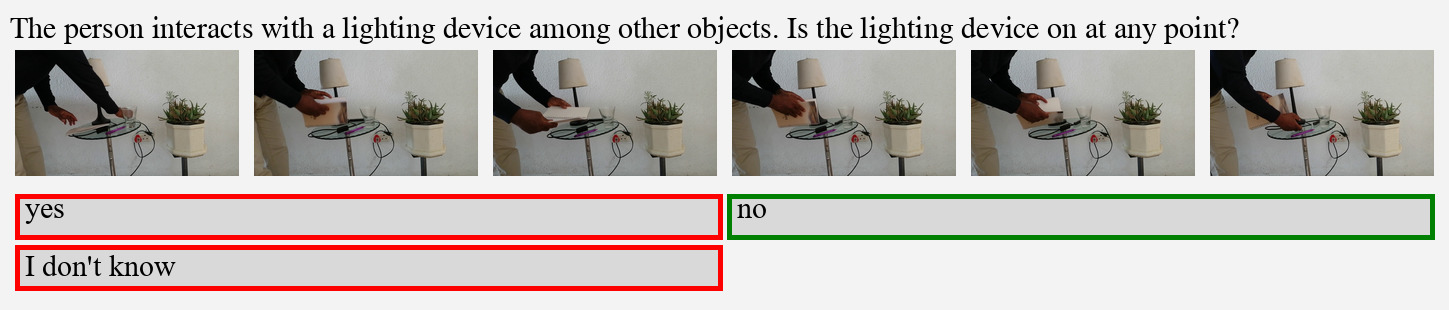}
\end{subfigure}

\begin{subfigure}{0.95\textwidth}
    \includegraphics[width=\linewidth]{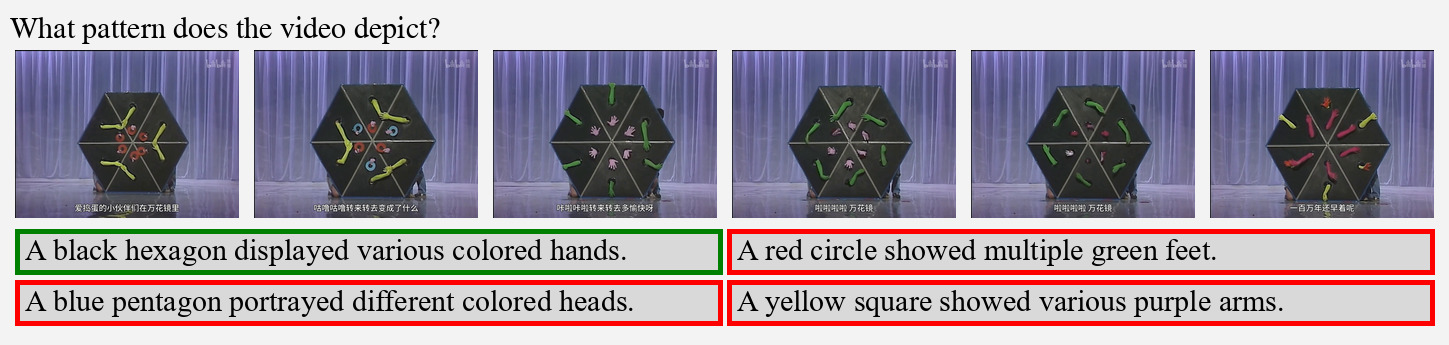}
\end{subfigure}

\begin{subfigure}{0.95\textwidth}
    \includegraphics[width=\linewidth]{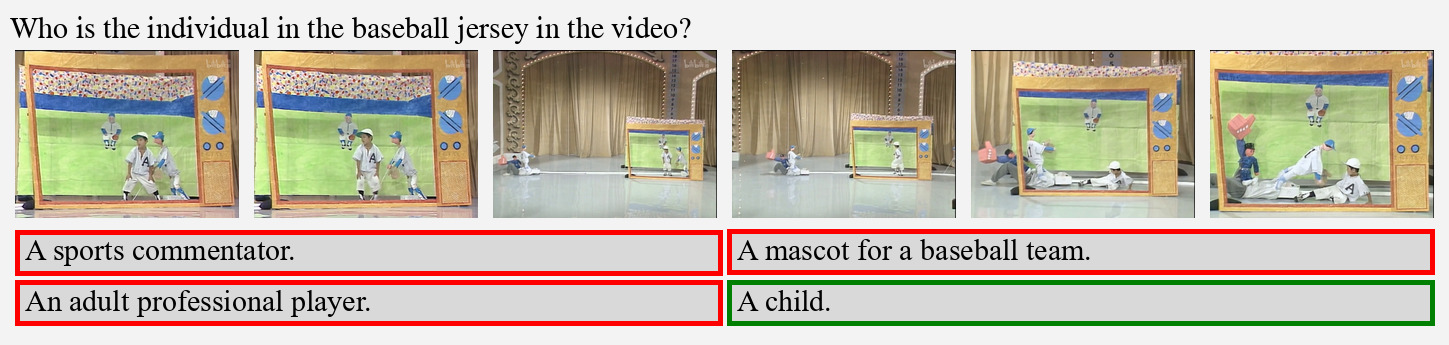}
\end{subfigure}

\begin{subfigure}{0.95\textwidth}
    \includegraphics[width=\linewidth]{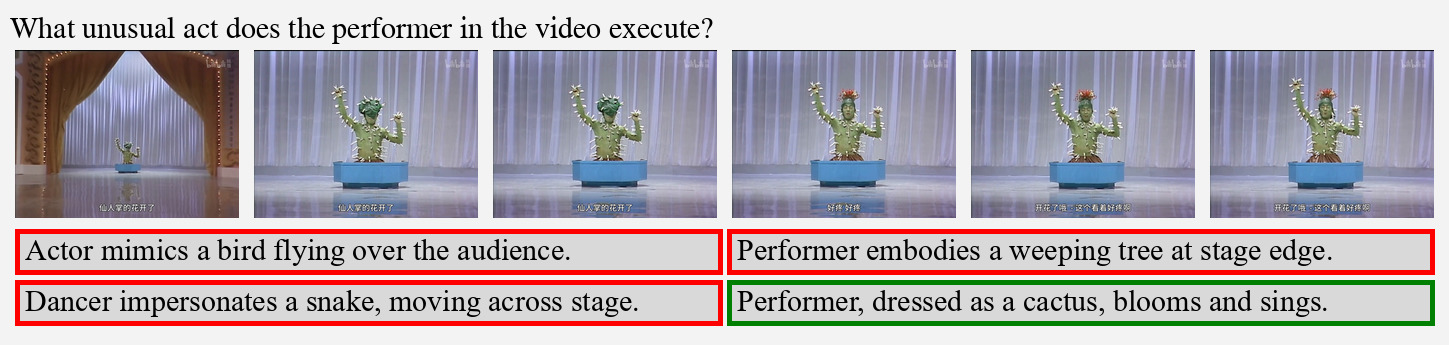}
\end{subfigure}

    \caption{\textbf{Spatial bias in MVBench (1).} Multiple-choice questions from various MVBench tasks can be solved using only a single frame from the video.}

\label{fig:tvbench_first}
\end{figure*}

\begin{figure*}
\centering
\centering
\begin{subfigure}{0.95\textwidth}
    \includegraphics[width=\linewidth]{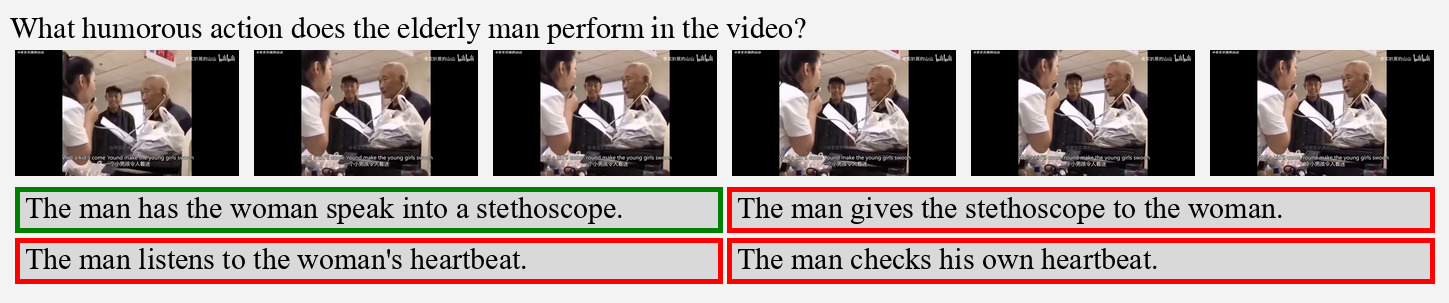}
\end{subfigure}

\begin{subfigure}{0.95\textwidth}
    \includegraphics[width=\linewidth]{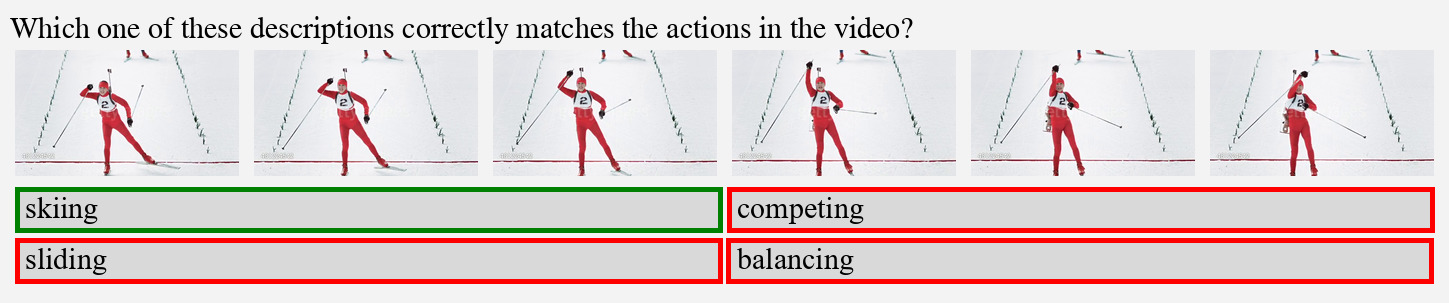}
\end{subfigure}

\begin{subfigure}{0.95\textwidth}
    \includegraphics[width=\linewidth]{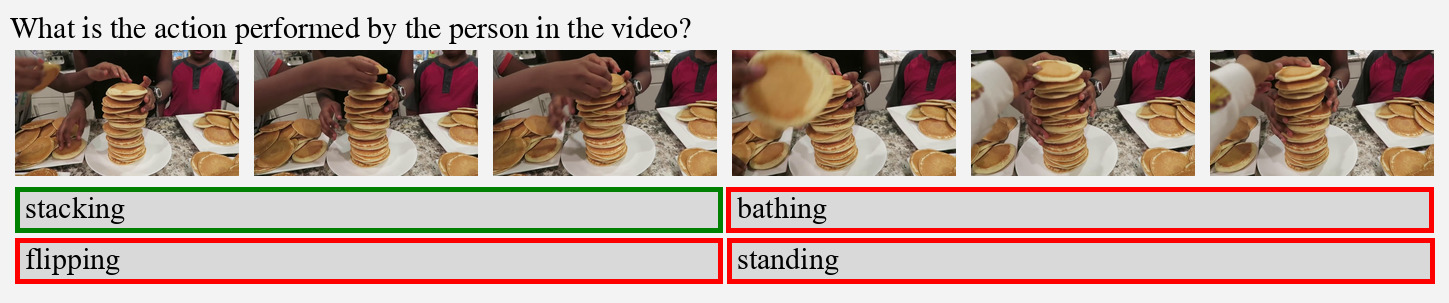}
\end{subfigure}

\begin{subfigure}{0.95\textwidth}
    \centering
    \includegraphics[width=\linewidth]{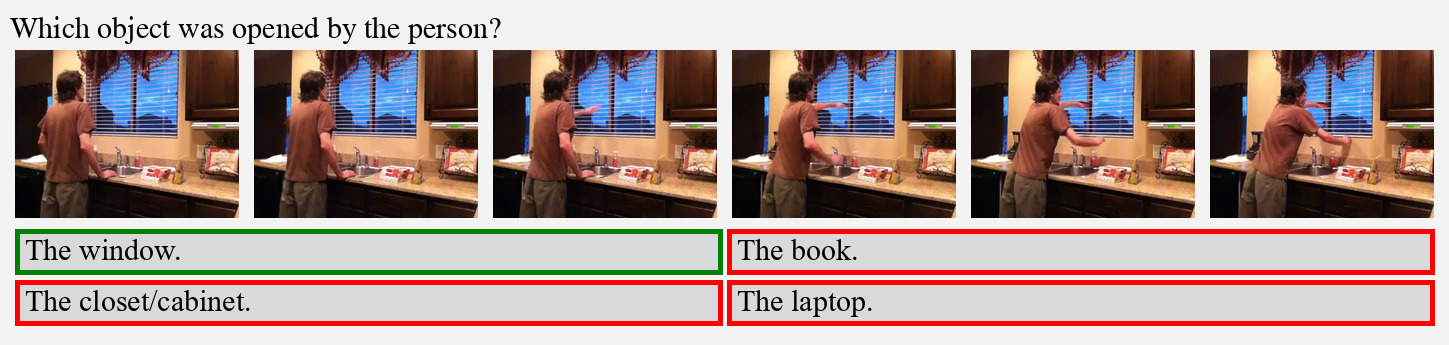}
\end{subfigure}

    \caption{\textbf{Spatial bias in MVBench (2).} Multiple-choice questions from various MVBench tasks can be solved using only a single frame from the video.}
\end{figure*}

\begin{figure*}
\centering
\centering
\begin{subfigure}{0.95\textwidth}
    \centering
    \includegraphics[width=\linewidth]{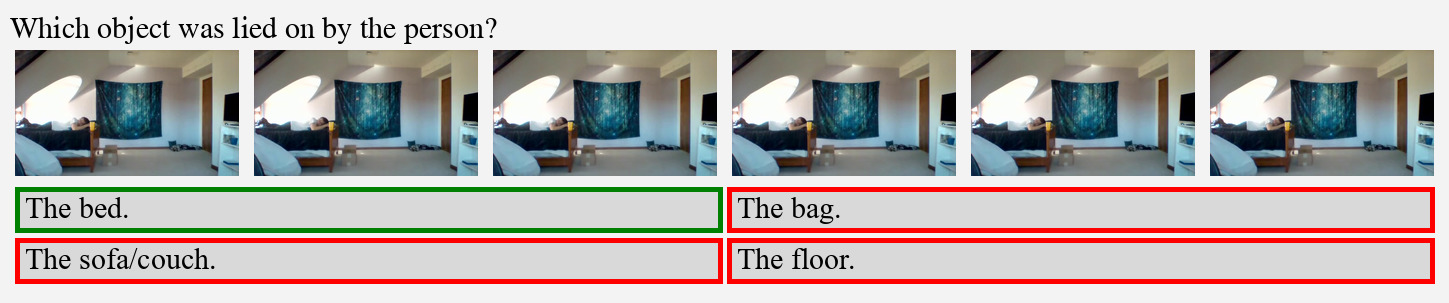}
\end{subfigure}
\begin{subfigure}{0.95\textwidth}
    \centering
    \includegraphics[width=\linewidth]{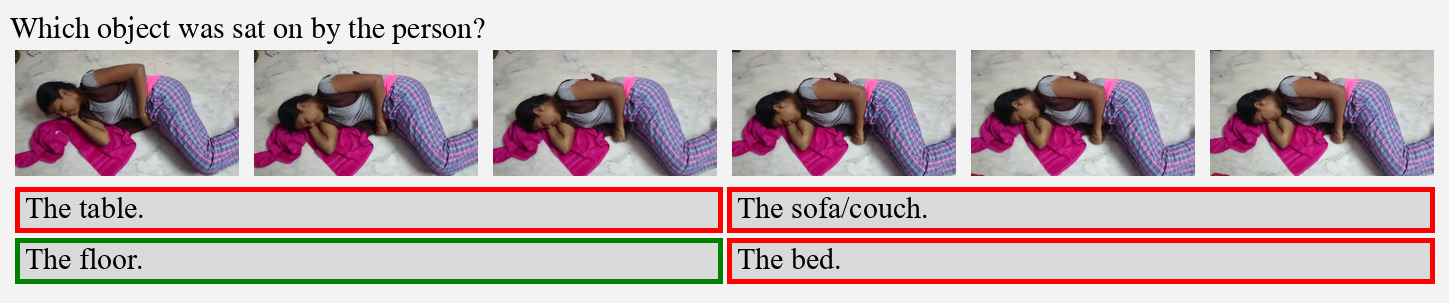}
\end{subfigure}
\begin{subfigure}{0.95\textwidth}
    \centering
    \includegraphics[width=\linewidth]{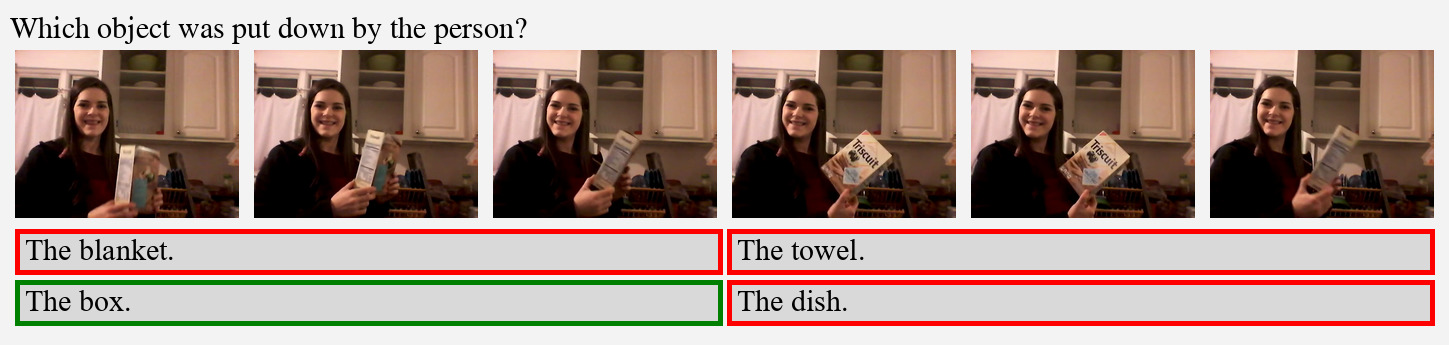}
\end{subfigure}

\begin{subfigure}{0.95\textwidth}
    \centering
    \includegraphics[width=\linewidth]{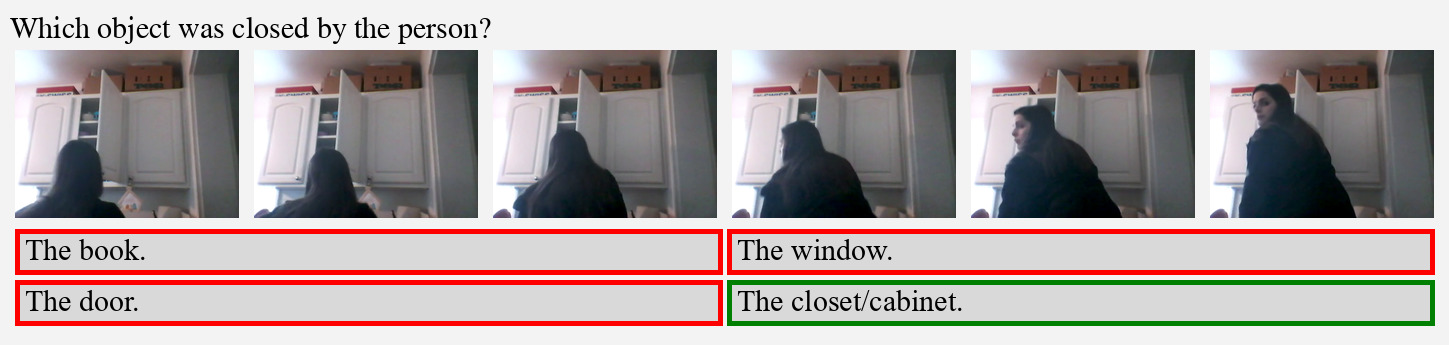}
\end{subfigure}

    \caption{\textbf{Spatial bias in MVBench (3).} Multiple-choice questions from various MVBench tasks can be solved using only a single frame from the video.}
\end{figure*}

\begin{figure*}
\centering
\begin{subfigure}{0.95\textwidth}
    \includegraphics[width=\linewidth]{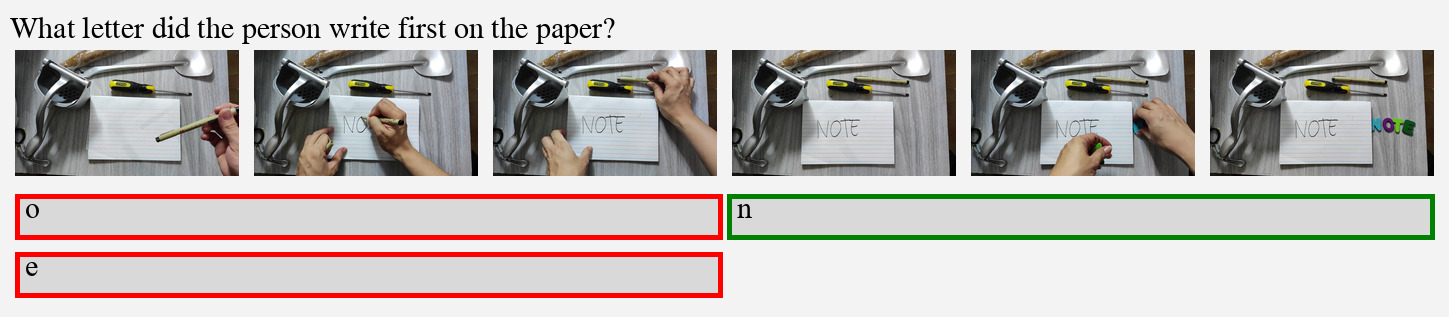}
\end{subfigure}

\begin{subfigure}{0.95\textwidth}
    \includegraphics[width=\linewidth]{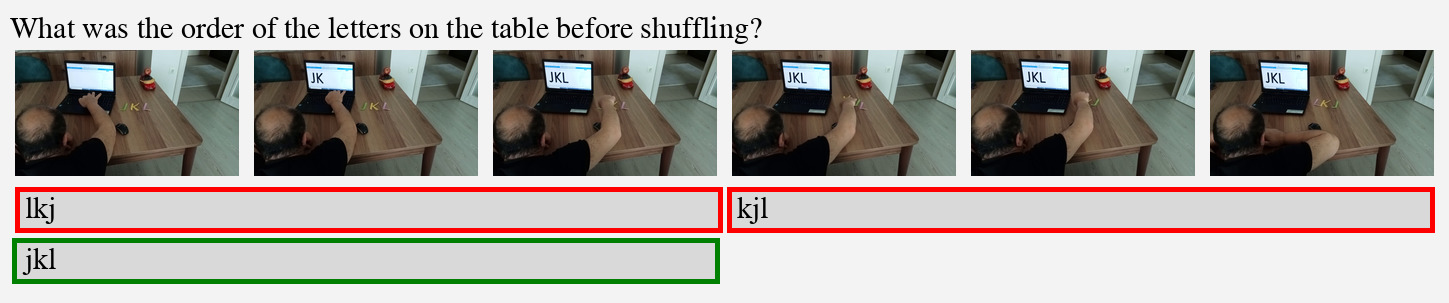}
\end{subfigure}

\begin{subfigure}{0.95\textwidth}
    \includegraphics[width=\linewidth]{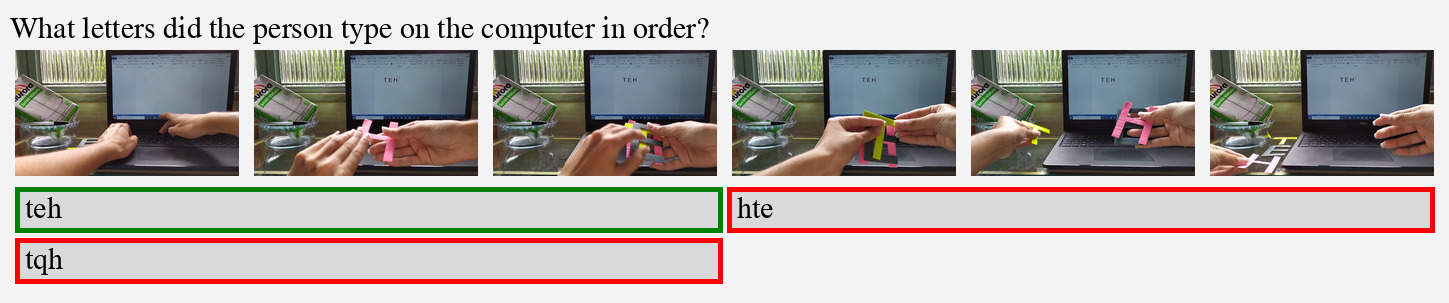}
\end{subfigure}

\begin{subfigure}{0.95\textwidth}
    \includegraphics[width=\linewidth]{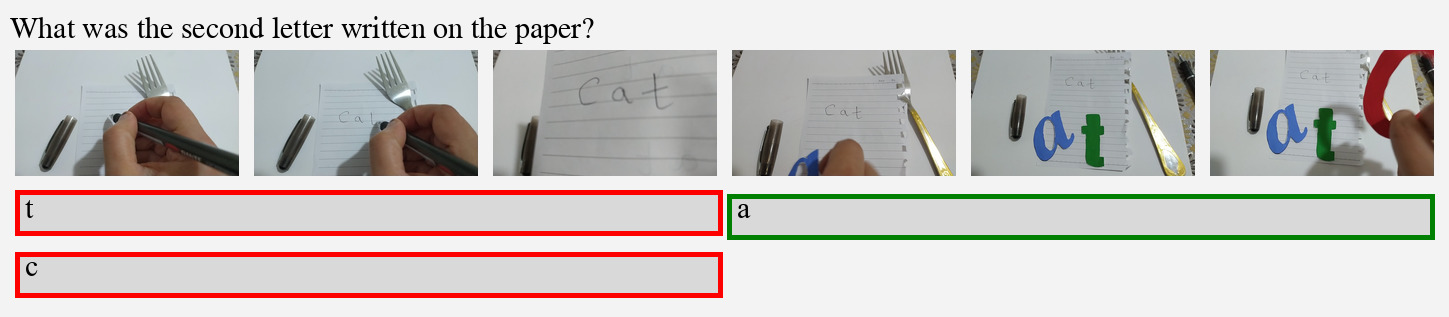}
\end{subfigure}

\begin{subfigure}{0.95\textwidth}
    \includegraphics[width=\linewidth]{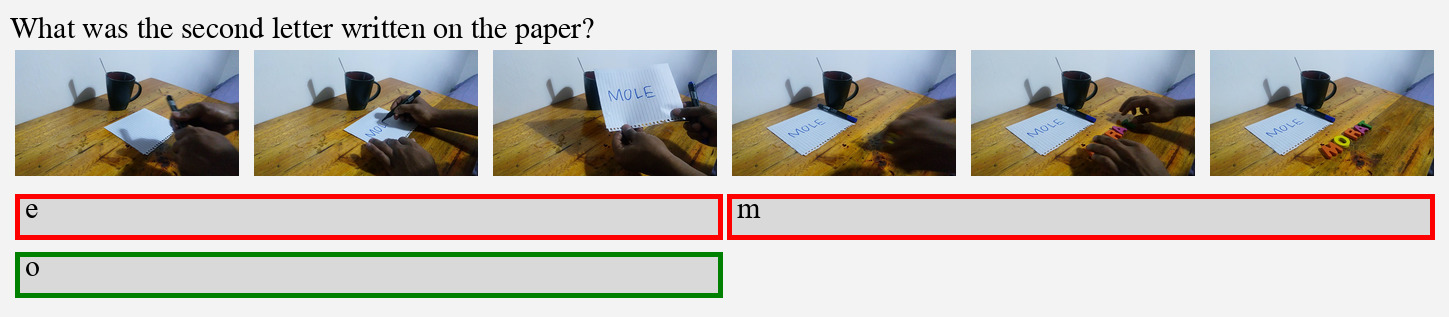}
\end{subfigure}

\begin{subfigure}{0.95\textwidth}
    \includegraphics[width=\linewidth]{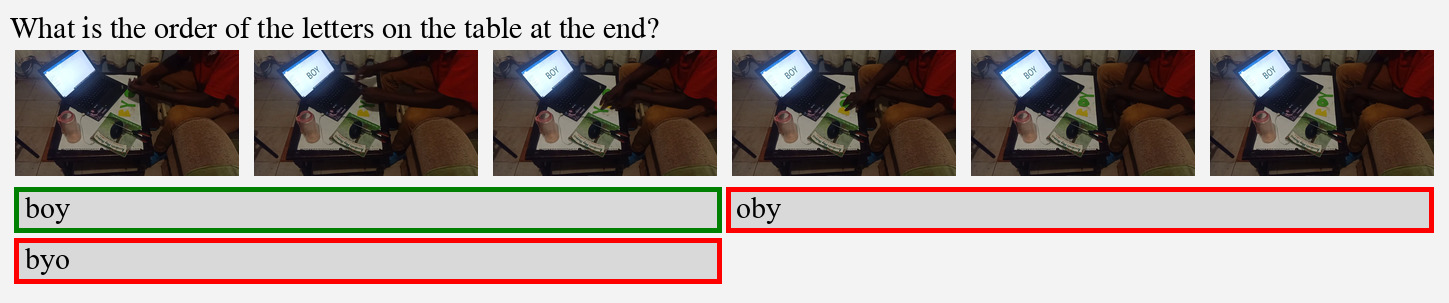}
\end{subfigure}

    \caption{\textbf{Spatial bias in MVBench (4).} A single frame containing all characters inquired is sufficient to answer MVBench.}
\end{figure*}

\begin{figure*}
\centering

\begin{subfigure}{0.95\textwidth}
    \includegraphics[width=\linewidth]{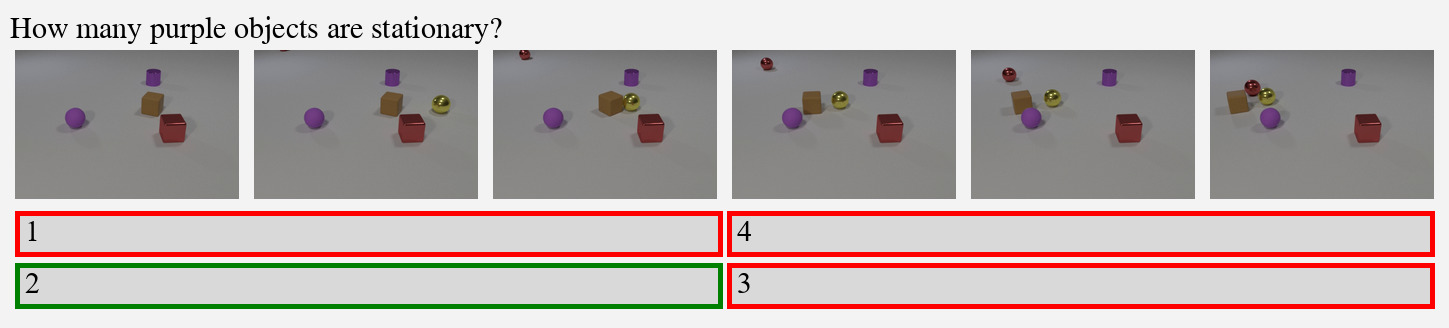}
\end{subfigure}

\begin{subfigure}{0.95\textwidth}
    \includegraphics[width=\linewidth]{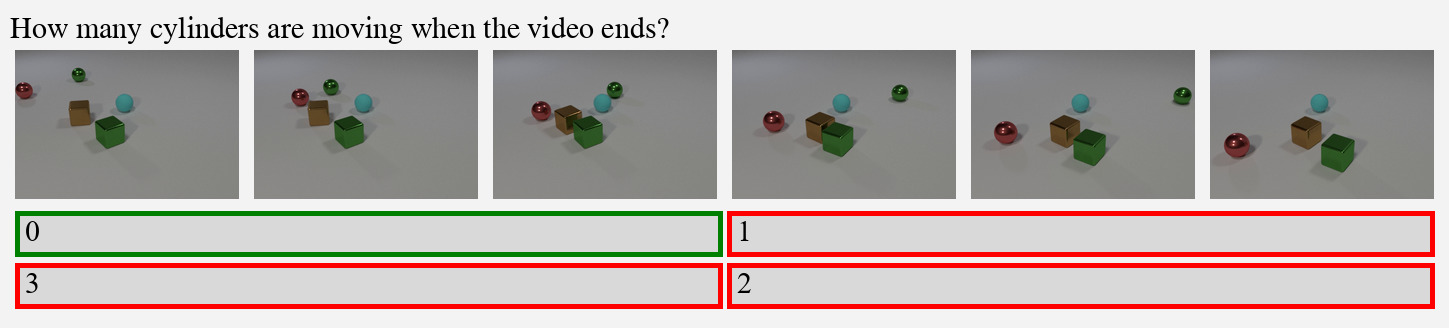}
\end{subfigure}

\begin{subfigure}{0.95\textwidth}
    \includegraphics[width=\linewidth]{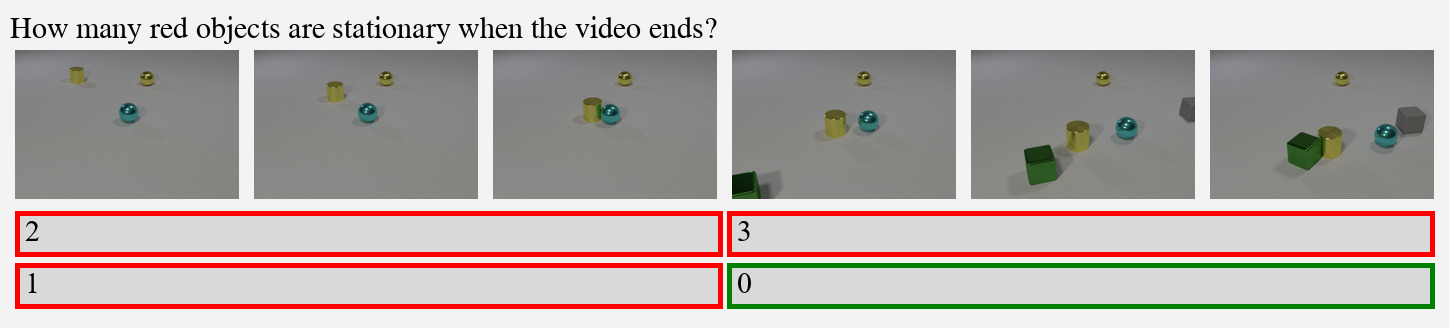}
\end{subfigure}

\begin{subfigure}{0.95\textwidth}
    \includegraphics[width=\linewidth]{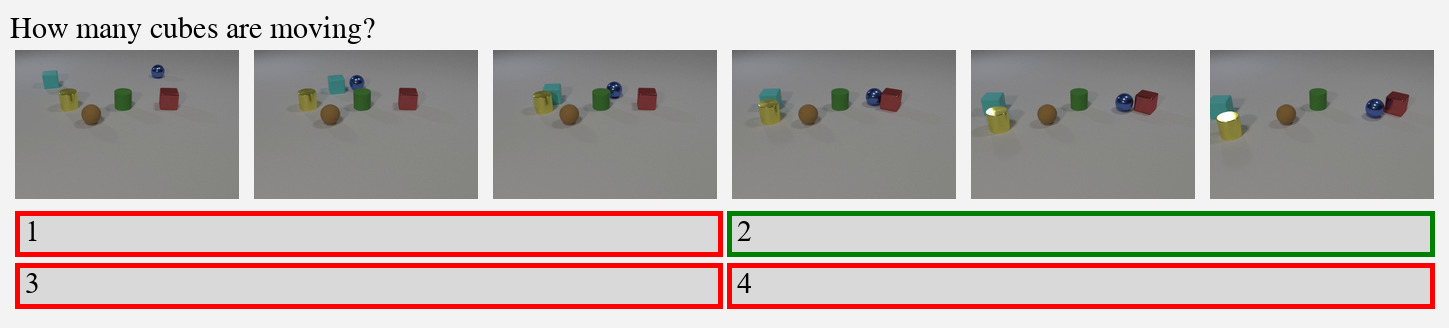}
\end{subfigure}

\begin{subfigure}{0.95\textwidth}
    \includegraphics[width=\linewidth]{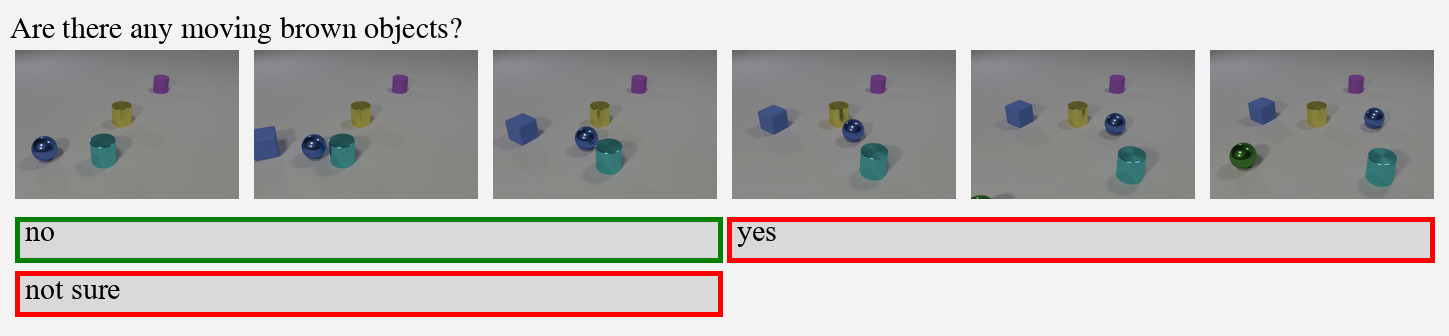}
\end{subfigure}

\caption{\textbf{Spatial bias in MVBench (5).} Temporal understanding is not needed as a single frame suffices to solve these questions on MVBench.}
\end{figure*}

\begin{figure*}
\centering
\centering
\begin{subfigure}{0.95\textwidth}
    \includegraphics[width=\linewidth]{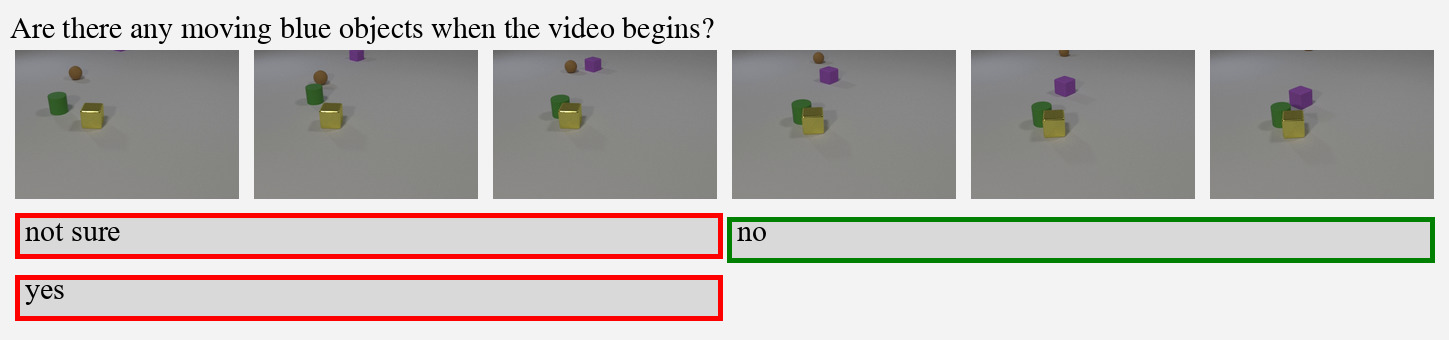}
\end{subfigure}

\begin{subfigure}{0.95\textwidth}
    \includegraphics[width=\linewidth]{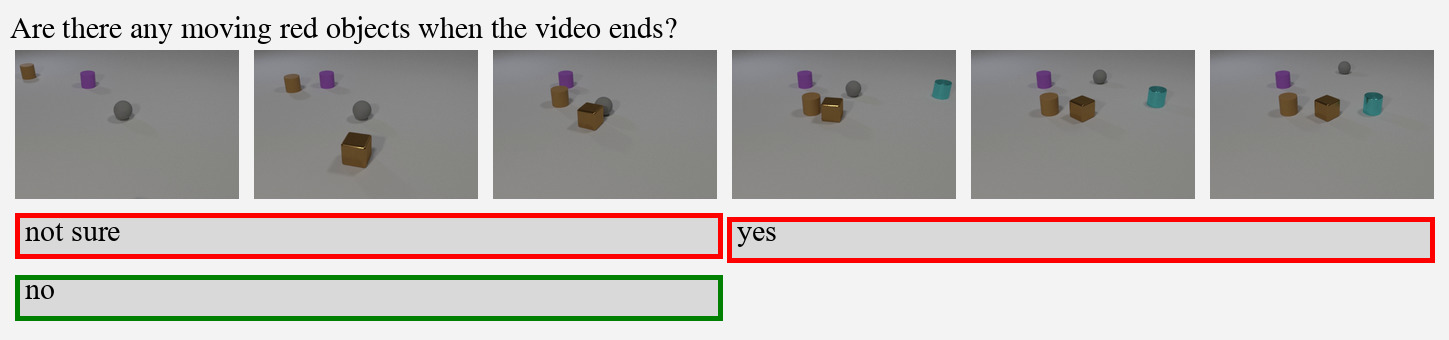}
\end{subfigure}

\begin{subfigure}{0.95\textwidth}
    \includegraphics[width=\linewidth]{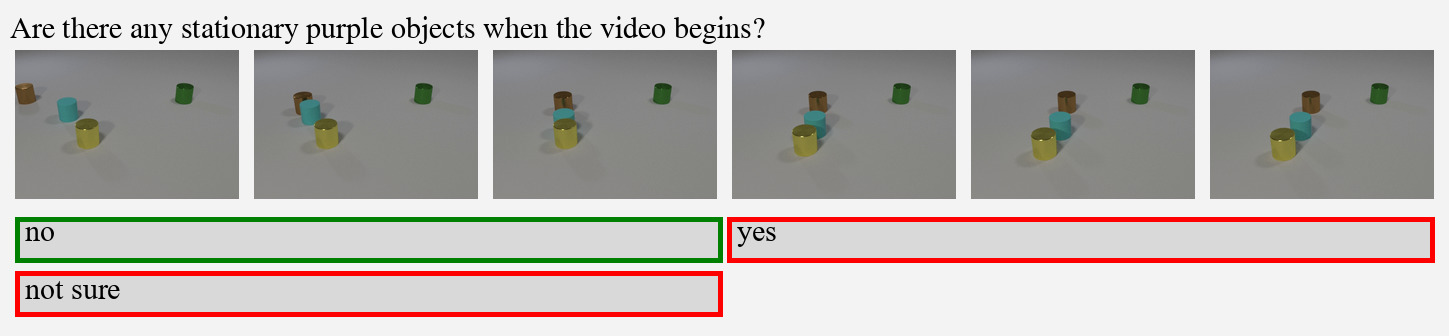}
\end{subfigure}

\begin{subfigure}{0.95\textwidth}
    \includegraphics[width=\linewidth]{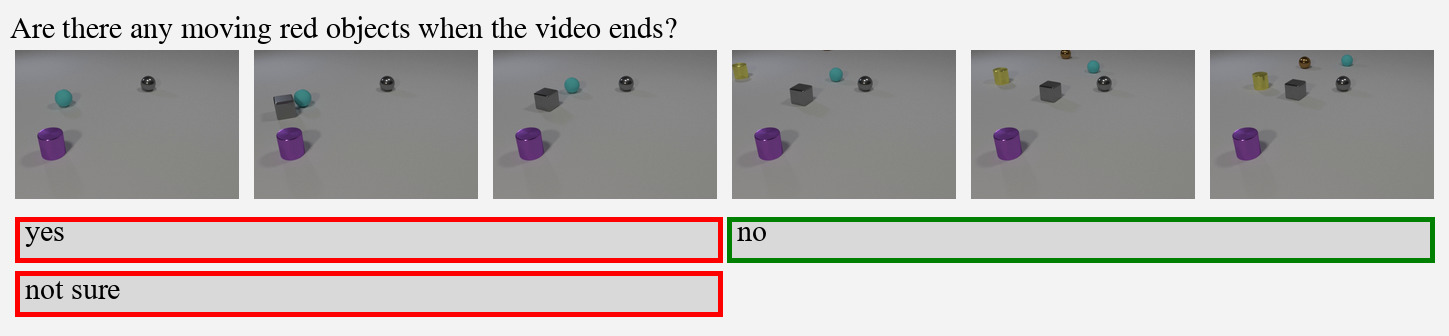}
\end{subfigure}

\begin{subfigure}{0.95\textwidth}
    \includegraphics[width=\linewidth]{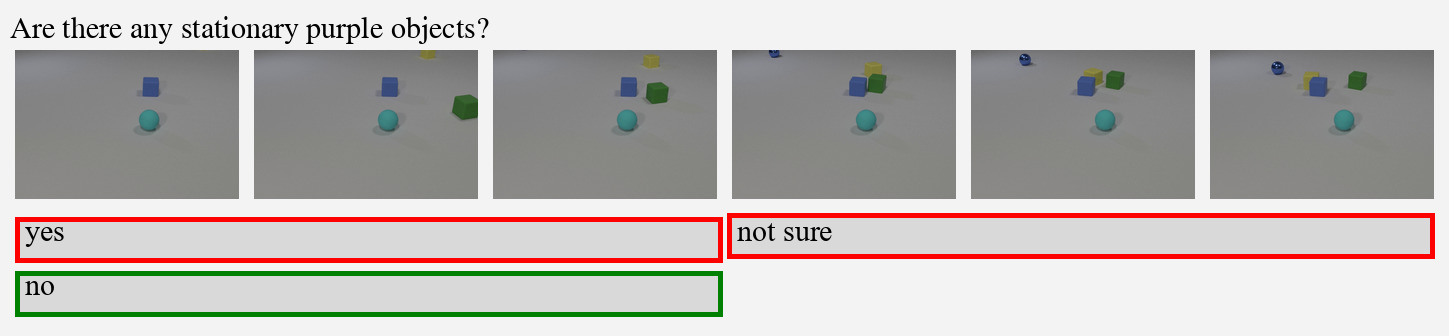}
\end{subfigure}
\caption{\textbf{Spatial bias in MVBench (6).} Temporal understanding is not needed as a single frame suffices to solve these questions on MVBench.}
\end{figure*}

\begin{figure*}
\centering
\centering
\begin{subfigure}{0.95\textwidth}
    \includegraphics[width=\linewidth]{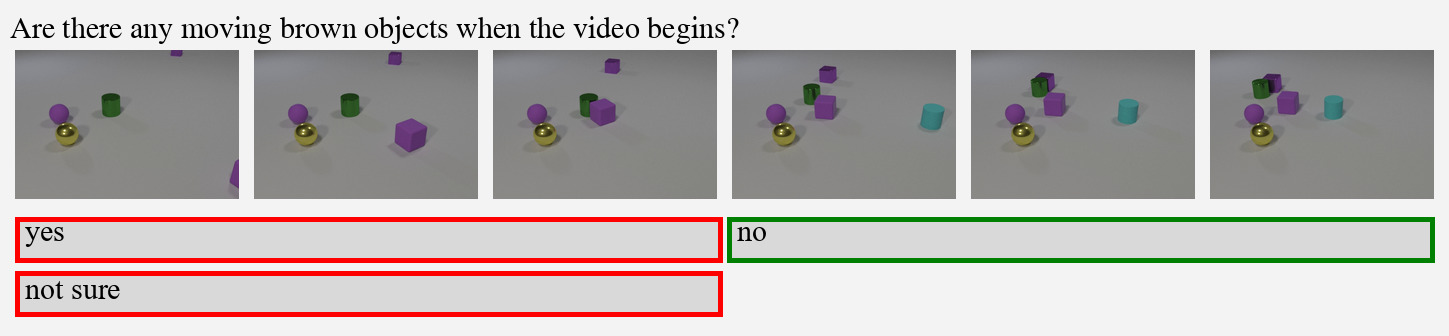}
\end{subfigure}

\begin{subfigure}{0.95\textwidth}
    \includegraphics[width=\linewidth]{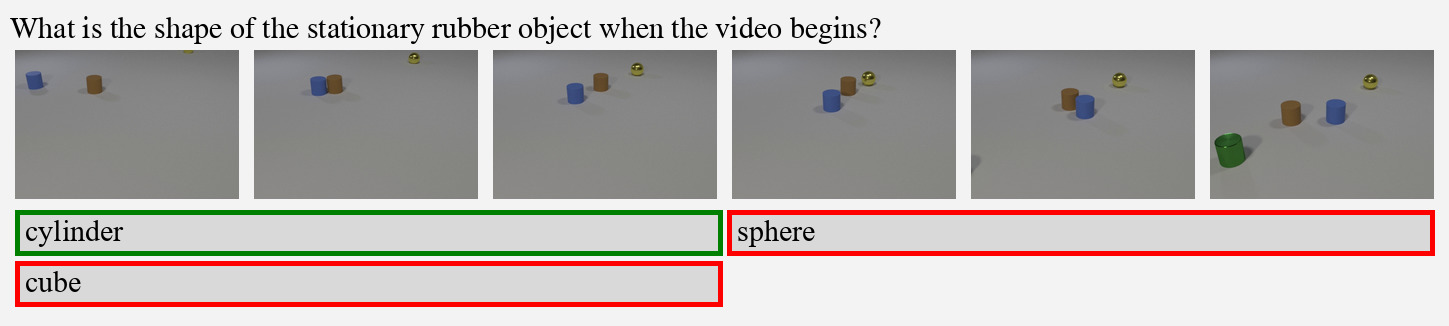}
\end{subfigure}

\begin{subfigure}{0.95\textwidth}
    \includegraphics[width=\linewidth]{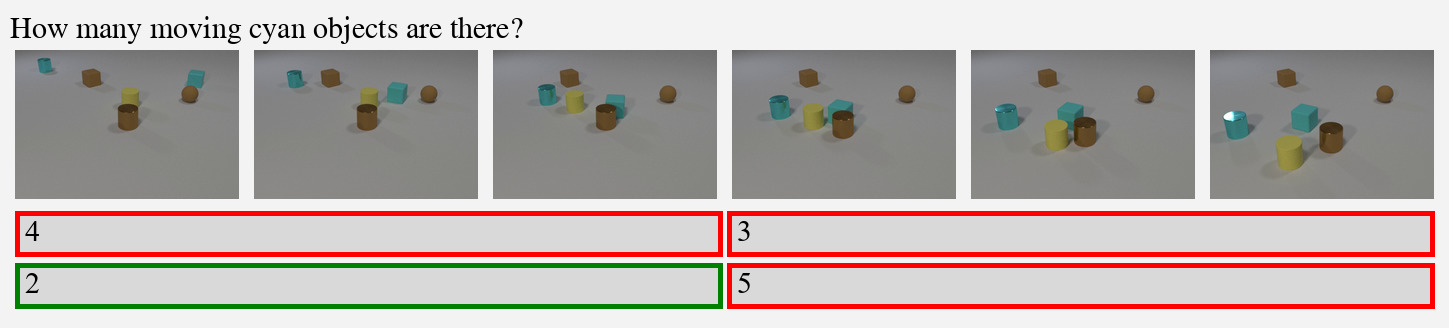}
\end{subfigure}

\begin{subfigure}{0.95\textwidth}
    \includegraphics[width=\linewidth]{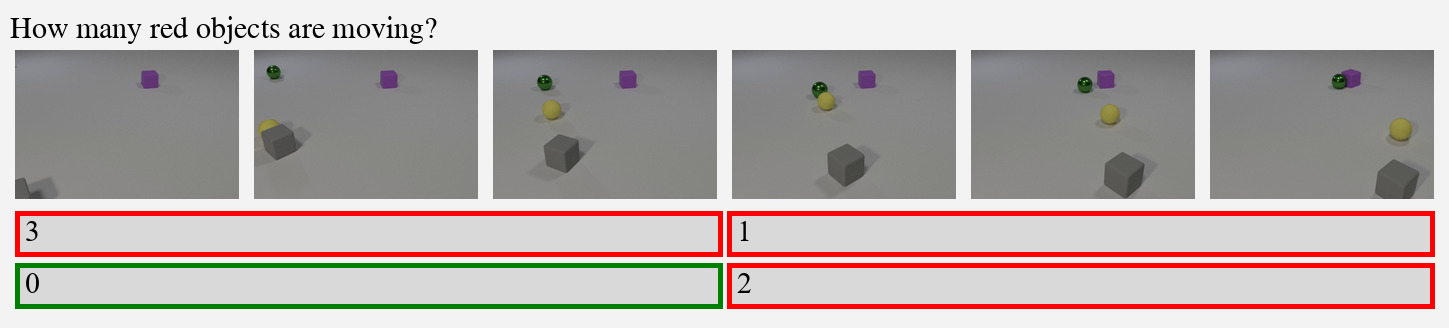}
\end{subfigure}
\begin{subfigure}{0.95\textwidth}
    \includegraphics[width=\linewidth]{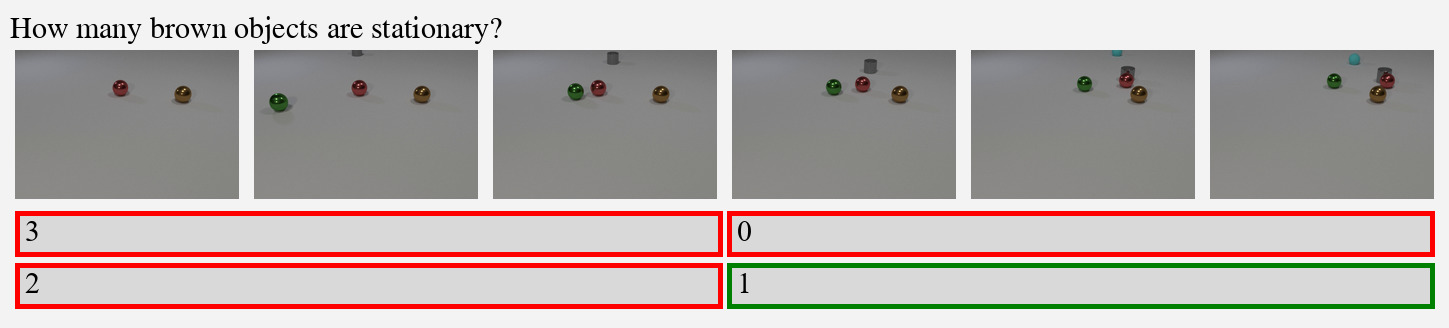}
\end{subfigure}
\caption{\textbf{Spatial bias in MVBench (7).} Temporal understanding is not needed as a single frame suffices to solve these questions on MVBench.}

\end{figure*}

\begin{figure*}
\centering
\begin{subfigure}{0.95\textwidth}
    \includegraphics[width=\linewidth]{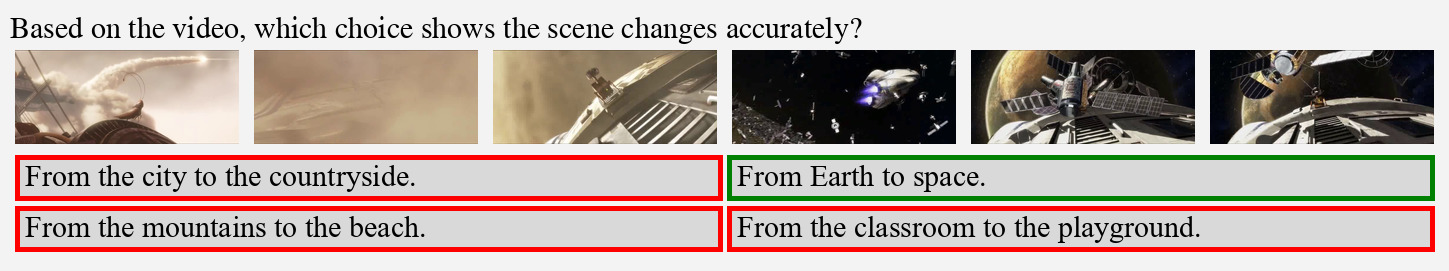}
\end{subfigure}

\begin{subfigure}{0.95\textwidth}
    \includegraphics[width=\linewidth]{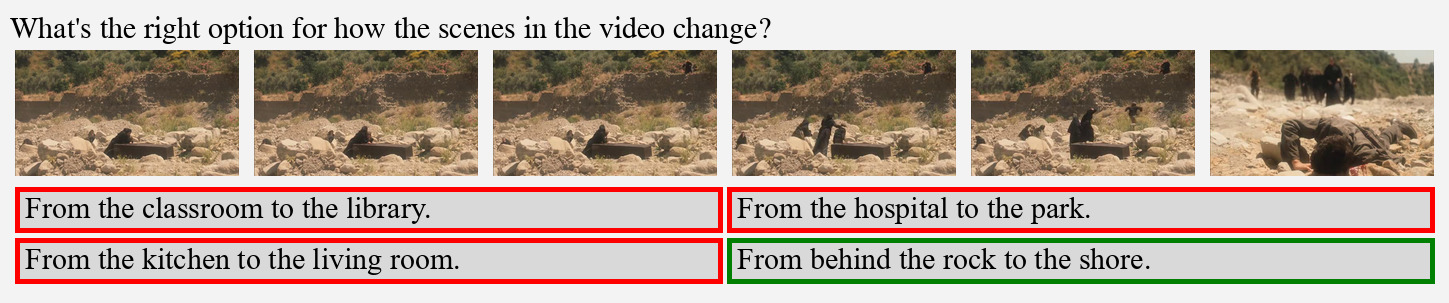}
\end{subfigure}

\begin{subfigure}{0.95\textwidth}
    \includegraphics[width=\linewidth]{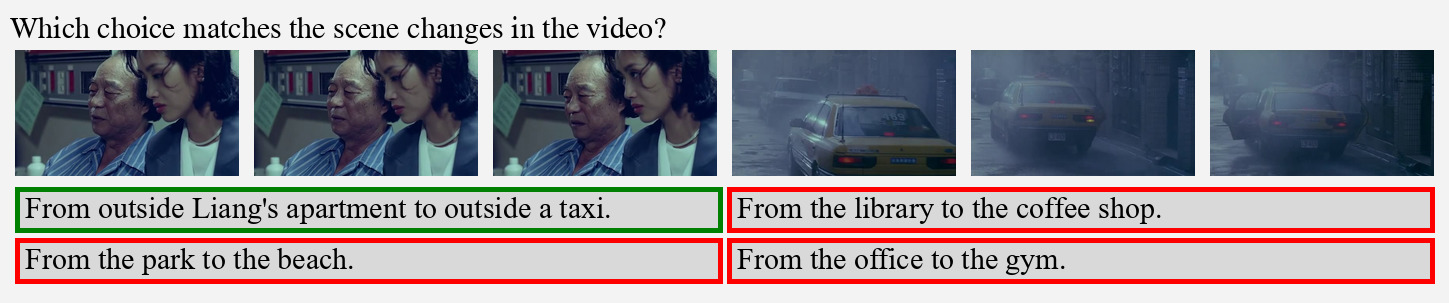}
\end{subfigure}

\begin{subfigure}{0.95\textwidth}
    \includegraphics[width=\linewidth]{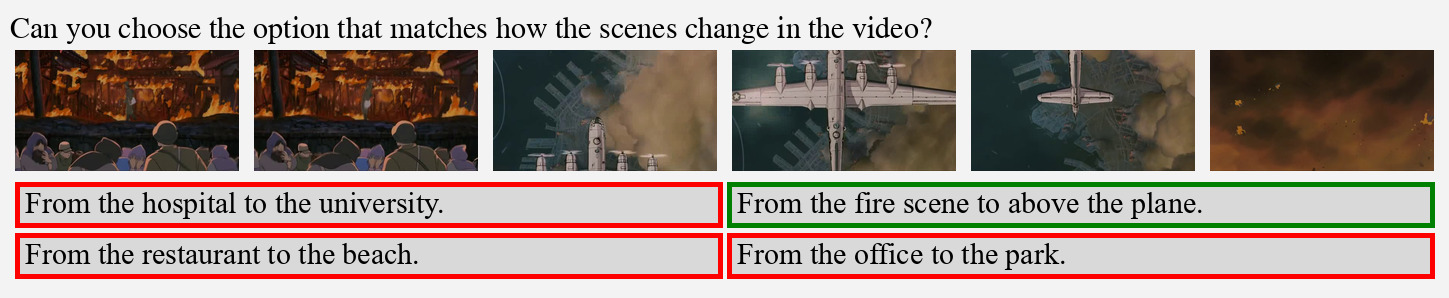}
\end{subfigure}

\begin{subfigure}{0.95\textwidth}
    \includegraphics[width=\linewidth]{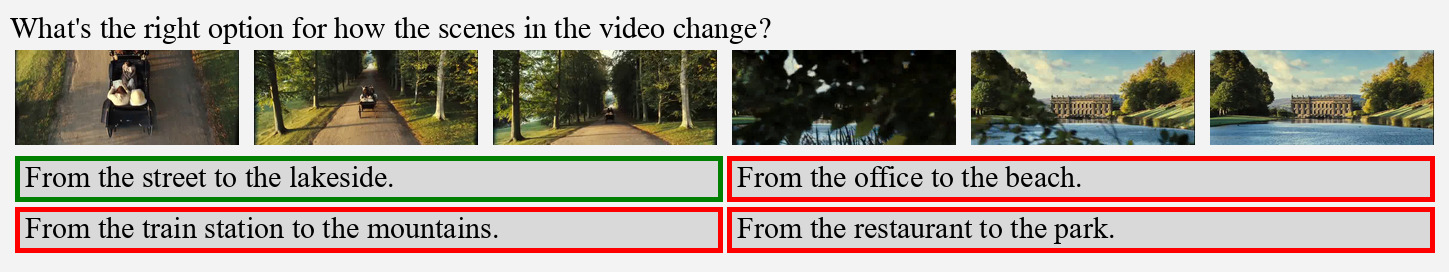}
\end{subfigure}
\caption{\textbf{Spatial bias in MVBench (8).} A single frame is sufficient to discard incorrect candidates, as the scenes described in these candidates never occurred in the video.}
\label{fig:tvbench_spatial_bias_last}
\end{figure*}

\begin{figure*}
\centering
\centering
\begin{subfigure}{0.95\textwidth}
    \includegraphics[width=\linewidth]{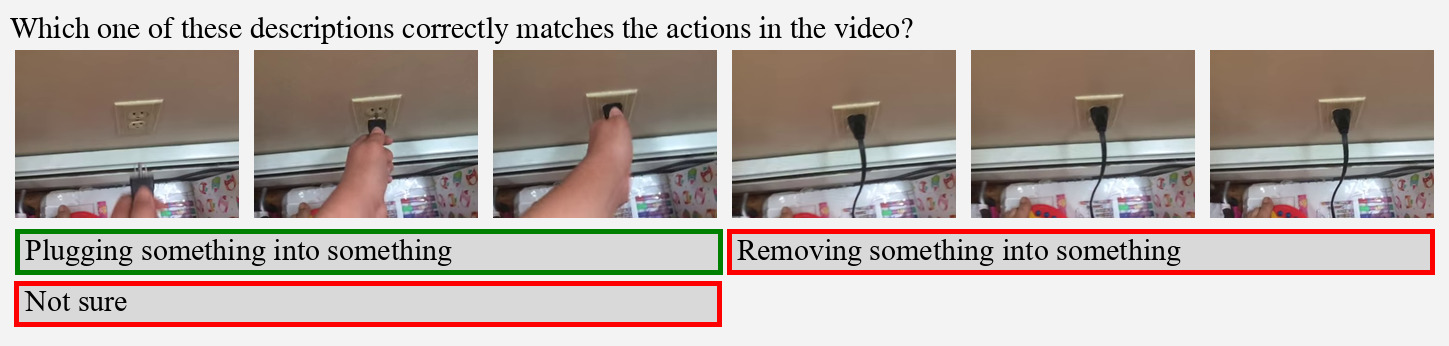}
\end{subfigure}

\begin{subfigure}{0.95\textwidth}
    \includegraphics[width=\linewidth]{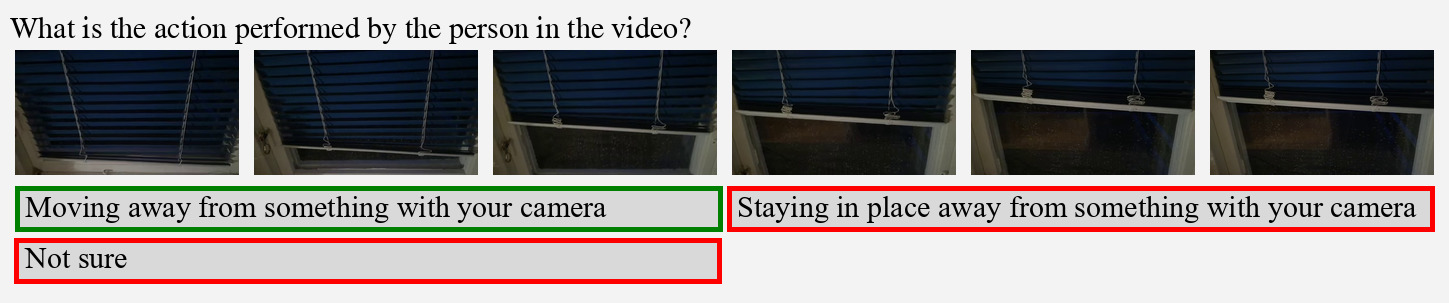}
\end{subfigure}

\begin{subfigure}{0.95\textwidth}
    \includegraphics[width=\linewidth]{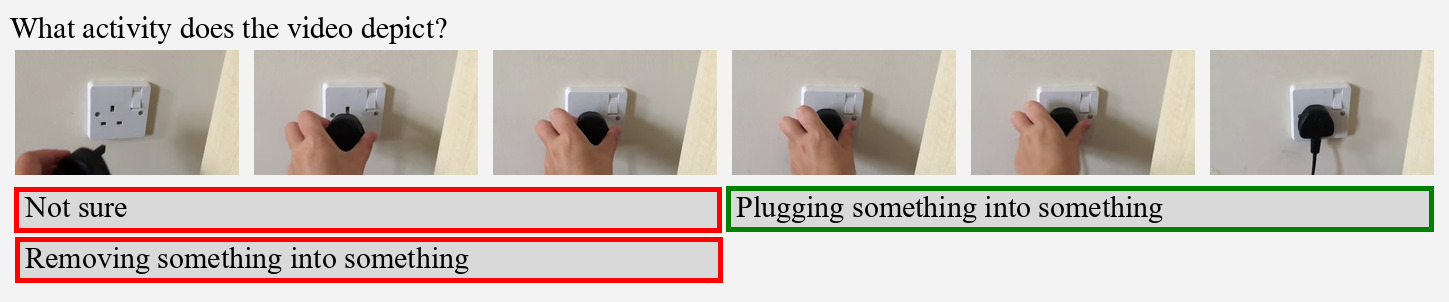}
\end{subfigure}

\begin{subfigure}{0.95\textwidth}
    \includegraphics[width=\linewidth]{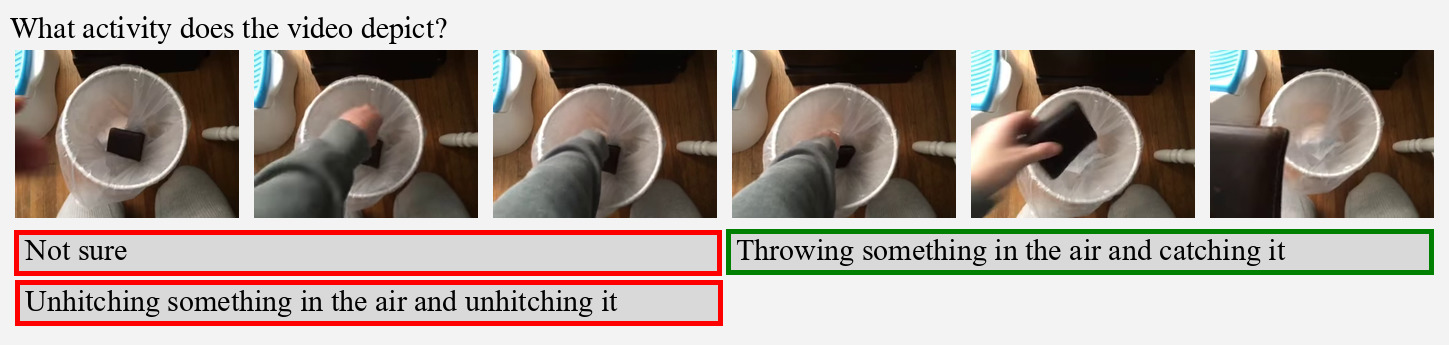}
\end{subfigure}

\begin{subfigure}{0.95\textwidth}
    \includegraphics[width=\linewidth]{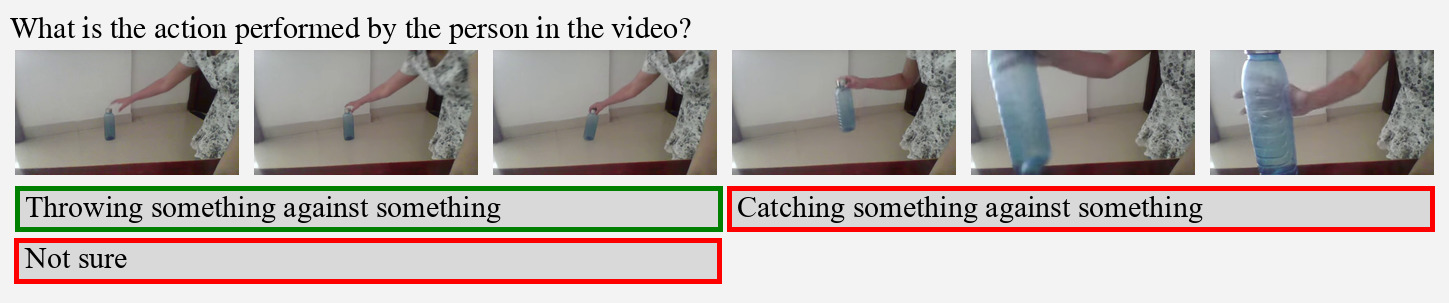}
\end{subfigure}
\caption{\textbf{Textual bias in MVBench (1).} Answer candidates are generated with an LLM. The ``Not sure'' candidate is never correct, while the other incorrect candidate makes no textual sense e.g. ``catching something against something''.}
\label{fig:text_bias_one}
\end{figure*}

\begin{figure*}
\centering
\begin{subfigure}{0.95\textwidth}
    \includegraphics[width=\linewidth]{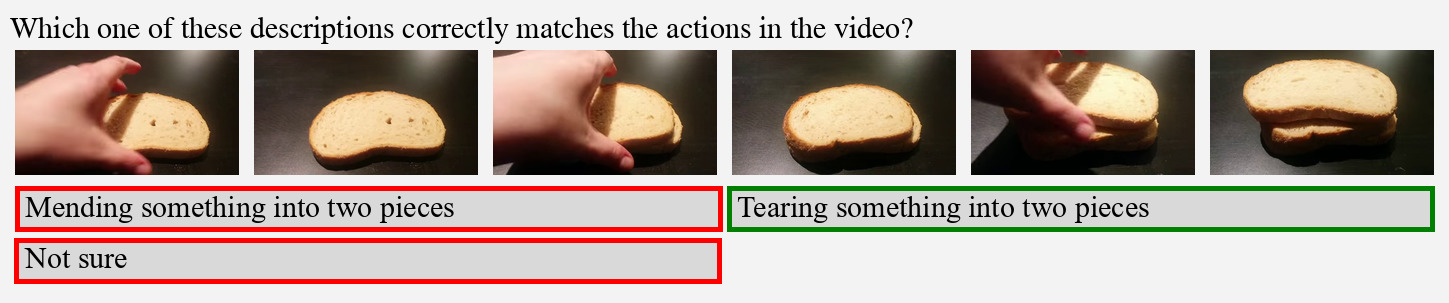}
\end{subfigure}

\begin{subfigure}{0.95\textwidth}
    \includegraphics[width=\linewidth]{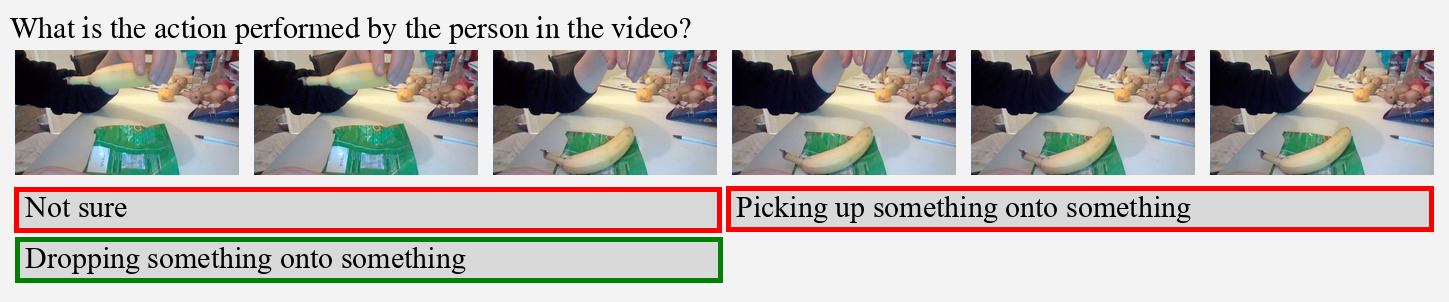}
\end{subfigure}
\caption{\textbf{Textual bias in MVBench (2).} Answer candidates are generated with an LLM. The ``Not sure'' candidate is never correct, while the other incorrect candidate makes no textual sense e.g. ``catching something against something''.}
\label{fig:text_bias_one-continued}
\end{figure*}

\begin{figure*}
\centering
\begin{subfigure}{0.95\textwidth}
    \includegraphics[width=\linewidth]{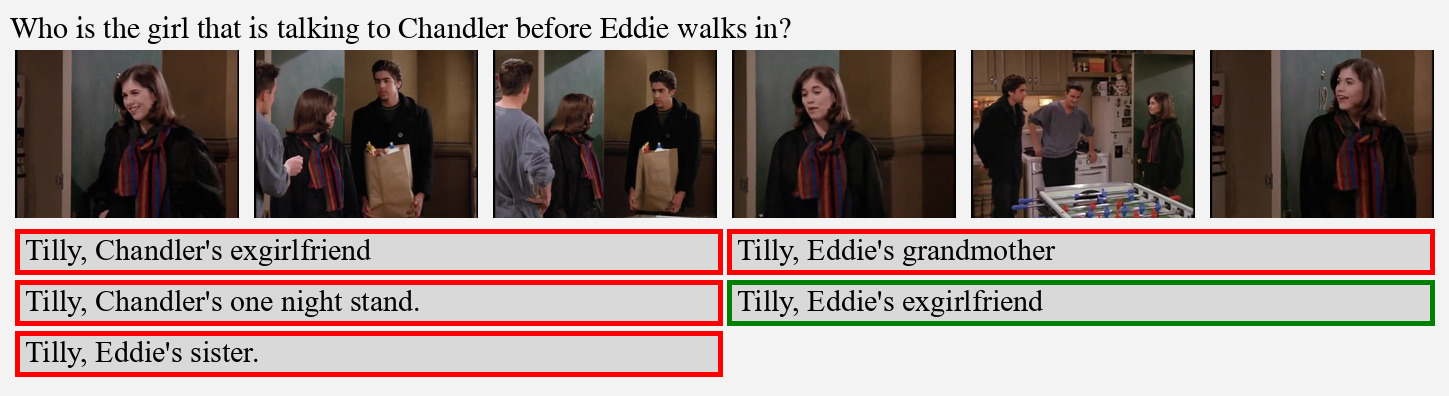}
\end{subfigure}

\begin{subfigure}{0.95\textwidth}
    \includegraphics[width=\linewidth]{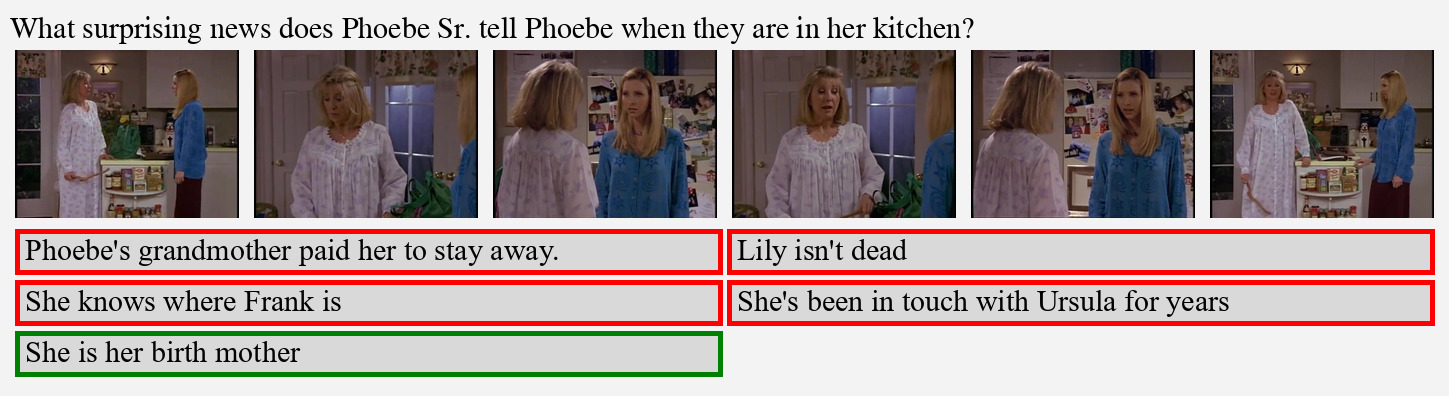}
\end{subfigure}

    \caption{\textbf{Textual bias in MVBench (3): Overreliance on world knowledge.} Answers can be inferred using world knowledge of TV shows. Questions cannot be answered from the video only as answers contain e.g. character names of the TV show.}
\end{figure*}

\begin{figure*}
\centering
\begin{subfigure}{0.95\textwidth}
    \includegraphics[width=\linewidth]{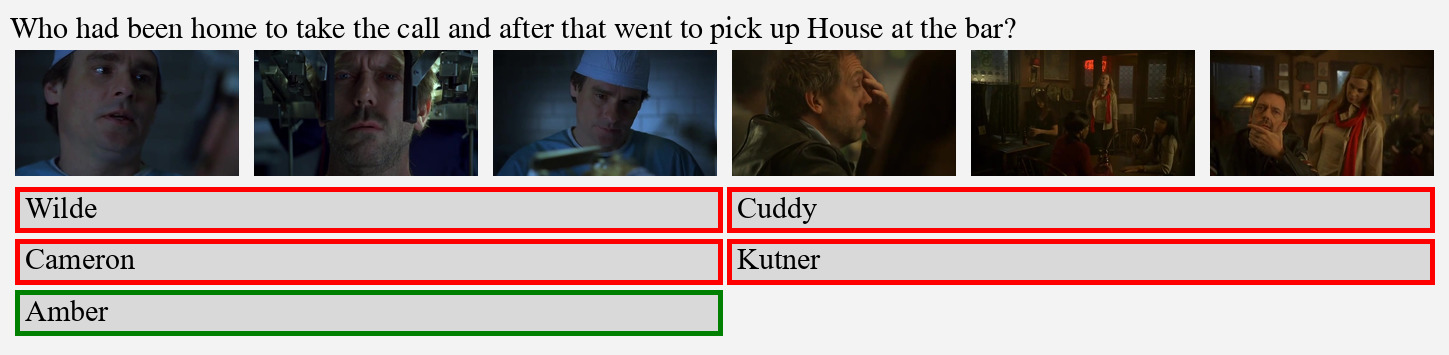}
\end{subfigure}

\begin{subfigure}{0.95\textwidth}
    \includegraphics[width=\linewidth]{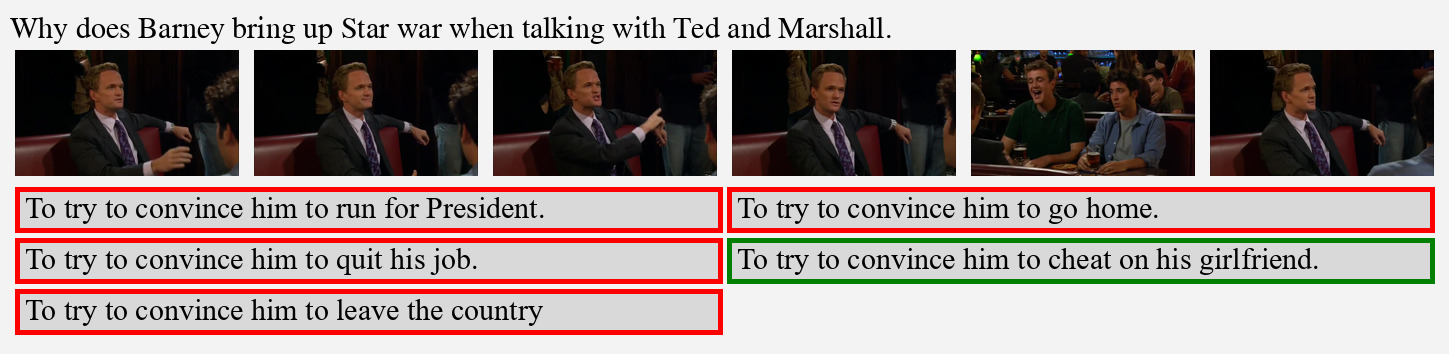}
\end{subfigure}

\begin{subfigure}{0.95\textwidth}
    \includegraphics[width=\linewidth]{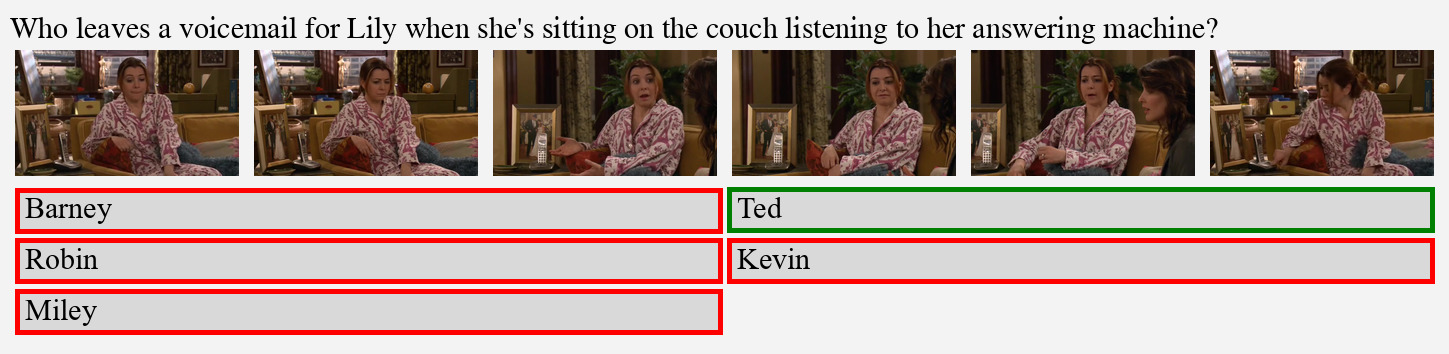}
\end{subfigure}

\begin{subfigure}{0.95\textwidth}
    \includegraphics[width=\linewidth]{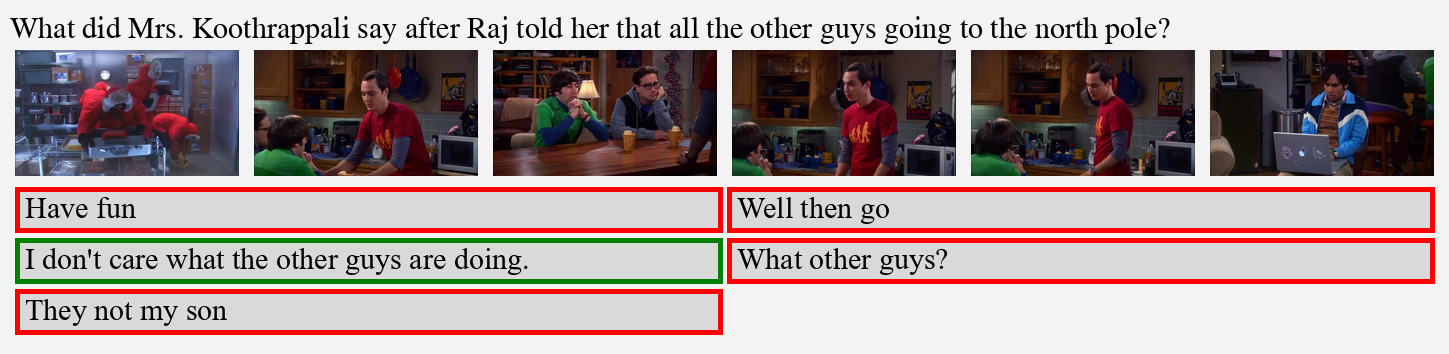}
\end{subfigure}

    \caption{\textbf{Textual bias in MVBench (4): Overreliance on world knowledge.} Answers can be inferred using world knowledge of TV shows. Questions cannot be answered from the video only as answers contain e.g. character names of the TV show.}
\end{figure*}

\begin{figure*}
\centering
\centering
\begin{subfigure}{0.95\textwidth}
    \includegraphics[width=\linewidth]{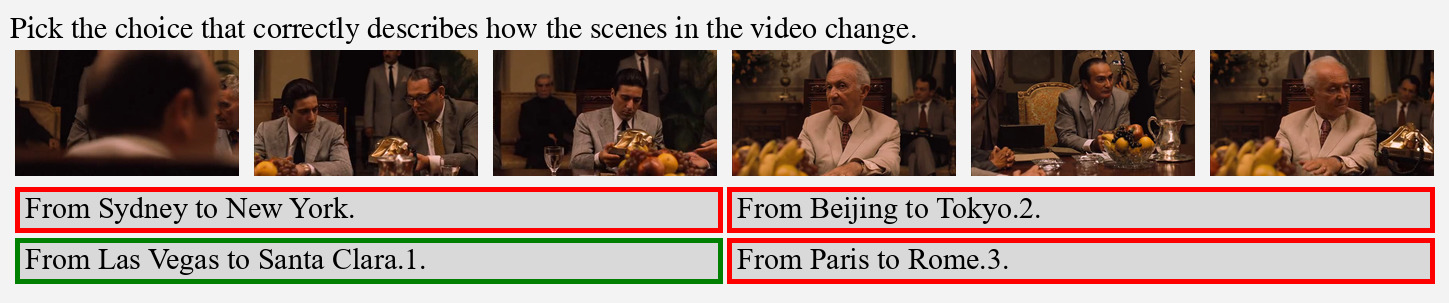}
\end{subfigure}

\begin{subfigure}{0.95\textwidth}
    \includegraphics[width=\linewidth]{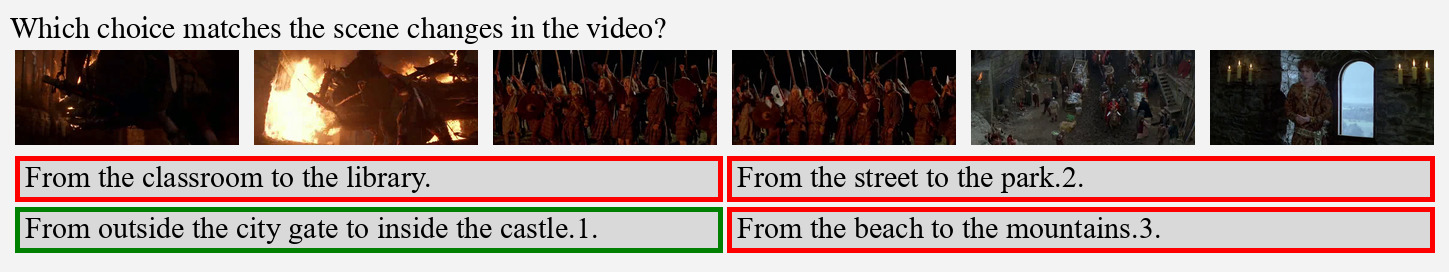}
\end{subfigure}
\begin{subfigure}{0.95\textwidth}
    \centering
    \includegraphics[width=\linewidth]{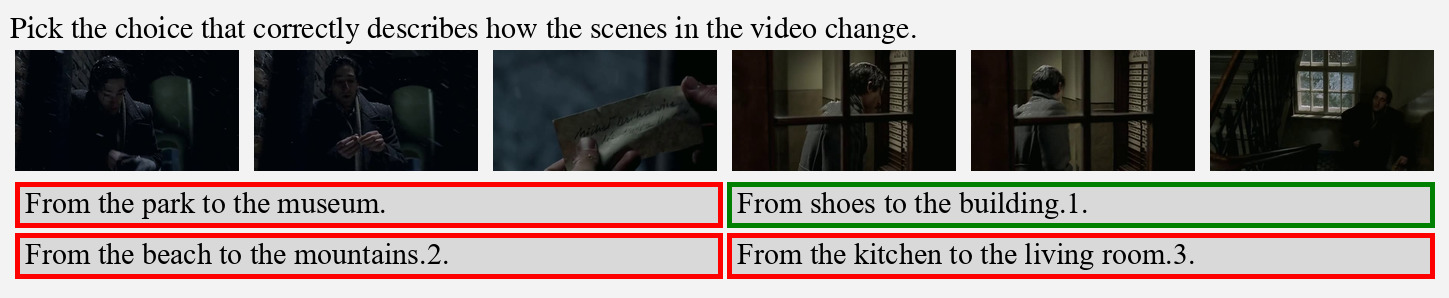}
\end{subfigure}
\begin{subfigure}{0.95\textwidth}
    \centering
    \includegraphics[width=\linewidth]{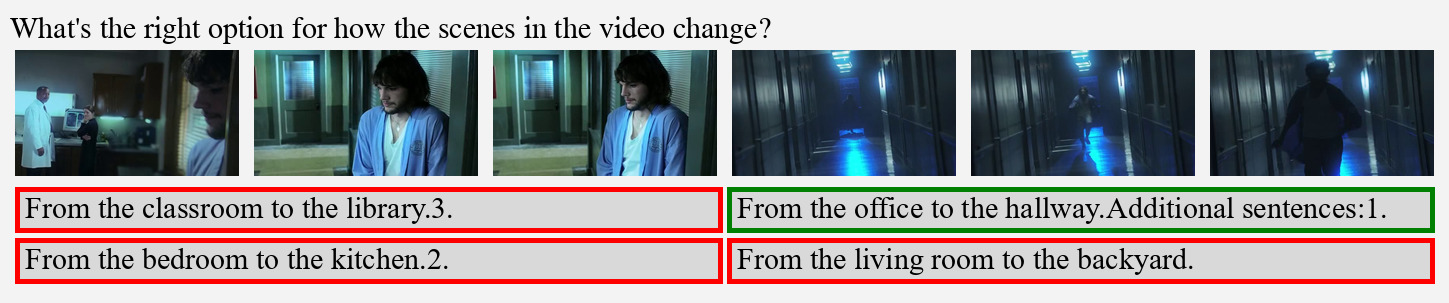}
\end{subfigure}
\begin{subfigure}{0.95\textwidth}
    \includegraphics[width=\linewidth]{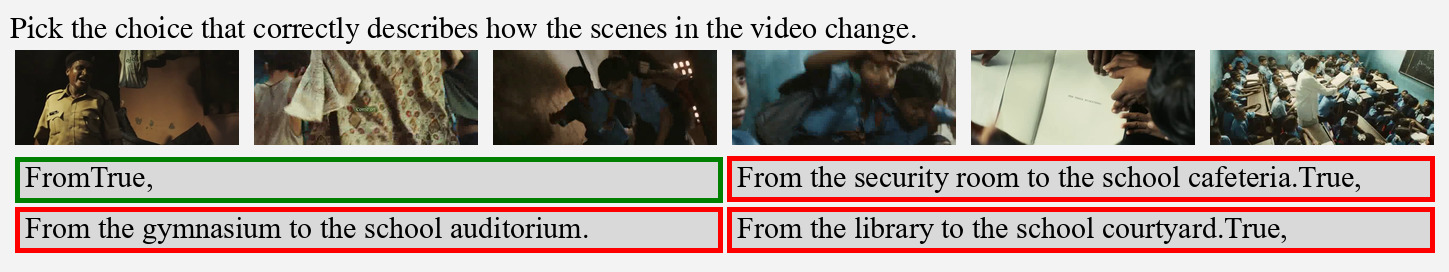}
\end{subfigure}
\caption{\textbf{Textual bias in MVBench (5).} Some candidates in MVBench contain artifacts due to the automatic LLM-based generation process. In these cases, the correct answers end with ``1'' or contain the keyword ``true''.}
\end{figure*}

\begin{figure*}
\centering
    \centering
    \includegraphics[width=\linewidth]{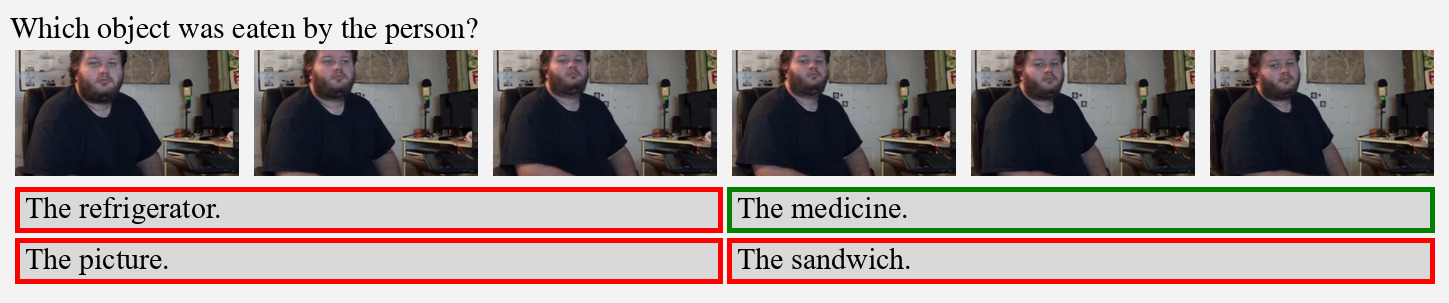}
    \caption{\textbf{Textual bias in MVBench (6).} Since two of the candidate objects cannot be eaten, the choice is reduced to two remaining candidates.}
\end{figure*}

\begin{figure*}
\centering
\centering
\begin{subfigure}{0.95\textwidth}
    \includegraphics[width=\linewidth]{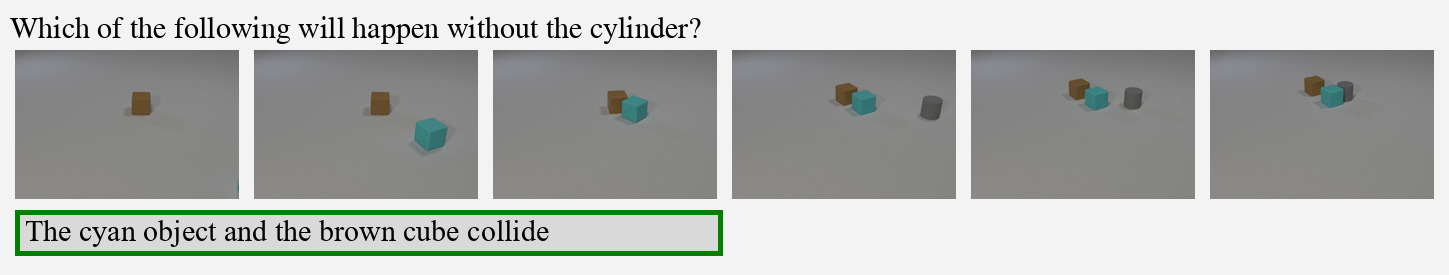}
\end{subfigure}

\begin{subfigure}{0.95\textwidth}
    \includegraphics[width=\linewidth]{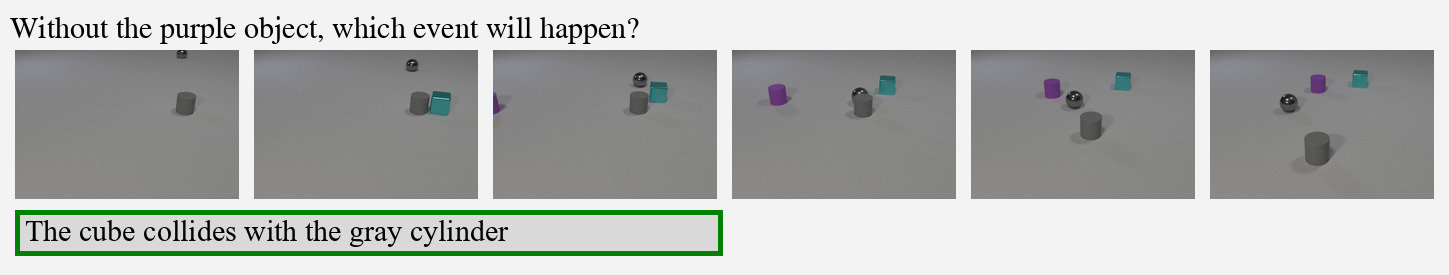}
\end{subfigure}
    \caption{\textbf{Textual bias in MVBench (7).} Just one candidate answer is provided for some QA pairs on MVBench.}
\label{fig:tvbench_last}
\end{figure*}

\fi

\end{document}